\documentclass[nonacm, sigconf, review=false, anonymous=false, screen=true]{acmart}

\renewcommand\footnotetextcopyrightpermission[1]{}

\AtBeginDocument{
  \providecommand\BibTeX{{
    \normalfont B\kern-0.5em{\scshape i\kern-0.25em b}\kern-0.8em\TeX}}}

\newcommand{\f}{\boldsymbol}
\newcommand{\alg}{(DQN-HER) }
\newcommand{\LP}{LP }
\newcommand{\LPcompact}{LP}
\newcommand{\LSP}{L\&SP }
\newcommand{\LSPcompact}{L\&SP}
\newcommand{\btc}{BTCUSD}
\newcommand{\btctwenty}{BTCUSDT}
\newcommand{\nifty}{NIFTY50}
\newcommand{\niftytwenty}{NIFTY50}

\newcommand{\AAPL}{AAPL}
\newcommand{\SPY}{SPY}
\newcommand{\XRP}{XRPUSDT}
\newcommand{\ETH}{ETHUSDT}

\newtheorem{remark}{Remark}

\usepackage{algorithm}
\usepackage{algorithmic}
\usepackage{subcaption}
\newcommand{\resetline}[1]{\setcounter{ALC@line}{\numexpr#1-1}}
\usepackage[]{todonotes}

\setcopyright{none}

\begin{document}

\makeatletter
\def\@ACM@checkaffil{
    \if@ACM@instpresent\else
    \ClassWarningNoLine{\@classname}{No institution present for an affiliation}%
    \fi
    \if@ACM@citypresent\else
    \ClassWarningNoLine{\@classname}{No city present for an affiliation}%
    \fi
    \if@ACM@countrypresent\else
        \ClassWarningNoLine{\@classname}{No country present for an affiliation}%
    \fi
}
\makeatother

\title[Multi-Objective reward generalization ... for applications in single-asset trading]{Multi-Objective reward generalization: Improving performance of Deep Reinforcement Learning for applications in single-asset trading}

\author{Federico Cornalba}
\authornote{These authors contributed equally to this work} 
\authornote{Funded by the Austrian Science Fund (FWF) through the project F65 (previously funded by the European Union's Horizon 2020 research and innovation programme under the Marie Sk\l{}odowska-Curie grant agreement No. 754411)} 
\authornote{Research work primarily conducted at Trality GmbH, Vienna, Austria} 
\affiliation{%
  \institution{Institute of Science and Technology Austria (ISTA)}}
\email{federico.cornalba@ista.ac.at} 

\author{Constantin Disselkamp}
\authornotemark[1]
\authornote{Received funding by the Austrian Research Promotion Agency (FFG)} 
\authornotemark[3]
\affiliation{%
  \institution{TRALITY GmbH}
  \streetaddress{P.O. Box 1212}
  \postcode{1070}
}

\email{constantin@trality.com}

\author{Davide Scassola}
\authornotemark[1]
\authornotemark[4]
\authornotemark[3]
\affiliation{%
  \institution{University of Trieste}
  \streetaddress{P.O. Box 1212}
  \postcode{1070}
}
\email{davide.scassola@phd.units.it}

\author{Christopher Helf}
\authornotemark[4]
\authornotemark[3]
\affiliation{%
  \institution{TRALITY GmbH}
  \streetaddress{P.O. Box 1212}
  \postcode{1070}
}
\email{christopher@trality.com}

\renewcommand{\shortauthors}{Cornalba, et al.}

\settopmatter{printacmref=false}

\begin{abstract}
We investigate the potential of Multi-Objective, Deep Reinforcement Learning for stock and cryptocurrency single-asset trading: in particular, we consider a Multi-Objective algorithm which generalizes the reward functions and discount factor
(i.e., these components are not specified a priori, but incorporated in the learning process).

  Firstly, using several important assets (cryptocurrency pairs \btc, \ETH, \XRP, and stock indexes \AAPL, \SPY, \nifty{}), we verify the reward generalization property of the proposed Multi-Objective algorithm,
    and provide preliminary statistical evidence showing increased predictive stability over the corresponding Single-Objective strategy. 
    Secondly, we show that the Multi-Objective algorithm has a clear edge over
    the corresponding Single-Objective strategy when the reward mechanism is
    sparse (i.e., when non-null feedback is infrequent over time).
    Finally, we discuss the generalization properties with respect to the discount factor.
  
    The entirety of our code is provided in open source format.
\end{abstract}



\settopmatter{printfolios=true}
\maketitle

\section{Introduction}
The algorithm developed by Fontaine and Friedman \cite{friedman2018generalizing} (see also~\cite{castelletti2012tree, ernst2005tree}) is a specific declination of Multi-Objective Reinforcement Learning (RL): 
while we postpone rigorous details to the next sections below, we can -- for now -- informally describe this as a RL declination which allows to simultaneously train all possible strategies associated with exploring a dynamic environment. 
Specifically, each strategy is identified by the reward mechanism associated with a given linear combination of multiple, pre-specified (and possibly conflicting) objective functions. 
This simultaneous learning process over all possible strategies, which gives the user the freedom to specify the combination of the objective functions after the training has been completed, makes the RL declination in \cite{friedman2018generalizing} highly interpretable and versatile.

While the techniques in \cite{friedman2018generalizing} have so far been successfully
 applied in several fields,
 applications of such a methodology for financial purposes are -- to the best of our knowledge -- still lacking. 
In this paper, we show the potential entailed by using such a Multi-Objective Reinforcement Learning algorithm
 -- along with meaningful variations of the same -- in the context of single-asset 
 stock and cryptocurrency applications. 
 We validate our results by deploying the algorithm on several
  important assets, namely: cryptocurrency pairs \btc, \ETH, \XRP, and stock indexes \AAPL, \SPY, \nifty. Additionally, we discuss the generalization with respect to the discount factor parameter\footnote{
        We stress that the term \emph{discount factor} in this paper refers to a RL 
        time-regularization parameter (see \eqref{CumulativeReward} below),
        and should not be confused with the finance-related homonym.}.

We now give more precise details for the terminology so far informally introduced.

\subsection{Reinforcement Learning: basics}
Reinforcement Learning (RL) is a subfield of Machine Learning specifically designed to handle learning processes for problems which involve a dynamic interaction with a given underlying environment. 
RL techniques have initially been deployed -- among many -- in the field of gaming \cite{mnih2013playing}, robotics \cite{kober2013reinforcement}, personalized recommendations \cite{zheng2018drn}, and resource management \cite{mao2016resource}.

\subsubsection{Single-Objective RL}
A RL algorithm learns to use the set of observable state variables $\f{s}$ (describing the current \emph{state} of the environment) to take the most appropriate admissible \emph{action} $a$. 
For a general RL algorithm, the state variables and action $(\f{s},a)$ and the environment's intrinsic stochasticity determine the next state of the environment that the algorithm will visit: 
the exact way in which this task is accomplished varies depending on the specific algorithm.
For example, the RL algorithms of so-called \emph{critic--only} form (see \cite{sutton2018reinforcement}) aim to maximize a single, cumulative reward (accounting for all actions taken in a given episode), which is in turn based on a pre-specified state--action \emph{reward function} $r$ (assigning a numerical reward $r(\f{s},a)$ to every pair $(\f{s},a)$ of given state and action undertaken).
The algorithm uses several \emph{episodes} (each accounting for an exploration of the environment which ends when a pre-specified end-state is reached) to train its decision-making capabilities (by progressively updating the so-called \emph{$Q$-values} \cite{sutton2018reinforcement} throughout the episodes).


\subsubsection{Multi-Objective RL}\label{MultiObjectiveRLGeneral} 
Multi-Objective\footnote{We use the terms `Multi/Single-Objective' and `Multi/Single-Reward' interchangeably throughout the paper.}  RL is a declination of RL devoted to learning in environments with vector-valued -- rather than scalar -- rewards. 
Such a setting -- which allows to cover realistic scenarios in which conflicting metrics are present -- has been gaining a lot of traction lately. Two rather comprehensive surveys on the topic\footnote{also including the multi-agent case, which we do not treat in this work} that we are aware of are \cite{ruadulescu2020multi,roijers2013survey}. 
With methodologies including -- among many others -- Pareto front-type analysis \cite{zitzler2008quality, reymond2019pareto}, dynamic multi-criteria average reward reinforcement learning \cite{natarajan2005dynamic}, 
convex hull value iteration \cite{barrett2008learning}, 
Hindsight Experience Replay techniques \cite{andrychowicz2017hindsight}, 
dynamic weights computation \cite{abels2019dynamic}, 
tunable dynamics in agent-based simulation \cite{kallstrom2019tunable}, 
Deep Optimistic Linear Support Learning (DOL) for high-dimensional decision problems \cite{mossalam2016multi}, Deep Q-networks techniques \cite{nguyen2020multi,tajmajer2017multi,tajmajer2018modular}, 
and collaborative agents systems \cite{dusparic2009distributed},
it is safe to say that the Multi-Objective RL paradigm is well-established in several non-finance related applications. 

\subsection{Reinforcement Learning in finance}

The problems we are interested in are related to using RL for profitable, risk-reduced trading of financial tools based on historical data. 
These problems are usually quite challenging to tackle using AI, and this is mostly due to three factors: 
i) high data noisiness of financial environment; 
ii) subjective definition of the financial environment, and 
iii) non reproducibility of the financial environment (only a single copy of any given time series is available for training).

Despite these difficulties, a prolific literature is available: the summary report \cite{fischer2018reinforcement} provides an exhaustive overview of the main works associated with the three most commonly used RL paradigms (i.e., \emph{critic}-, \emph{actor}-, and \emph{actor/critic}- based techniques) for financial applications.
Among critic-based works, we find superiority of RL over standard supervised learning approaches \cite{neuneier1995optimal}, performance improvement assessments with respect to varying reward functions and hyperparameters \cite{RePEc:ven:wpaper:2014:15}, Deep Q-learning (DQL) extensions to trading systems \cite{jin2016portfolio}, evolutionary reinforcement learning in the form of genetic algorithms \cite{dempster2001computational,gu2011trading}, identification of seasonal effects \cite{tan2011stock, eilers2014intelligent}, high-frequency analysis \cite{sherstov2004three}, trade execution optimization \cite{nevmyvaka2006reinforcement}, dynamical allocation of high-dimensional assets, \cite{kaur2017algorithmic, jangmin2006adaptive}, and hedging basis risk assessment \cite{watts2015hedging}. 
For actor-based methods, we mention recurrent reinforcement learning baselines \cite{moody1998performance, gold2003fx}, multi-layered risk management systems \cite{dempster2006automated}, and high-level feature extraction \cite{deng2016deep, jiang2017deep}.
Finally, we quote \cite{li2007short, bekiros2010heterogeneous, RePEc:sce:scecf1:146} as representatives of the actor/critic-based category.

\subsubsection{Multi-Objective RL in finance}\label{MultiObjectiveRLInFinance} 

While the Multi-Objective RL paradigm is well-established in several non-finance related applications (see discussion of Subsection \ref{MultiObjectiveRLGeneral}), it appears to still be relatively under-explored\footnote{Only counting the publicly available, non-proprietary literature} in the context of financial markets.

In this context, the most commonly taken approach to Multi-Objective RL is to indirectly embed the desired multi-reward effects in parts of the model other than the reward mechanism itself (e.g., collaborating market agents \cite{lee2002multi,lee2007multiagent,lee2002stock}). Another approach is to consider an intrinsic Multi-Objective approach, but without generalization (i.e., the reward weights are set a priori, and are not part of the learning process). This is the case for the two reference works \cite{bisht2020deep,si2017multi}, which we summarize in Section \ref{RelatedWork}.

\section{Our contribution}
We use a generalized, intrinsically Multi-Objective RL strategy for stock and cryptocurrency trading.
We implement this by considering extensions of Multi-Objective Deep Q-Learning RL algorithm with experience replay and target network stabilization given in \cite{friedman2018generalizing}, and deploying it on several important cryptocurrency pairs and stock indexes.

\subsection{Main Results}
Our main findings -- which we have validated on several datasets 
including \AAPL, \SPY, \ETH, \XRP, \btc, \nifty -- are summarized as follows.

\begin{description}
\item $\bullet$ \emph{Generalization}. We show that our Multi-Objective RL algorithm generalizes well between four reward mechanisms (last logarithmic return, Sharpe ratio, average logarithmic return, and a sparse reward triggered by closing positions).
\item $\bullet$ \emph{Stability on prediction}. We use two metrics (Sharpe Ratio and cumulative profits) to show that the prediction of our Multi-Objective algorithm is more stable than the corresponding Single-Objective strategy's.
\item $\bullet$ \emph{Advantage for sparse rewards}. We show that the results of the Multi-Objective algorithm are significantly better than those of the corresponding Single-Objective algorithm in the case of sparse rewards\footnote{i.e., rewards that give non-null feedback infrequently over time}.
\item $\bullet$ \emph{Discount factor generalization}. We provide partial evidence of generalization of the discount factor: this parameter is the RL 
        time-regularization parameter in \eqref{CumulativeReward} below.
\item $\bullet$ \emph{Impact of fees}. As per the nature of the current underlying RL algorithm (for both Multi-Objective
    and Single-Objective), trading fees diminish every performance to zero. 
    This behavior is to be expected, see mitigating circumstances described in 
    Subsection \ref{ZeroFeeSubsec} and Subsection \ref{UnderlyingRL_AlgoSubsec} below.

\end{description}

\subsection{Validation of results}

In order to show the robustness of our analysis, we adhere to the following general strategy.

\begin{description}
\item $\bullet$ \emph{Large number of datasets}. We run our experiments on a large group of single-asset financial time series: in particular, these include cryptocurrency pairs (\btc, \ETH, \XRP) and stock indexes (\AAPL, \SPY, \nifty).
\item $\bullet$ \emph{Different initializations}. In order to detach the impact of the chosen neural network's random initialization from the actual results, we perform several independent trainings for each given dataset we run our algorithm on. This is used to assess the distribution of gains for our algorithm. 
\item $\bullet$ \emph{Cross-validation}. We further confirm the effectiveness of our method using a plain hold-out method and \emph{walk-forward} cross-validation with anchoring (i.e., where the training set starting date is the same for all folds).\end{description}

\subsection{Practical implications and considerations}

We highlight some of the most important aspects of our approach from an applied perspective.

\subsubsection{Simplicity and interpretability}
We deploy an intuitive and interpretable Multi-Objective RL algorithm
in order to account for several -- well established -- reward mechanisms in single-asset trading.
We view this methodology as having a strong applied connotation, and being complementary
to other existing Multi-Objective strategies for financial problems
\cite{lee2002multi,bisht2020deep,si2017multi}.

\subsubsection{Zero-fee setting}\label{ZeroFeeSubsec}
Although our analysis is conducted in a zero-fee market context,
such a context nonetheless applies to many meaningful markets.
For retail traders zero-fee spot trading is possible on multiple markets,
for example \btc spot on the cryptocurrency exchange Binance.
For institutional traders even more opportunities exist, since they can often operate
in low to zero or fixed fee market contexts by directly working
together with market makers.
\subsubsection{Underlying RL algorithm}\label{UnderlyingRL_AlgoSubsec}
Our main focus is always to compare our generalized Multi-Objective RL methodology to a corresponding Single-Objective strategy: 
said differently, we do not so much dwell on increasing the performance of the underlying RL strategy 
(in fact, we choose a rather simple critic-based RL declination), 
as we do on evaluating the difference of the Single-/Multi-Objective methods for the same underlying RL method.
\begin{remark}
The adaptation of this methodology with more sophisticated underlying RL methods is deferred to future works.
\end{remark}

\subsection{Structure of the paper}
We summarize the main contributions of the related works \cite{bisht2020deep,si2017multi} in Section \ref{RelatedWork}. 
We provide the abstract setup of our proposed Algorithm in Section \ref{Abstract}, and fill in the necessary quantitative details in Section \ref{Specifics}. 
After spelling out the main technical features related to our code (see Section \ref{Experiments}), we discuss our main results in Section \ref{Results}.
Conclusions (respectively, future outlook) are given in Section \ref{Conclusions} (respectively, Section \ref{Outlook}). 
The precise details concerning our chosen underlying RL algorithm are provided in the Appendix \ref{AppendixRL_Algos}.

\section{Related work}\label{RelatedWork}

The works \cite{bisht2020deep,si2017multi} use multi-reward scalarization to improve on the following, well-established benchmark strategies in the context of price prediction in single-asset trading: i) an actor-only, RL algorithm with total portfolio value as single reward, and; ii) a standard Buy-and-Hold strategy.
More specifically, the authors take as rewards the average and standard deviations from the classical definition of the Sharpe ratio, and combine them with pre-defined weights to favor risk-adjustment.
In another variation, the resulting scalarized metric is modified to further penalize negative volatility.

The authors use a two-block LSTM neural network to directly map the last previously taken action ({\tt Buy}/{\tt Sell}/{\tt Hold}) and the available state variables to the next action.
The first LSTM block is used for high-level feature extraction, and the other one for reward-oriented decision-making.
From the experimental results the authors conclude superiority of their method over to the two benchmarks in terms of cumulative profit and algorithm convergence, although an analysis of statistical significance is not provided\footnote{such an analysis is extremely tough, given that it is very hard to obtain statistically consistent positive returns in volatile single-asset problems}.

While the scalarization approach is effective, it nonetheless has the downside of having to a priori specify the balance of the individual rewards (via their weights).
This introduces a human factor into the balancing of rewards, and also restricts the scope of the learning process.
Driven by this, we choose not to scalarize the reward metrics, so that the weights can be included in the learning process.
To the very best of our knowledge, ours is the first application of Multi-Reward RL in the sense of \cite{friedman2018generalizing} to financial data.

\section{Abstract definition of the model}\label{Abstract}

We consider variations of the Deep-Q-Learning algorithm with Hindsight Experience Replay and target network stabilization \cite{sutton2018reinforcement} \alg for both standard Single-Reward or Multi-Reward structure (in the sense of \cite{friedman2018generalizing}), and apply them to single asset trading problems.

\subsection{The classical abstract setup}
The basic structure of \alg is concerned with maximizing cumulative rewards of the type
\begin{align}\label{CumulativeReward}
R_t = \sum_{i=t}^{T}{\gamma^{i}r_i},
\end{align}
where $\gamma\in(0,1)$ is the so-called discount factor.
The discount factor determines the time preference of the agent and regularizes the reward as \(T \rightarrow \infty\).
A small discount factor makes short-term rewards more favorable.
The algorithm fits a neural network taking the current state $\f{s}_t$ as input and giving an estimate of the maximum cumulative reward of type \eqref{CumulativeReward} achievable by subsequently taking each permitted action $a_t$.
The learning process is linked to the Bellman's equation update 
\begin{align}\label{BellmanUpdate}
[Q(\f{s}_t)]_{a_t} & = (1-\alpha)[Q(\f{s}_t)]_{a_t} \nonumber \\
& \quad + \alpha \left(r(\f{s}_t,a_t) + \gamma \max_{a_{t+1}}{[Q(\f{s}_{t+1})]_{a_{t+1}}}\right)
\end{align}
for a given learning rate $\alpha\in(0,1)$, and where $r(\f{s_t},a_t)$ is the reward for taking action $a_t$ in state $\f{s}_t$.

\subsubsection{Multi-Reward adaptation in the sense of \cite{friedman2018generalizing}}

With respect to the previous case, the neural network's input is augmented by a reward weight vector $\f{w}$, which is used to compute the total reward $\f{w}\cdot \f{r}(\f{s}_t,a_t)$ (here $\f{r}$ is the vector of rewards, and $\cdot$ denotes the standard scalar product).
The Single-Reward case can be seen as a declination of the Multi-Reward case with constant suitable one-hot encoding vectors $\f{w}$.
The \alg algorithm is summarized in Algorithm \ref{alg:HERRL} for the reader's convenience.

\subsection{Our abstract setup}\label{AbstractSetup}
The methodology we adopt in this paper is summarized in Algorithm \ref{alg:HERRL-DFG}, and stems from the underlying basic Algorithm \ref{alg:HERRL}.
For the sake of clarity, Algorithm \ref{alg:HERRL-DFG} highlights only the changes between the two algorithms.
These modifications are related to:
\begin{description}
\item i) an option for `random access point': subject to specification by the user, each training episode may use a subset of the full price history of the training set, where the starting point is randomly sampled and the length is fixed;
\item ii) the generalization of the discount factor $\gamma$ as suggested in \cite{friedman2018generalizing}: this means that the neural network's input also comprised the discount factor $\gamma$ (i.e., input is augmented from $(\f{s},\f{w})$ to $(\f{s},\f{w},\gamma)$);
\item iii) the specific choice of normalization spelled out in Subsection \ref{Normalization} below.
\end{description}

\subsubsection{Choice of Normalization}\label{Normalization}
We choose to indirectly normalize the neural network's output variables (i.e., the approximate $Q$-values) by rescaling the rewards in the Bellman's update \eqref{BellmanUpdate}.
More precisely, whenever a mini-batch 
$$
\{(\f{s}_i,\gamma_i,\f{w}_i,\f{w}_i\cdot \f{r}(\f{s}_i,a_i),a_i,\f{s}_{i, new})\}_{i\in B}, \qquad B\subset \{1,\dots,\#\mathcal{R}\}
$$ 
is randomly sampled from the replay $\mathcal{R}$ (see Algorithm \ref{alg:HERRL-DFG}, line \ref{sampleandnormalize}), the rescaled vectors  
\begin{align}\label{RewardRescaling}
\displaystyle
\tilde{\f{r}}(\f{s}_i,a_i) & = \frac{\Sigma^{-1/2} \f{r}(\f{s}_i,a_i)}{\|\f{w}_i\|_2},\qquad i\in B
\end{align}
and associated scalar rewards $\{\f{w}_i\cdot \tilde{\f{r}}(\f{s}_i,a_i)\}_{i\in B}$ are fed to the training in place of the original sets $\{\f{r}(\f{s}_i,a_i)\}_{i\in B}$ and $\{\f{w}_i\cdot \f{r}(\f{s}_i,a_i)\}_{i\in B}$.
Here, the matrix $\Sigma$ denotes the approximate covariance matrix computed using the reward vectors from the entire replay, namely 
$$
\Sigma = \mbox{Cov}\left(\{\f{r}(\f{s}_i,a_i)\}_{i\in \{1,\dots,\#\mathcal{R}\}}\right).
$$
With this choice, the overall scalar rewards $\{\f{w}_i\cdot \tilde{\f{r}}(\f{s}_i,a_i)\}_{i\in B}$ are normalized, in the sense that 
\begin{align}\label{CovarianceAfterNormalization}
\mbox{Cov}\left(\{\f{w}_i\cdot \tilde{\f{r}}(\f{s}_i,a_i)\}_{i\in B}\right) = 1.
\end{align}

\subsubsection{Length of the Experience Replay}

We exclusively use a \emph{same-age} type experience replay:
more precisely, the experience replay's oldest element (measured in number of network updates following its creation) is the same for both Single-- and Multi--Reward case. 
In the Multi-Reward case, where extra -- not visited -- experiences are thrown into the replay (Algorithm \ref{alg:HERRL}-line \ref{add-experiences}), this results in a longer replay.

\section{Specific details of our model}\label{Specifics}

After having laid out the general structure of our RL setup (see Algorithm \ref{alg:HERRL-DFG}), we give precise substance to all quantities involved.

\subsection{State variables $\f{s}_t$}

We define the state variables $\f{s}_t$ as the vector comprising both the current position in the market (which we name $p_t$, and whose precise details are given in Subsection \ref{admissableactionspositions} below) and a fixed lookback of length $\ell$ over the most recent log returns of close prices $\{z_t\}_t$: more explicitly, we set  
\begin{align}\label{StateVariables}
\f{s}_t := \left(\left\{\ln z_s - \ln z_{s-1}\right\}_{s = t-(\ell-1)}^{t},\,p_t\right) \in \mathbb{R}^{\ell}\times \mathbb{R}.
\end{align}

\subsection{Admissible Actions and Positions}\label{admissableactionspositions}
As far as actions are concerned, we analyze two scenarios:
\begin{itemize}
\item {\tt Long} Positions only (\LPcompact): The agent is only allowed to perform two actions ({\tt Buy}/{\tt Hold})\footnote{selling previously acquired assets}, and consequently only switch between trading positions {\tt Long}/{\tt Neutral}.
\item {\tt Long} and {\tt Short} Positions (\LSPcompact): the agent is allowed to perform three actions ({\tt Buy}/{\tt Sell}/{\tt Hold}), and consequently switch between trading positions {\tt Long}/{\tt Short}/{\tt Neutral}.
\end{itemize}

\subsection{Rewards and Profit}\label{RewardsAndProfit}
For a given single-asset dataset with close prices $\{z_t\}_t$, we define the logarithmic (portfolio) return at time $t$ as
\begin{align}\label{LogReturns}
\ell r_t :=
\left\{
\begin{array}{rl}
\ln z_t - \ln z_{t-1}, & \mbox{ if in {\tt Long} at time }t-1,  \\
-\left(\ln z_t - \ln z_{t-1}\right), & \mbox{ if in {\tt Short} at time }t-1, \\
0, & \mbox{ if in {\tt Neutral} at time }t-1.
\end{array}
\right.
\end{align}
Let $L\in\mathbb{N}$ be fixed. We focus on three well-established rewards (at a reference given point in time $t$), namely:
\begin{description}
\item i) the last logarithmic return (LR), which is exactly \eqref{LogReturns};
\item ii) the average logarithmic return (ALR), given by
\begin{align*}
\mbox{ALR} := \mbox{mean}\left[\left\{\ell r_s\right\}_{s=t-(L-1)}^{t}\right];
\end{align*}
\item ii) the non-annualized Sharpe Ratio (SR), given by
\begin{align*}
\mbox{SR} := \frac{\mbox{mean}\left[\left\{\ell r_s\right\}_{s=t-(L-1)}^{t}\right]}{\mbox{std}\left[\left\{\ell r_s\right\}_{s=t-(L-1)}^{t}\right]},
\end{align*}
\end{description}
as well as the sparse, less conventional reward:
\begin{description}
\item iv) a `profit-only-when-(position)-closed' (POWC) reward, defined as
\begin{align*}
\mbox{POWC} :=
\left\{
\begin{array}{rl}
\ln z_t - \ln z_{t_{LT}}, & \mbox{ if {\tt Long} closed at time }t-1,  \\
-\left(\ln z_t - \ln z_{t_{LT}}\right), & \mbox{ if {\tt Short} closed at time }t-1,\\
0, & \mbox{ otherwise},
\end{array}
\right.
\end{align*}
where $t_{LT}$ is the time of last trade (i.e., last position change).
\end{description}

\section{Experiments}\label{Experiments}

We substantiate all necessary components involved in the simulations of our model (given in Algorithm \ref{alg:HERRL-DFG}). 

\subsection{Codebase}

For the structure of our code, we took some inspiration from two open source repositories: the FOREX and stock environment at 
\begin{center}
\url{https://github.com/AminHP/gym-anytrading}
\end{center} 
and the minimal Deep Q-Learning implementation at 
\begin{center}
\url{https://github.com/mswang12/minDQN}\,.
\end{center}

\subsubsection{Open source directory and reproducibility}

Our entire code is provided in open source format at
\begin{center}
\url{https://github.com/trality/fire}\,.
\end{center}
In particular: the instructions for reproducibility are contained in the {\tt README.md} file therein; we provide the entirety of the datasets considered in our experiments.

\subsection{Datasets}

We perform several runs of experiments on a variety of relevant single-asset datasets, both in cryptocurrency and stock trading.  

In the interest of increasing the training capabilities of our experiments (see Subsection \ref{Predictability} below), we always include an evaluation set in addition to train and test sets. The percentages of the data associated with training/evaluation/test sets are roughly \emph{train}:64\%--\emph{eval}:16\%--\emph{test}:20\%.
All datasets are of sufficient length as to provide a reasonable compromise between experiment running times, and significance of predictions.
\subsubsection{Cryptocurrency pairs}

We consider the following datasets: \emph{hourly}-data points for the \btc{ }pair (August 2017--June 2020); 
\emph{hourly}-data points for the \ETH{ }pair (August 2017--June 2020);
\emph{hourly}-data points for the \XRP{ }pair (May 2018--March 2021).

\subsubsection{Stock indexes} We consider the indexes \AAPL, \SPY, \nifty. The date range for \AAPL, \SPY{ }is January $2000$--September $2022$ (\emph{daily}-data points), while for \nifty{ } is March $2020$--June $2020$ (\emph{minute}-data points).

\begin{figure}[h]
   \centering
\includegraphics[width=1\linewidth]{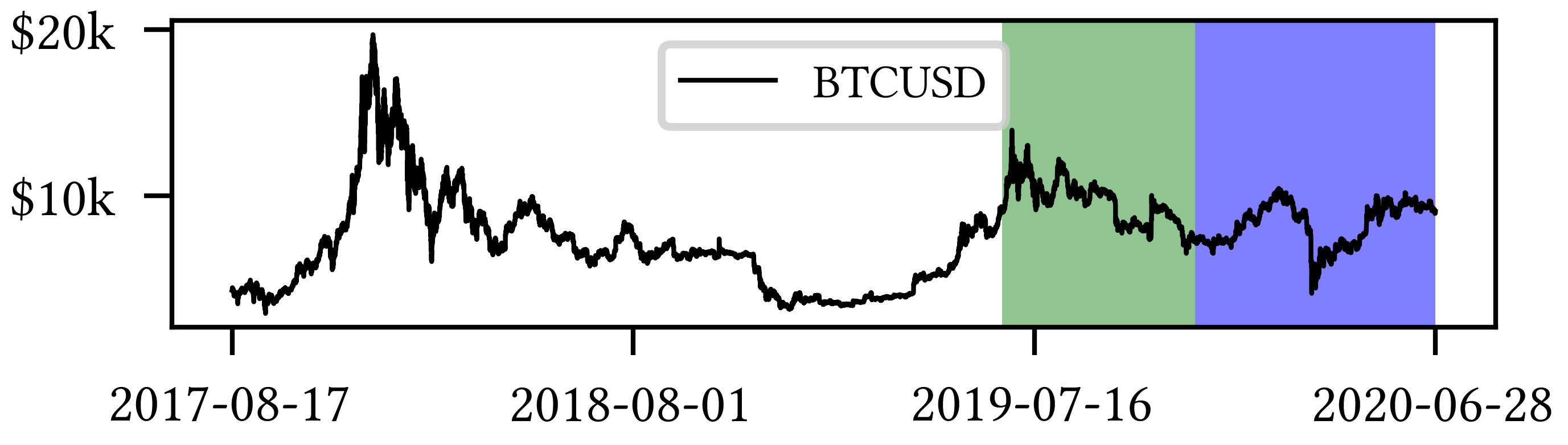}
\includegraphics[width=1\linewidth]{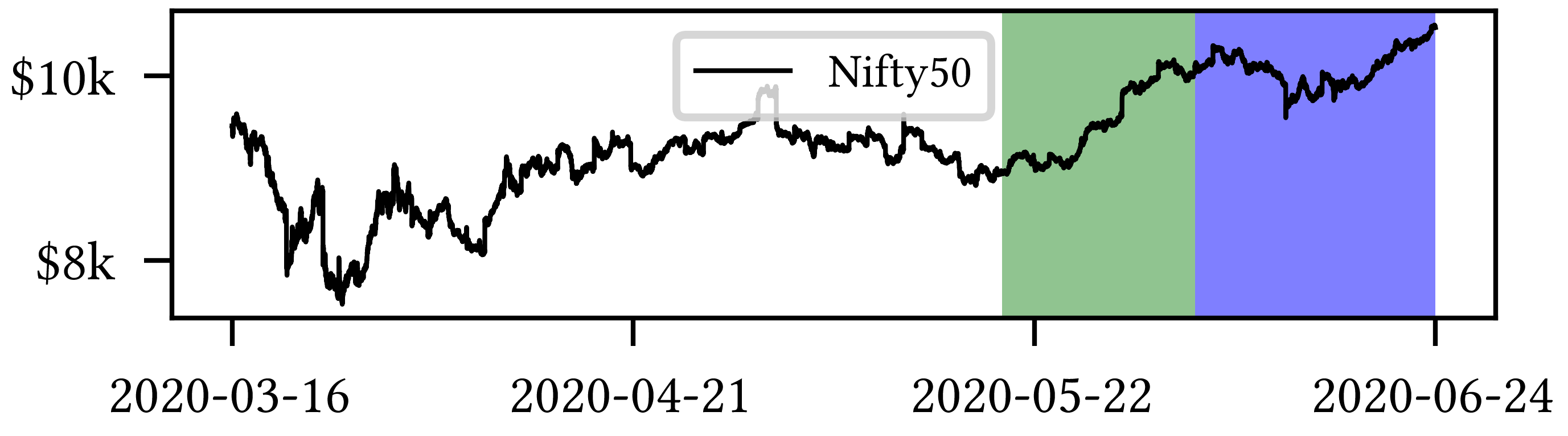}
   \caption{Close prices for train (white), evaluation (green), and test set (blue). Top (\emph{Bottom}): \btc{ }(\emph{\nifty}). We notice the mostly downward trend (respectively, \emph{strongly} upward trend) on the \btc{ }(\emph{\nifty}) evaluation set.}
  \Description{DataSetsCryptocurrency}
  \label{DataSetsCryptocurrency}
\end{figure}

A snapshot example from the datasets \btc{ } and \nifty{ }is shown in Figure \ref{DataSetsCryptocurrency}. 

\subsection{Quantities of interest and benchmarks}

All our considerations will be based on the following -- quite standard -- quantities:

\begin{itemize}
\item Total Reward: the cumulative reward over the considered portion of the dataset.
\item Total Profit: the cumulative gain/loss obtained by buying or selling with all the available capital at every trade.
\item Sharpe Ratio: the average return per step, divided by the standard deviation of all returns.
\end{itemize}
Crucially, results of Multi and Single-rewards simulations are compared against each other, as well as -- individually -- also against the Buy-and-Hold strategy.

\subsection{Measures for code efficiency}

\subsubsection{Basic measures} The most important measures taken in this regard are as follows.
Firstly, as we are primarily interested in assessing the potential superiority of a Multi-Reward approach over a Single-Reward one (see discussion in Subsection \ref{UnderlyingRL_AlgoSubsec}), we decide to stick to a simple Multi Layer Perceptron (MLP) Neural Network (Algorithm \ref{alg:HERRL}- line \ref{state-actionNN}). 
Secondly, for the purpose of checking the performance in between training, we run the currently available model on full training and evaluation sets only for an evenly distributed subspace of episodes. 

\subsubsection{Option of random access point}
If we choose the walk-forward cross-validation, the training in each episode is performed on a randomly selected, contiguous subset of the full training set with pre-specified length (this reduces the overall training cost).

\subsubsection{Vectorized computation of $Q$-values}
Except for the trading position $p_t$, the time evolution of the state variables vector given in \eqref{StateVariables} is otherwise entirely predictable (as prices $z_t$ obviously do not change in between training episodes).
This implies that, given a predetermined set of $n$ actions, the algorithm can efficiently vectorize the evaluation of the neural network for each separate trading position, and then deploy the results to speedily compute the associated $n$ future steps.
This method is feasible as the cardinality of admissible values of non-predictable state variables (i.e., the trading position) is low (three at most, in the \LSP case).

\section{Results}\label{Results}

The results arising from several Multi and Single-Reward experiments (using Algorithm \ref{alg:HERRL-DFG}) on several datasets (\btc, \ETH, \XRP, \AAPL, \SPY, \nifty) give us four general indications, which we discuss below in detail in as many dedicated subsections. 
We support our analysis using different types of plots, namely:
\begin{description}
\item i) distribution of performance (box-plots for various rewards over train/eval/test sets) obtained using several independent experiment realizations, see Figure \ref{distributions} as an example;
\item ii) cumulative rewards on train/eval sets (as functions of epochs), see Figure \ref{POWCRewardsLP} as an example, and;
\item iii) for each epoch, the Sharpe Ratio on eval set of the current best model, see Figure \ref{Predictability1} as an example.
\end{description}

\subsection{Multi-Reward generalization properties}\label{TrainingGen}

The first crucial conclusion that we can comfortably jump to is that the Multi-Reward strategy generalizes well over all different rewards, and this can be seen on pretty much all plots which compare Multi- and Single-Reward. 

\begin{figure}
    \includegraphics[width=0.49\linewidth]{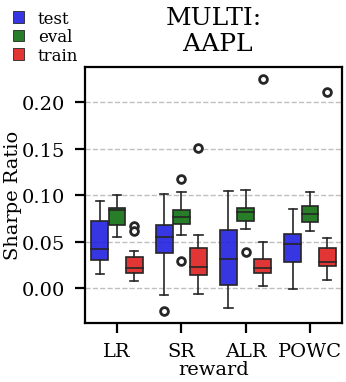}
    \includegraphics[width=0.49\linewidth]{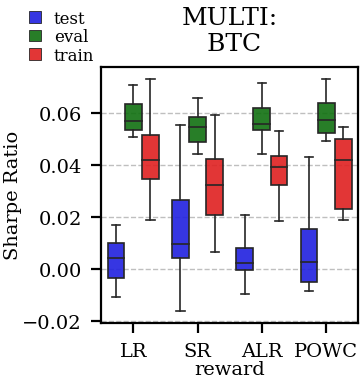}
    \includegraphics[width=0.49\linewidth]{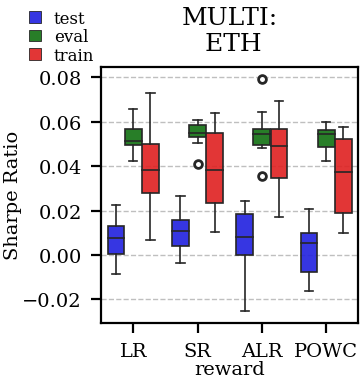}
    \includegraphics[width=0.49\linewidth]{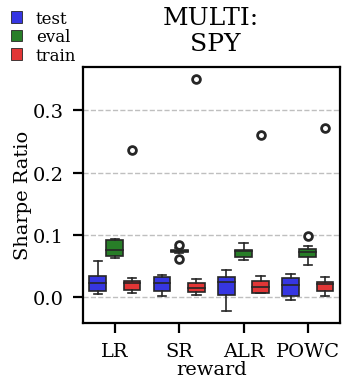}
    \caption{Distribution of the performance of multiple experiments (9, 13, 14, 17)
    with different random initialization for different assets on training,
    evaluation and test datasets, with multi reward.
        }\label{Fig_multi}
\end{figure}

\begin{figure}
    \includegraphics[width=0.49\linewidth]{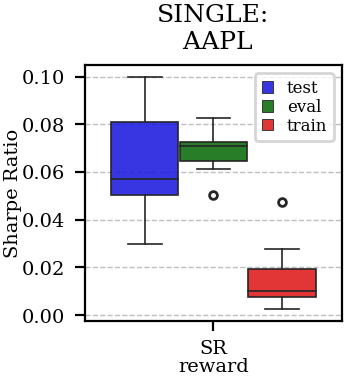}
    \includegraphics[width=0.49\linewidth]{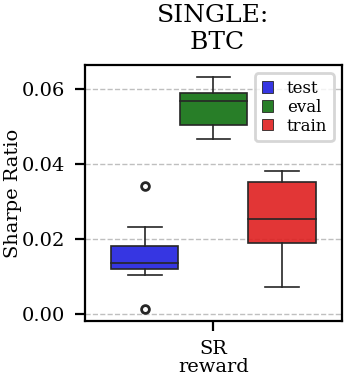}
    \includegraphics[width=0.49\linewidth]{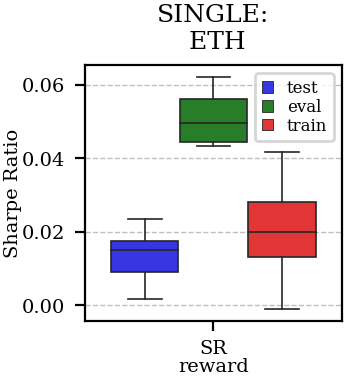}
    \includegraphics[width=0.49\linewidth]{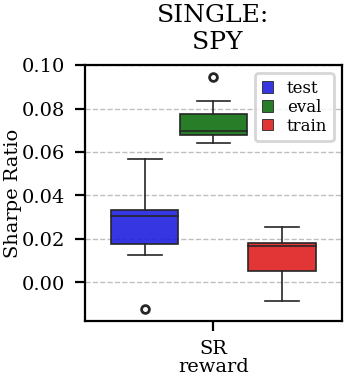}
    \caption{Distribution of the performance of multiple experiments (7, 8) with different
    random initialization for different assets on training, evaluation and test datasets,
    with single reward (SR). 
    }\label{single_various_SR}
\end{figure}

\begin{figure}
    \includegraphics[width=0.49\linewidth]{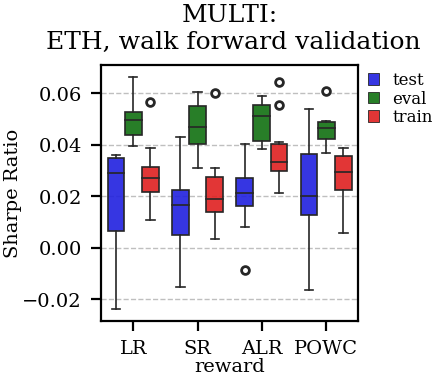}
    \includegraphics[width=0.49\linewidth]{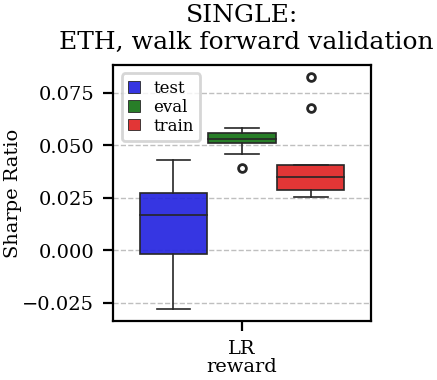}
    \includegraphics[width=0.49\linewidth]{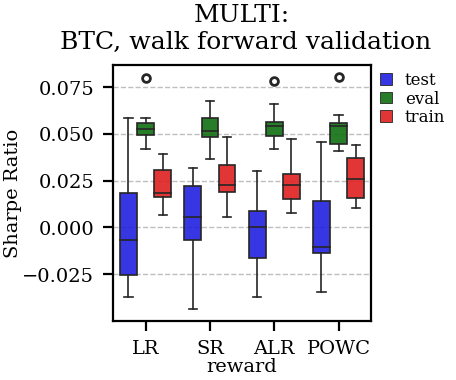}
    \includegraphics[width=0.49\linewidth]{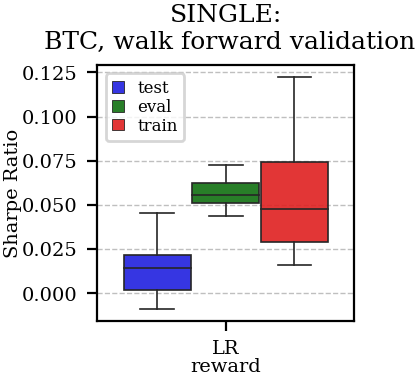}
    \caption{Distribution of the performance of multiple experiments (10) with different
    random initialization. Results are shown for \ETH{ }and \btc{ }on training, evaluation and test datasets.
    We compare multi and single reward using anchored walk-forward validation.
    }\label{walk_for_LR}
\end{figure}

Firstly, the generalization can be observed in average terms, for instance, in Figure \ref{distributions} (Single vs Multi for SR metric on \AAPL{ }and \btc), Figure \ref{walk_for_LR} (Single vs Multi for LR metric on \ETH{ }and \btc{ }with walk-forward validation) and from the comparison of Figures \ref{Fig_multi} and \ref{single_various_SR}.

\begin{figure}[H]
    \includegraphics[width=0.49\linewidth]{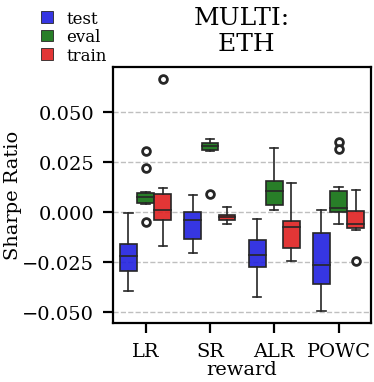}
    \includegraphics[width=0.49\linewidth]{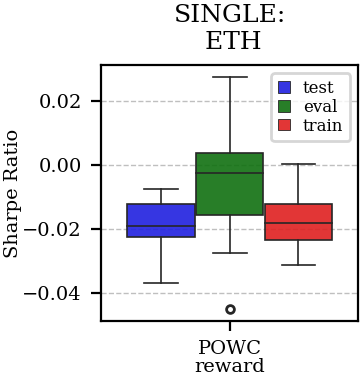}
    \caption{Distribution of the performance of multiple experiments (8, 10) with different
    random initialization for \ETH{ }on training, evaluation and test datasets,
    with multi and single (POWC) reward including fees of $0.03\%$.
    }\label{fig_fees}
\end{figure}

Secondly, the generalization is also visible for cumulative rewards in Figures \ref{pred1} \& \ref{pred2} (Single- vs Multi- for SR metric on \btc{ }and ALR metric on \nifty{}).
Thirdly, as far as the predictive power is concerned, the Multi-Reward method is as performing as -- and sometimes better performing than -- the Single-Reward counterpart, see Figures \ref{Predictability1}--\ref{Predictability2}--\ref{Predictability3}.
\begin{remark}
The training saturation levels may differ from those of the corresponding Single-Reward simulations, although this is likely caused by an apparent regularization effect of the Multi-Reward setting.
\end{remark}
\begin{remark}
The performance in the case of non-null fees (see Figure \ref{fig_fees}) is poor: this is not related to the generalization (which seems to hold also in this case), but rather to the simple nature of the underlying RL algorithm, see discussion in Subsection \ref{UnderlyingRL_AlgoSubsec}.
\end{remark} 

\subsection{Multi-Reward improvement on strongly position-dependent rewards}\label{SufferingPOWC} 
 
Let us consider a trading reward which 
\begin{description}
\item i) is strongly dependent on a specific trading position, and;
\item ii) is \emph{sparse} (meaning that it might take several time steps for such a reward to return a non-zero value).
\end{description}
Intuition says that it is highly likely that the Single-Reward RL algorithm will struggle to learn based on such a reward. 
On the other hand, it is expected that a Multi-Reward algorithm will perform better, due to the influence of easier rewards with different but similar goals. 
Furthermore, the performance difference between Multi-Objective and Single-Objective method is expected to be even more pronounced when there are fewer trading positions allowed (thus further restricting the Single-case capabilities to learn). 

Below, we confirm these intuitions for the POWC reward -- which satisfies i) and ii) above -- by running the Multi-Reward RL code with all four rewards considered in Subsection \ref{RewardsAndProfit} in its dictionary.

\subsubsection{Case \LP} 

\begin{figure}[h]
   \centering
\includegraphics[]{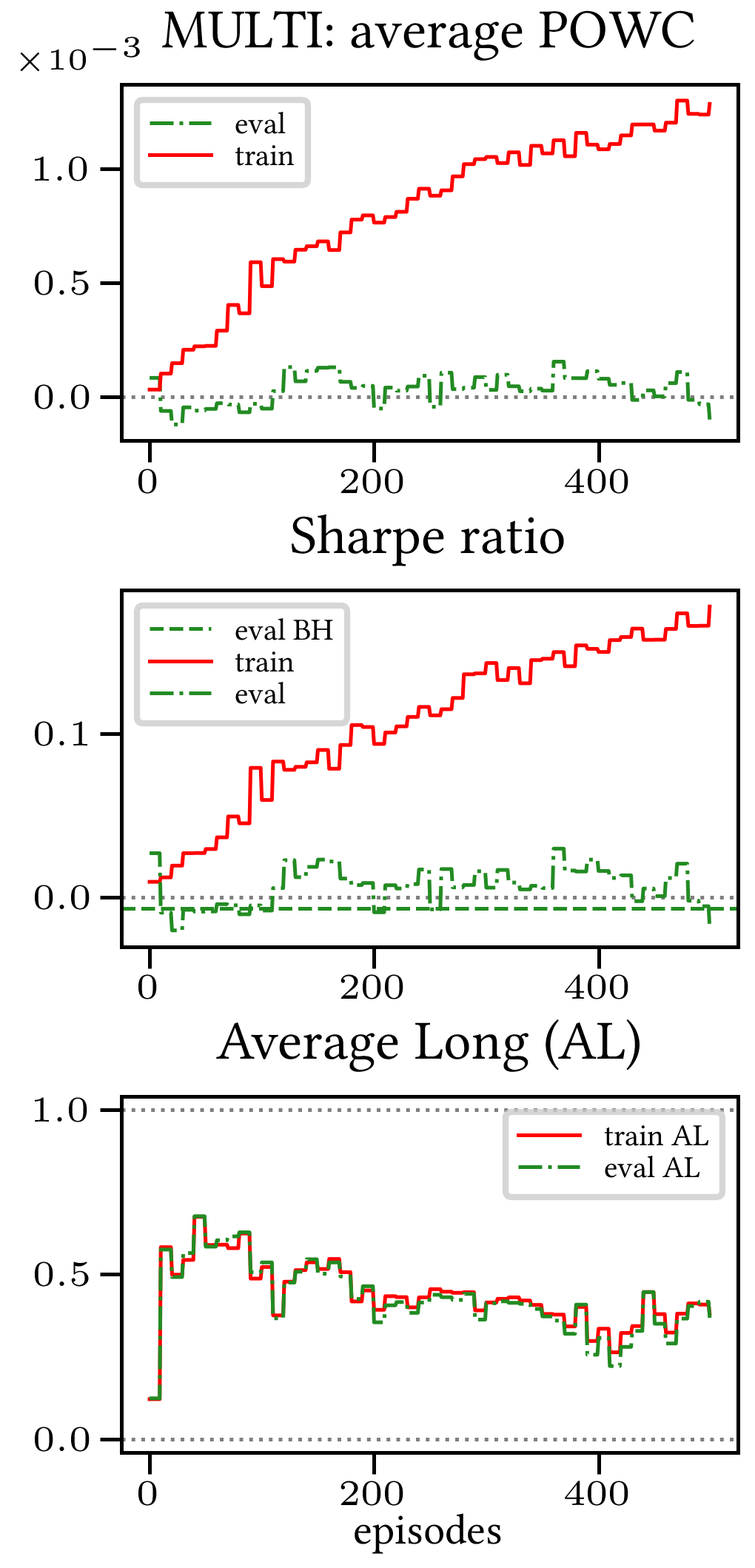}\includegraphics[]{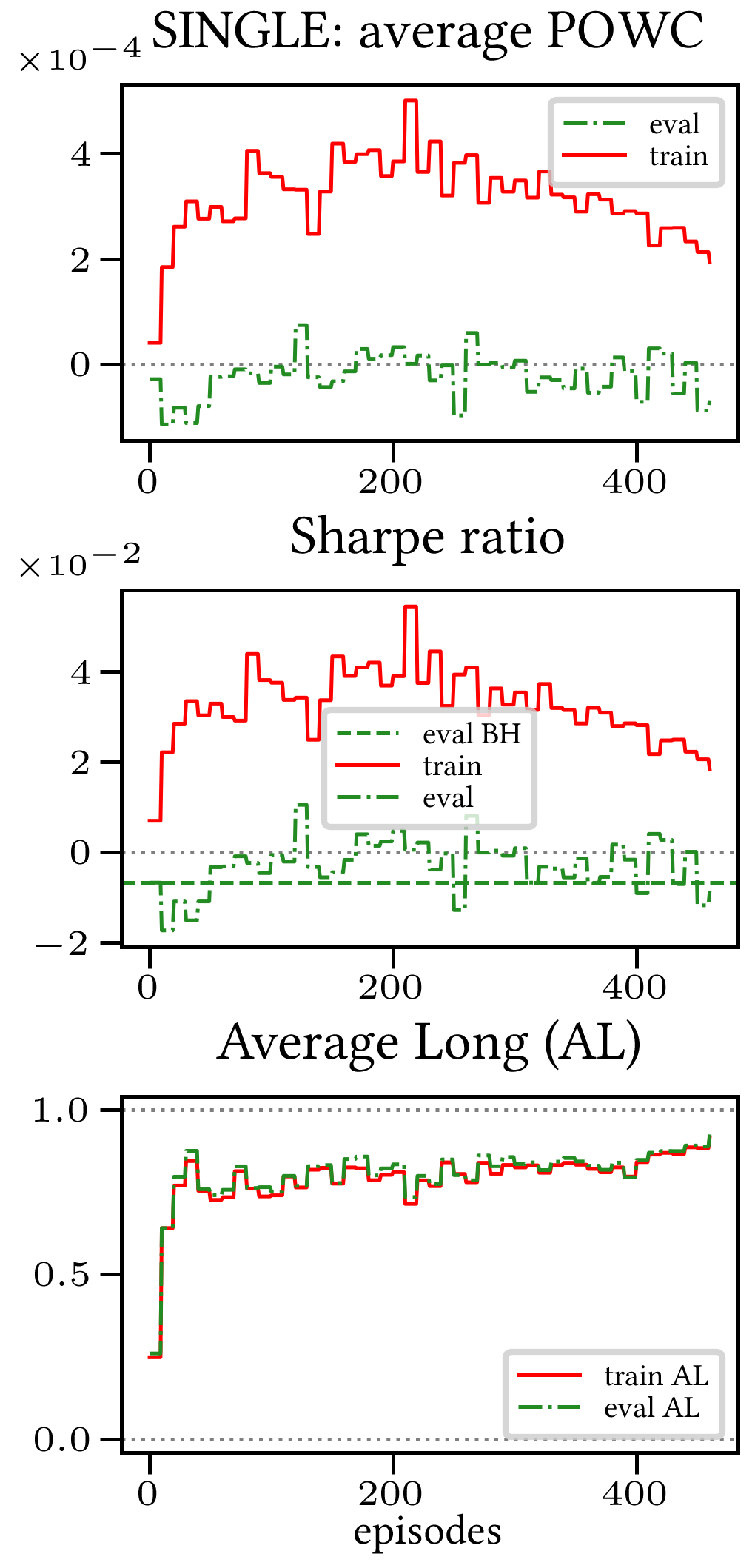}
  \caption{Reward performances, POWC/\LPcompact/\btctwenty}
  \Description{POWCRewardsLP}
  \label{POWCRewardsLP}
\end{figure}

\begin{figure}[h]
   \centering
\includegraphics[]{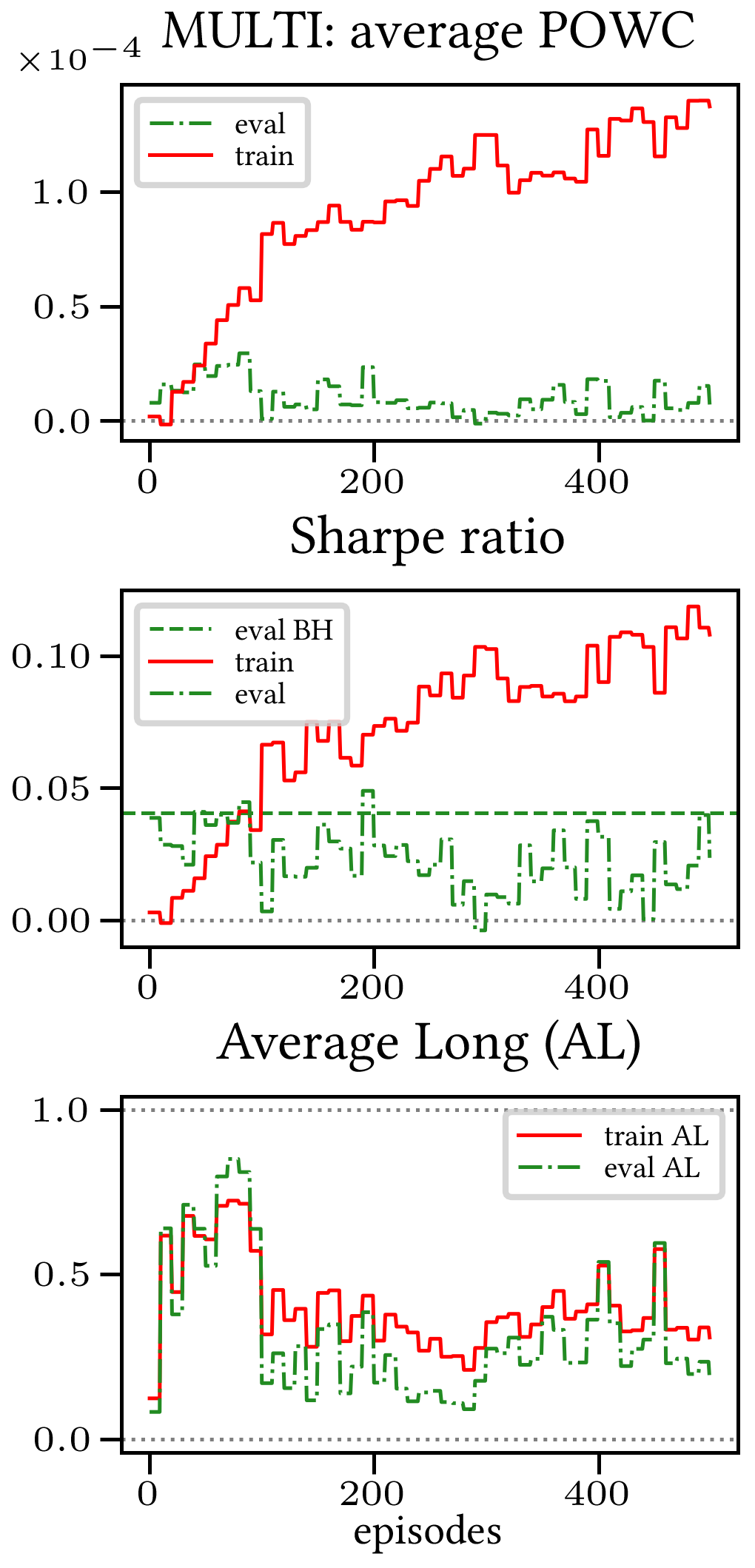}\includegraphics[]{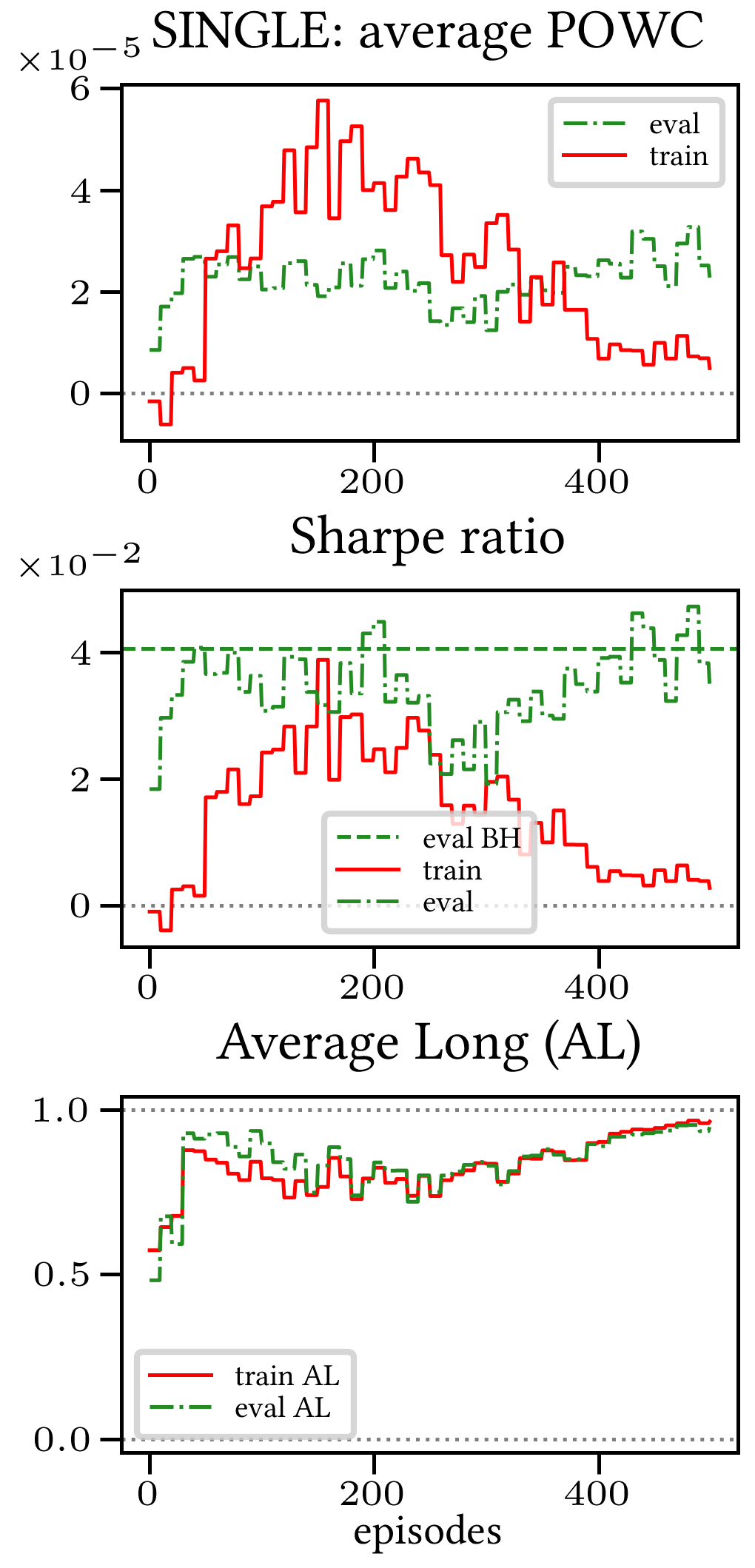}
  \caption{Reward performances, POWC/\LPcompact/\niftytwenty}
  \Description{POWCRewardsLPNifty}
  \label{POWCRewardsLPNifty}
\end{figure}

\begin{figure}[h]
   \centering
\includegraphics[width=0.49\linewidth]{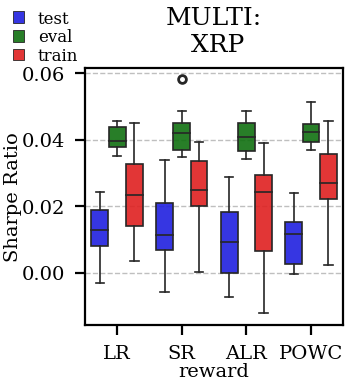}\includegraphics[width=0.49\linewidth]{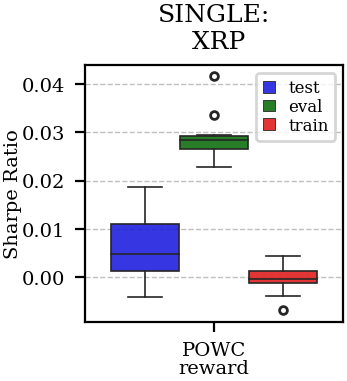}
  \caption{\LPcompact{ }case: Distribution of the performance of multiple experiments with different
    random initialization for \XRP{ }and POWC on training, evaluation and test datasets,
    with multi and single reward. }
  \Description{POWCRewardsXRPBoxplots}
  \label{POWCRewardsXRPBoxplots}
\end{figure}

\begin{figure}[h]
   \centering
\includegraphics[width=0.49\linewidth]{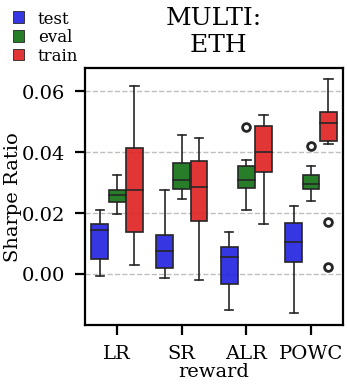}\includegraphics[width=0.49\linewidth]{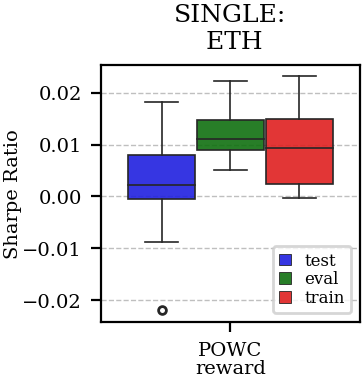}
  \caption{\LPcompact{ }case: Distribution of the performance of multiple experiments with different
    random initialization for \ETH{ }and POWC on training, evaluation and test datasets,
    with multi and single reward. }
  \Description{POWCRewardsETHBoxplots}
  \label{POWCRewardsETHBoxplots}
\end{figure}

When opening {\tt Short} positions is not allowed, the POWC reward provides a non-zero feedback only when {\tt Long} positions are closed. 
This extremely sparse feedback is likely to be the  justification of the poor cumulative training performance (Figures \ref{POWCRewardsLP} and \ref{POWCRewardsLPNifty}) where Single-Reward saturates the training at a much lower level than Multi-Reward. 
In contrast, the training for the Multi-case algorithm is much more consistent, as it can benefit from rewards with similar goals as POWC, but with more frequent feedback (e.g., LR).
In the Multi-Reward case, the prediction performance exceeds the Buy-and-Hold threshold on the \btc{ }dataset in a more consistent and stable way than in the training saturation regime of the Single-Reward case. 
Furthermore, the average of {\tt Long} positions\footnote{The average of {\tt Long} positions shows the time of capital exposure to the market.} is lower (which means less risk is taken) and are also relatively stable. The difference in strategy between Multi and Single-Reward algorithm can be inferred by the different convergence of {\tt Long} Positions. 
The analysis is further consolidated by considering the stark contrast in saturation levels in the distributional plots for the \XRP{ }and \ETH, see Figures \ref{POWCRewardsXRPBoxplots}--\ref{POWCRewardsETHBoxplots}.

\subsubsection{Case \LSP}

\begin{figure}
    \includegraphics[width=0.49\linewidth]{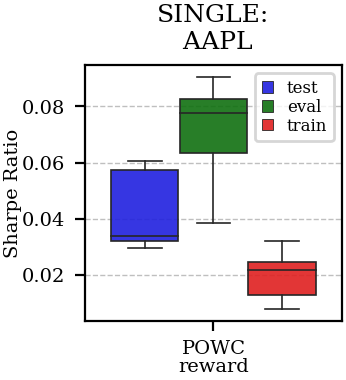}
    \includegraphics[width=0.49\linewidth]{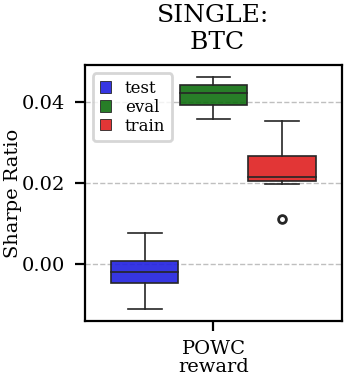}
    \includegraphics[width=0.49\linewidth]{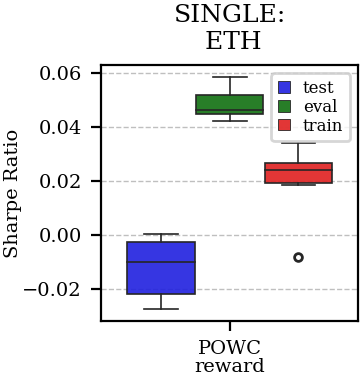}
    \includegraphics[width=0.49\linewidth]{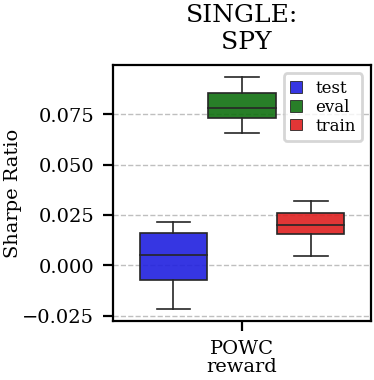}
    \caption{Distribution of the performance of multiple experiments (7) with different
    random initialization for different assets on training, evaluation and test datasets,
    with single reward (POWC). 
    }\label{single_various_POWC}
\end{figure}

\begin{figure}[h]
   \centering
\includegraphics[]{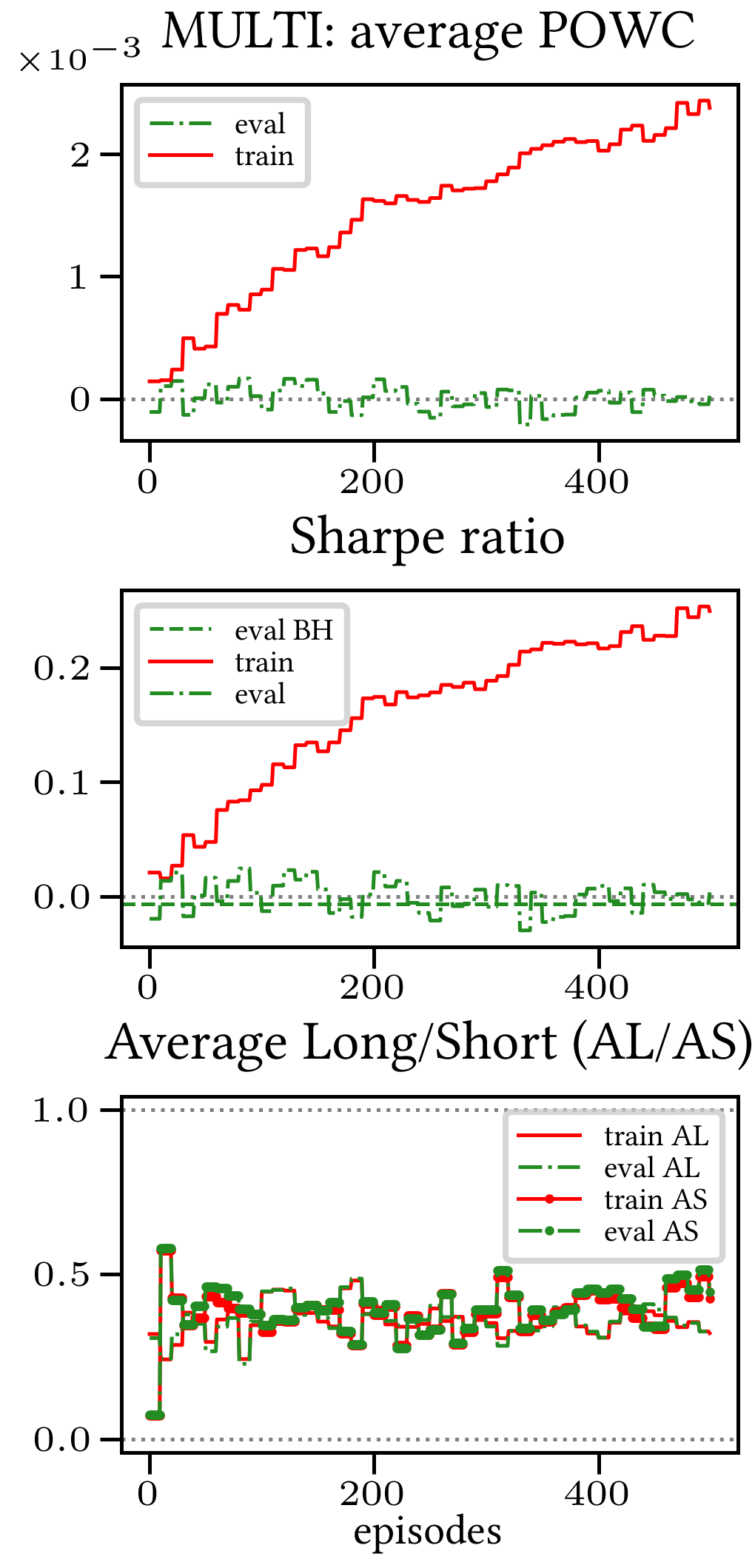}\includegraphics[]{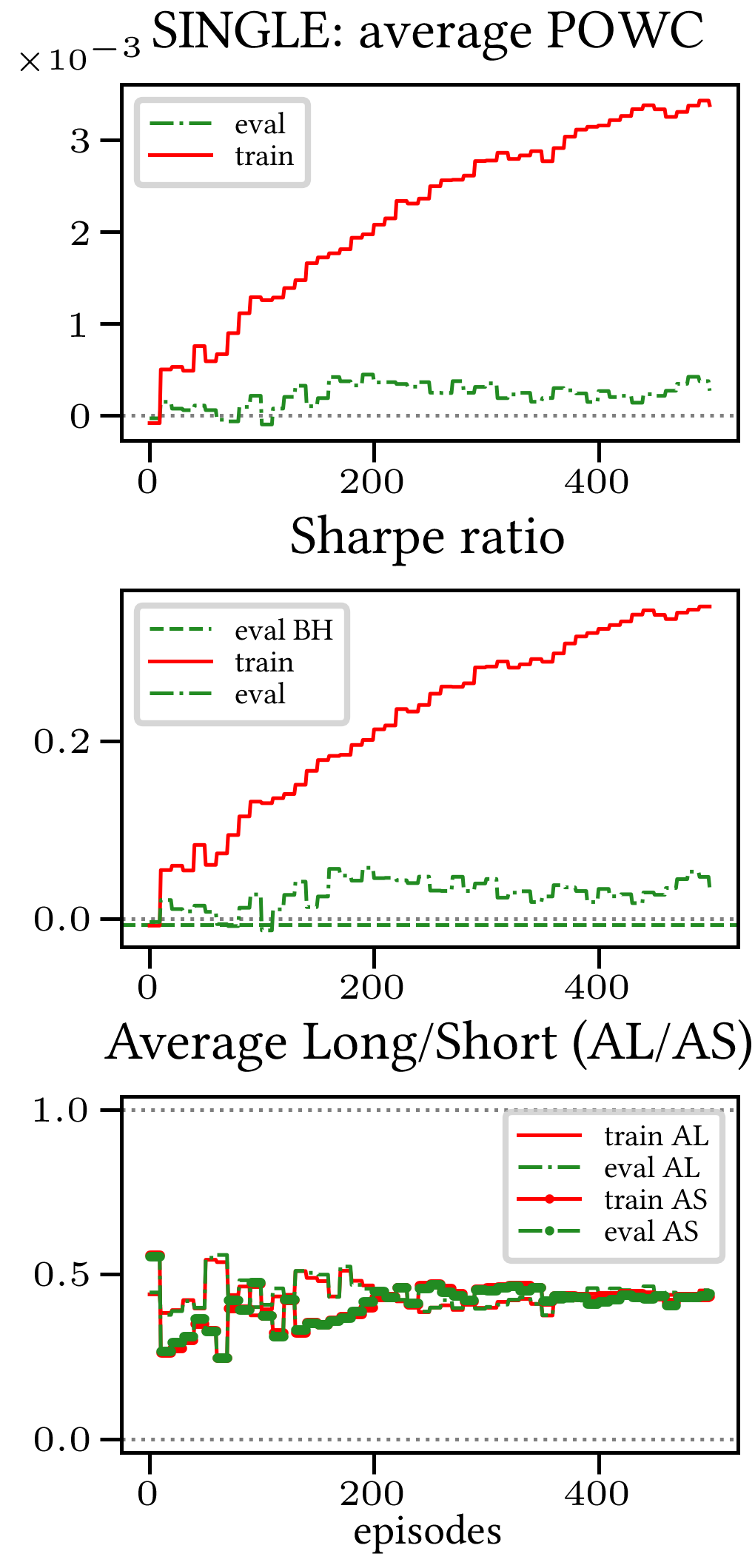}
  \caption{Reward performances, POWC/\LSPcompact/\btctwenty}
  \Description{POWCRewardsLSP}
  \label{POWCRewardsLSP}
\end{figure}

\begin{figure}[h]
   \centering
\includegraphics[]{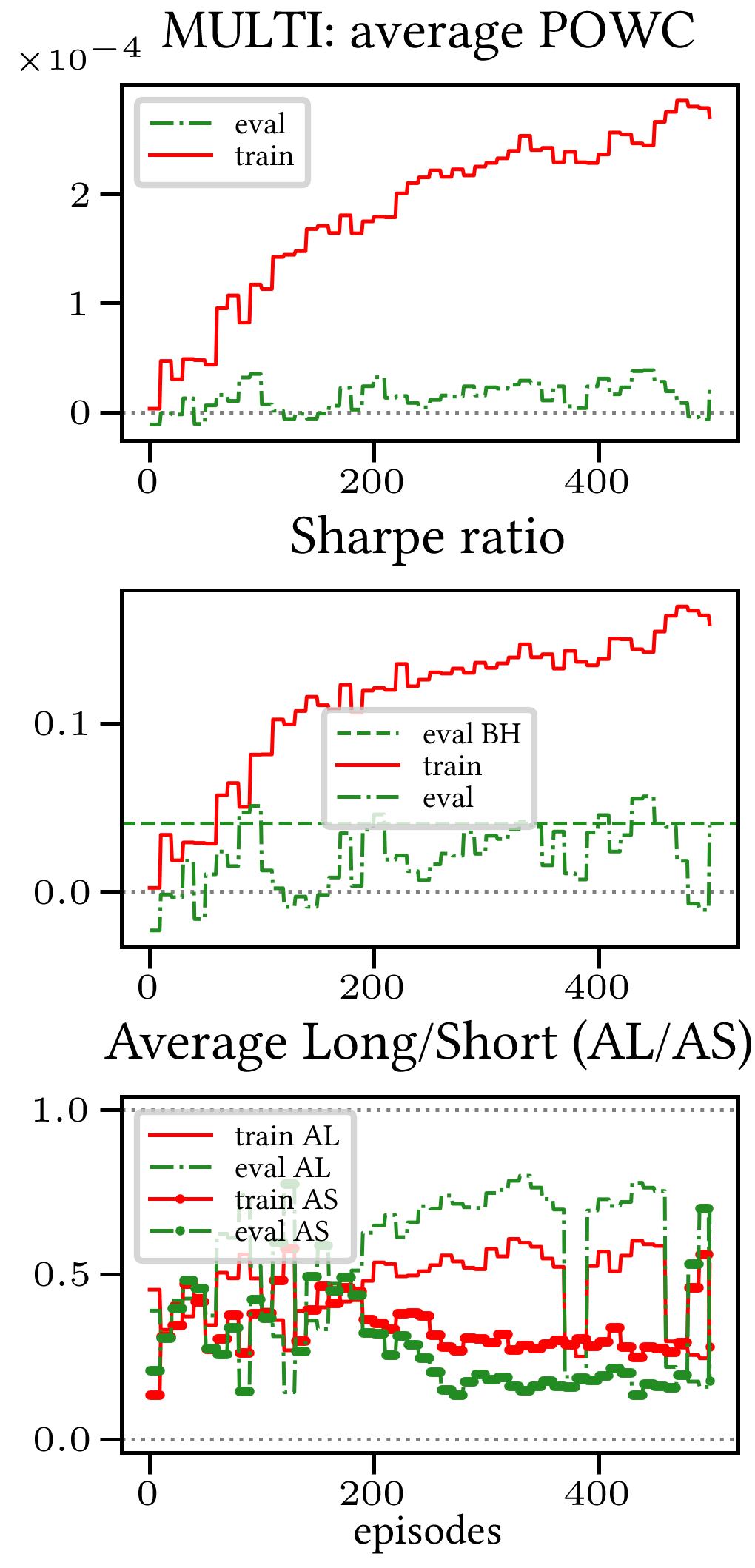}\includegraphics[]{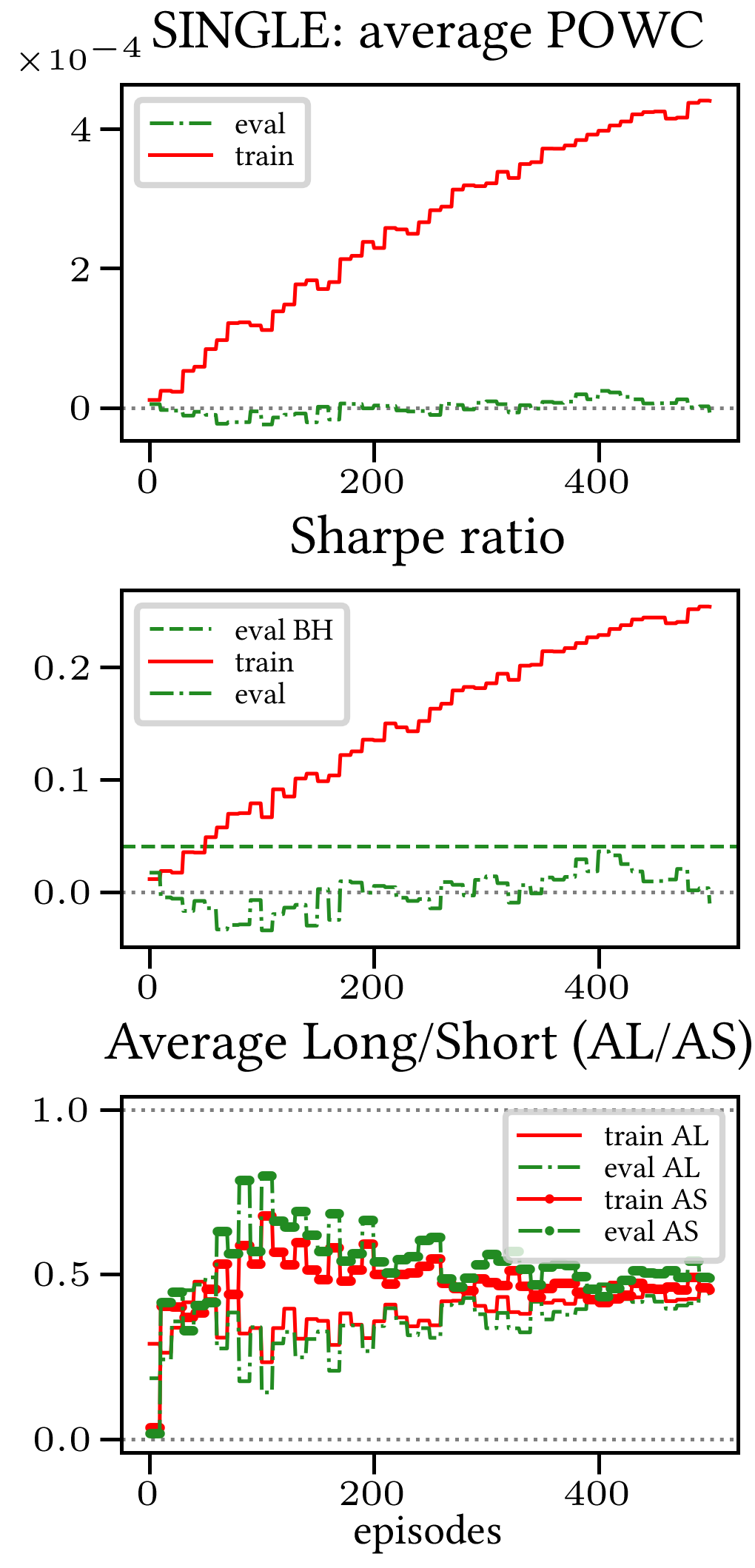}
  \caption{Reward performances, POWC/\LSPcompact/\niftytwenty}
  \Description{POWCRewardsLSPNifty}
  \label{POWCRewardsLSPNifty}
\end{figure}

As expected, the difference between Multi- and Single- case is much narrower in this case, as {\tt Short} positions can now be opened, see Figures \ref{POWCRewardsLSP} and \ref{POWCRewardsLSPNifty}. Except for the training saturation, relevant differences are not noted. This is also visible in distributional plots (for instance, consider the \SPY{ }pair in Figures \ref{Fig_multi} and \ref{single_various_POWC}).

 \subsection{Random discount factor generalization}\label{DiscountGen}

\begin{figure}[h]
   \centering
\includegraphics[]{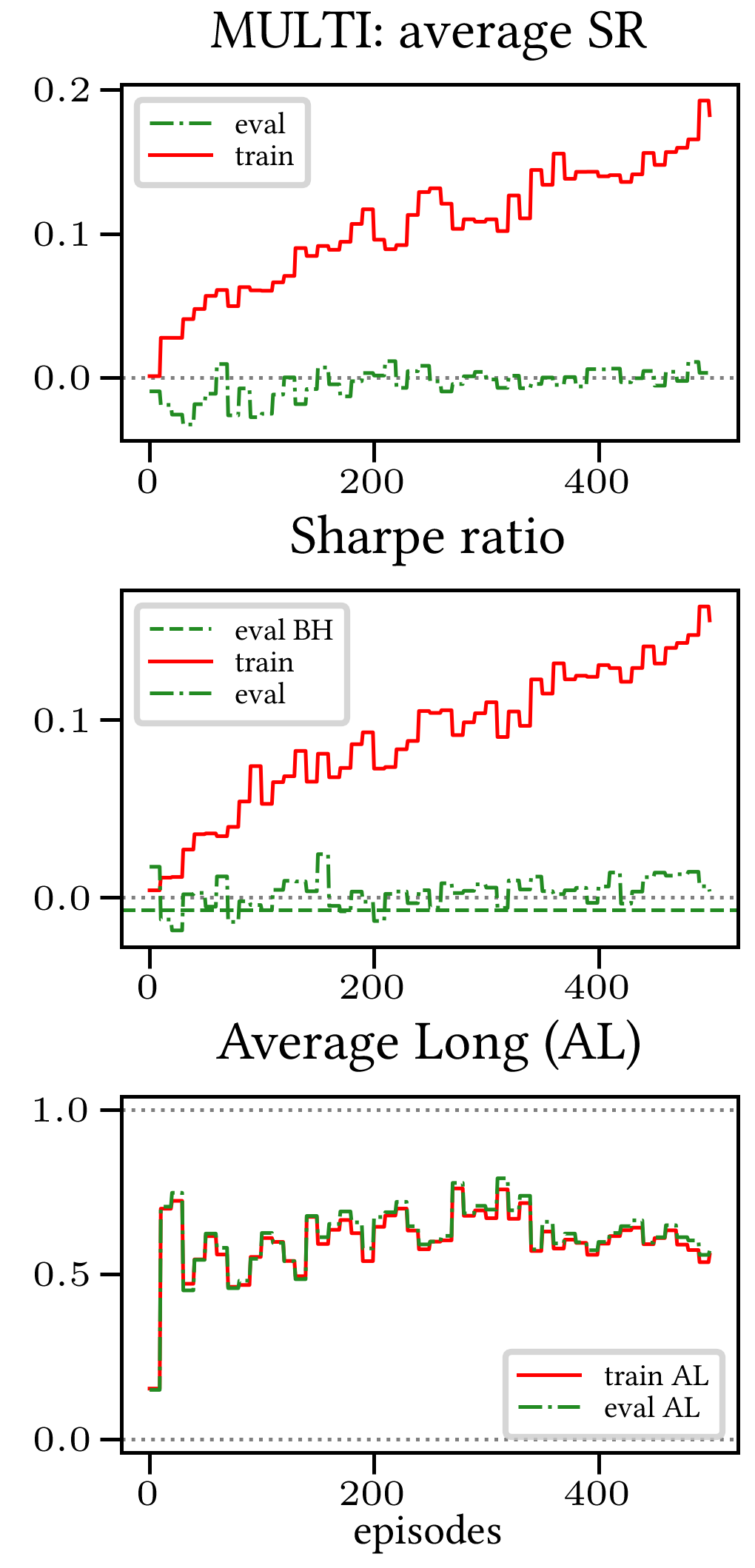}\includegraphics[]{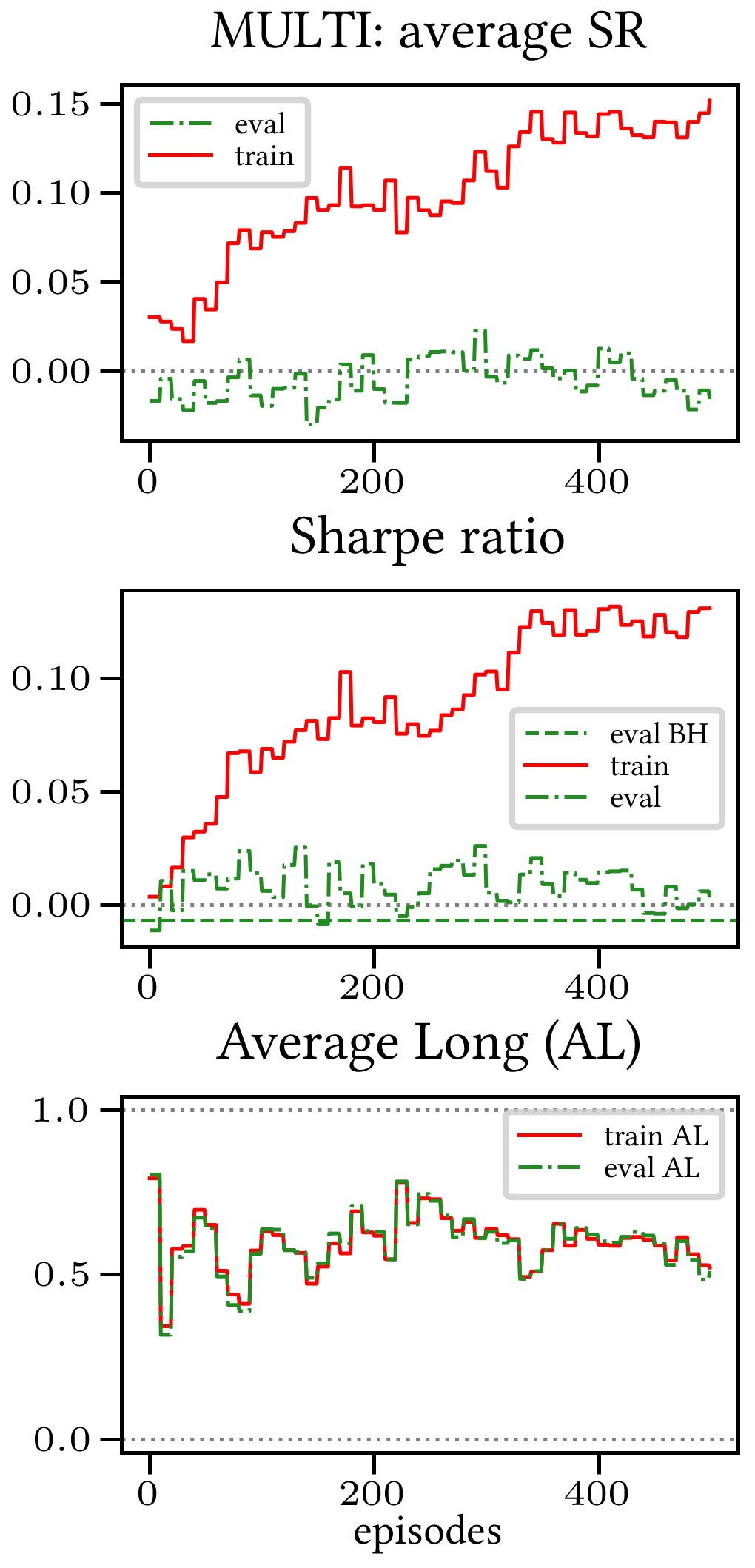}
  \caption{Reward Performances, SR/\LPcompact/\btctwenty. Left (\emph{Right}): Non-random (\emph{Random}) discount factor}
  \Description{rgammaRewardsLP-SR}
  \label{rgammaRewardsLP-SR}
\end{figure}

\begin{figure}[h]
   \centering
\includegraphics[]{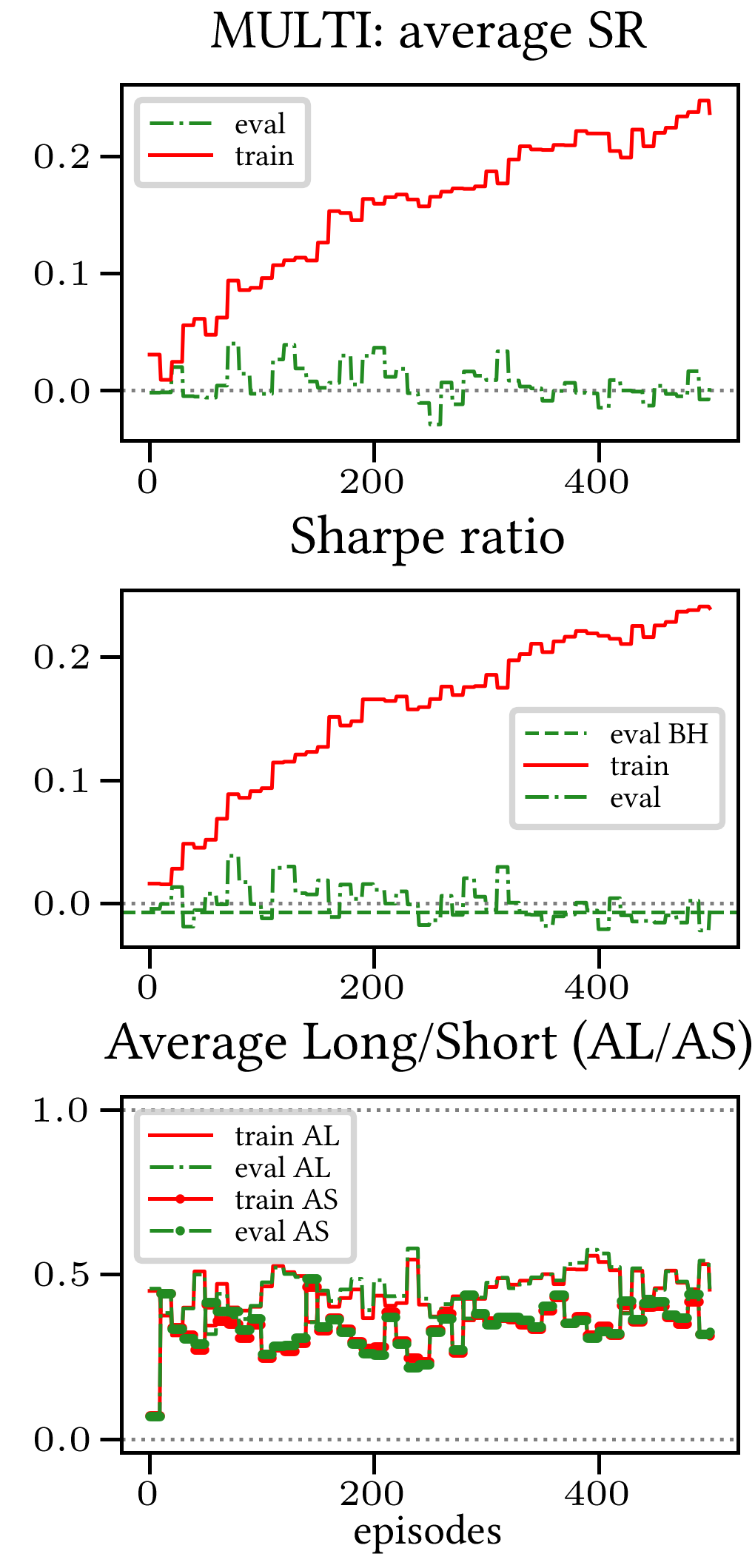}\includegraphics[]{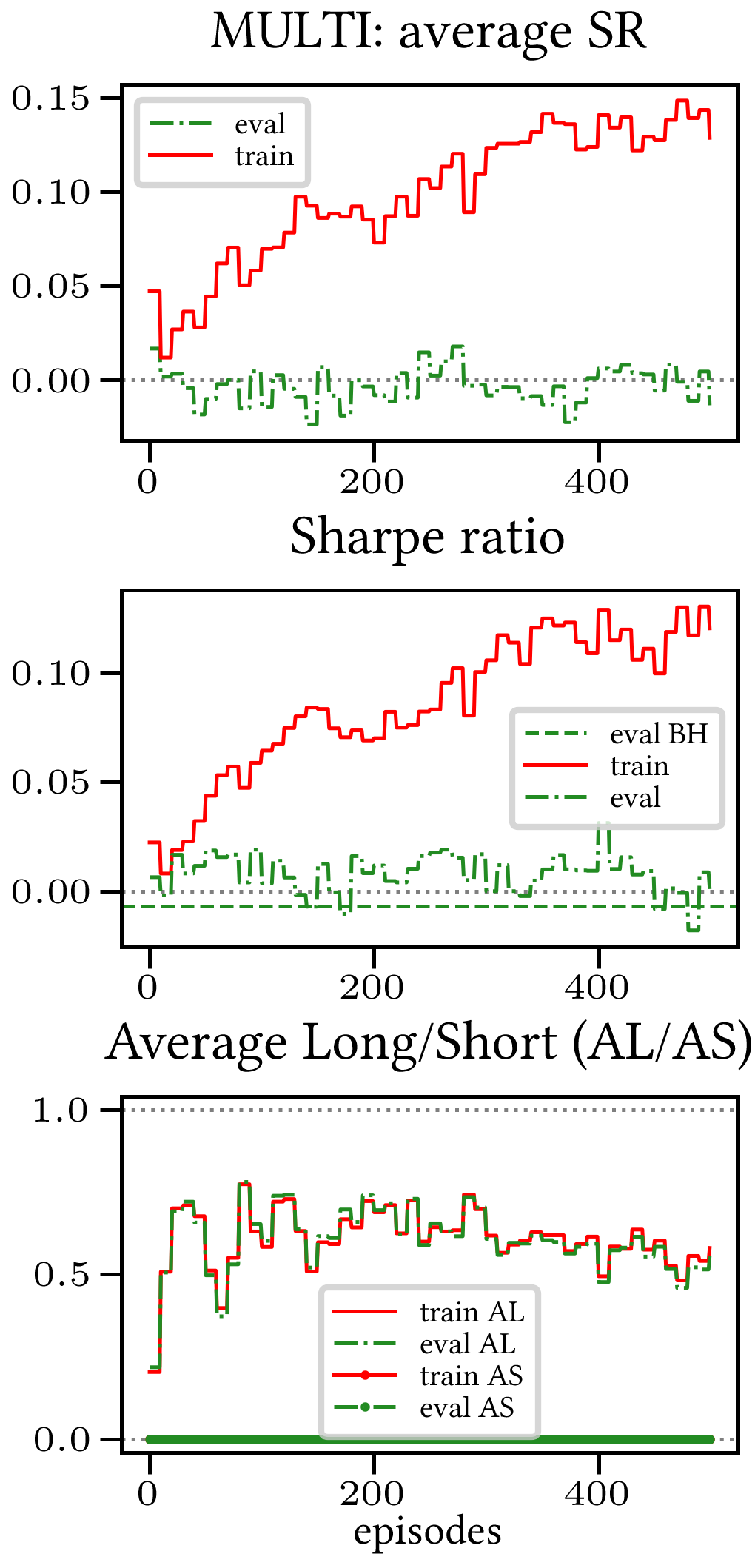}
  \caption{Reward Performances, SR/\LSPcompact/\btctwenty. Left (\emph{Right}): Non-random (\emph{Random}) discount factor}
  \Description{rgammaRewardsLSP-SR}
  \label{rgammaRewardsLSP-SR}
\end{figure}

We have run several simulations with both fixed and randomly sampled discount factor (as suggested in \cite{friedman2018generalizing}) on all datasets. 
The results are consistent across all simulations: therefore, we only show the results for the \btc{ }dataset in Figures \ref{rgammaRewardsLP-SR} and \ref{rgammaRewardsLSP-SR} (Multi-Reward case only, for reward SR in both the \LP and \LSP cases), as they are representative of all the remaining simulations.

We were able to notice the following general trends:

\begin{description}
\item - \emph{Generalization}. A graphical comparison of performance indicators suggests that the algorithm generalizes with respect to the value of the random discount factor.
\item - \emph{Training saturation levels}. 
There is a visible difference of saturation levels, with the random discount factor version saturating at a consistently lower level than its non-random discount factor counterpart.
It is plausible that the discount factor generalization serves the purpose of a neural network regularizer. The difference is more pronounced in the \LSP case, i.e., when the agent is allowed to open {\tt Short} positions.
\item - \emph{Evaluation set saturation levels and average positions}.  No significant differences are noticeable between the two cases.
\end{description} 

Taking everything into account, our simulations point in the direction of validating the discount factor generalization assumption provided in \cite{friedman2018generalizing}. 
Nevertheless, more extensive testing is necessary in order to fully confirm this.

\subsection{Consistent indications for predictability}\label{Predictability}

Although a full statistical justification of the obtained results is beyond the
scope of this paper, we nonetheless have achieved indications that validate the
effectiveness and robustness of the Multi-Reward approach. We further detail this statement.

\subsubsection{Consistent performance with respect to the Buy-and-Hold strategy}
\begin{figure}[h]
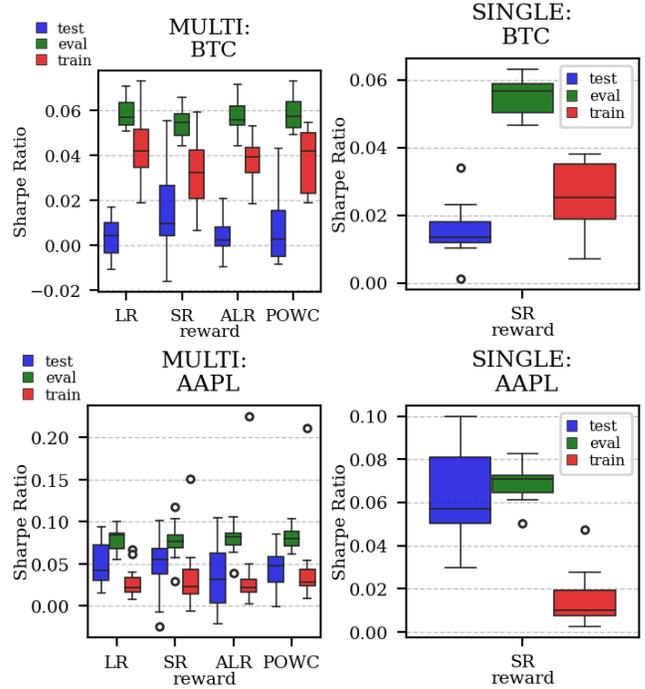

    \includegraphics[width=0.49\linewidth]{figures_paper/distributions/m_btc_13.png}
    \includegraphics[width=0.49\linewidth]{figures_paper/distributions/s_SR_btc_8.png}
    \includegraphics[width=0.49\linewidth]{figures_paper/distributions/m_aapl_17.png}
    \includegraphics[width=0.49\linewidth]{figures_paper/distributions/s_SR_aapl_7.png}
    \caption{Distribution of the performance of multiple experiments with different
    random initialization for \btc{} and AAPL on training, evaluation and test datasets,
    with multi and single reward. 
    }\label{distributions}
\end{figure}

In the majority of simulations, the  Multi-Reward is capable of
improving over the Buy-and-Hold strategy benchmark (in terms of Sharpe ratio),
on the evaluation and, more importantly, on the test set (Figure~\ref{distributions}).
Our validation has primarily relied on a distributional analysis across several experiments with independent initialization (Figures \ref{Fig_multi}, \ref{single_various_SR}, \ref{walk_for_LR}, \ref{single_various_POWC}, \ref{distributions}).

\subsubsection{Comparing performance on training, evaluation, test sets} 
\begin{figure}[h]
\includegraphics[width=0.49\linewidth]{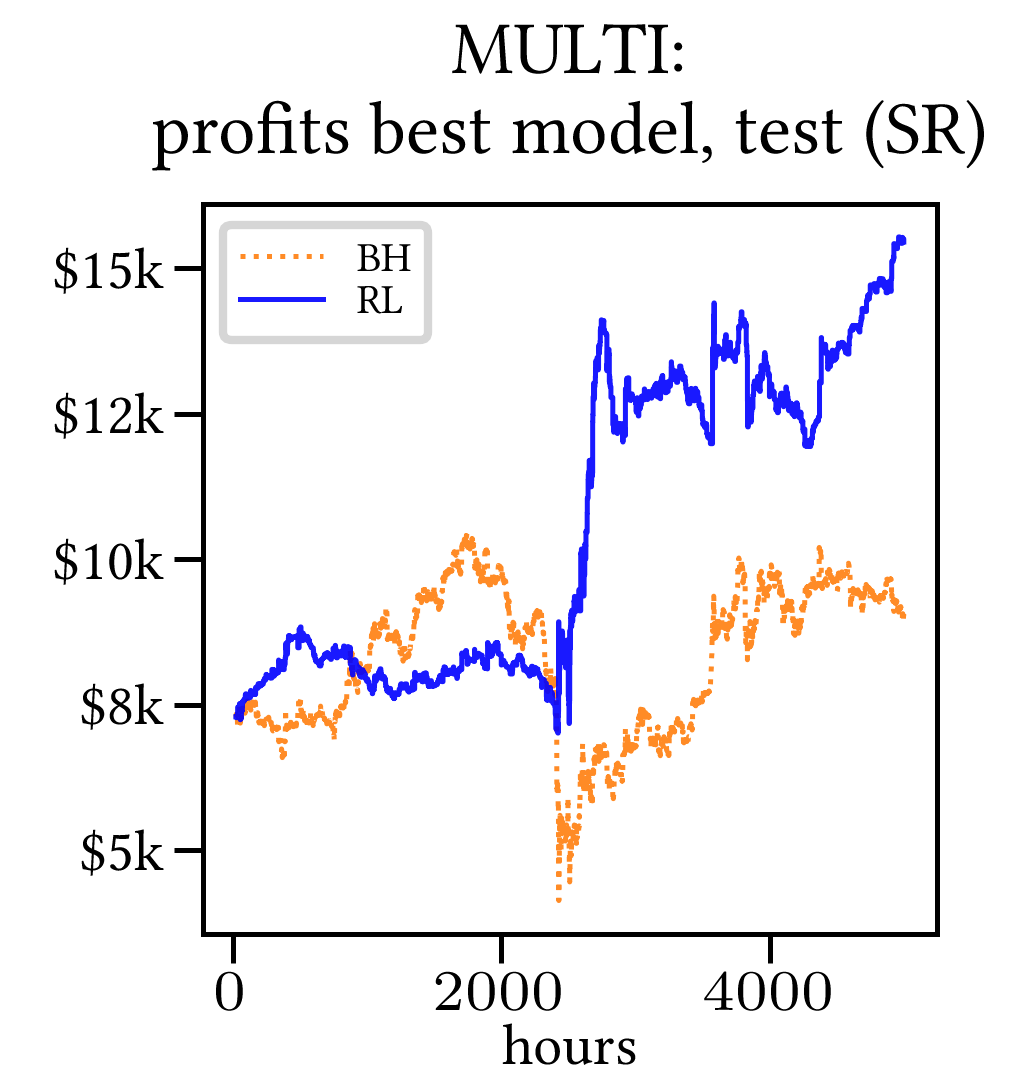}\includegraphics[width=0.49\linewidth]{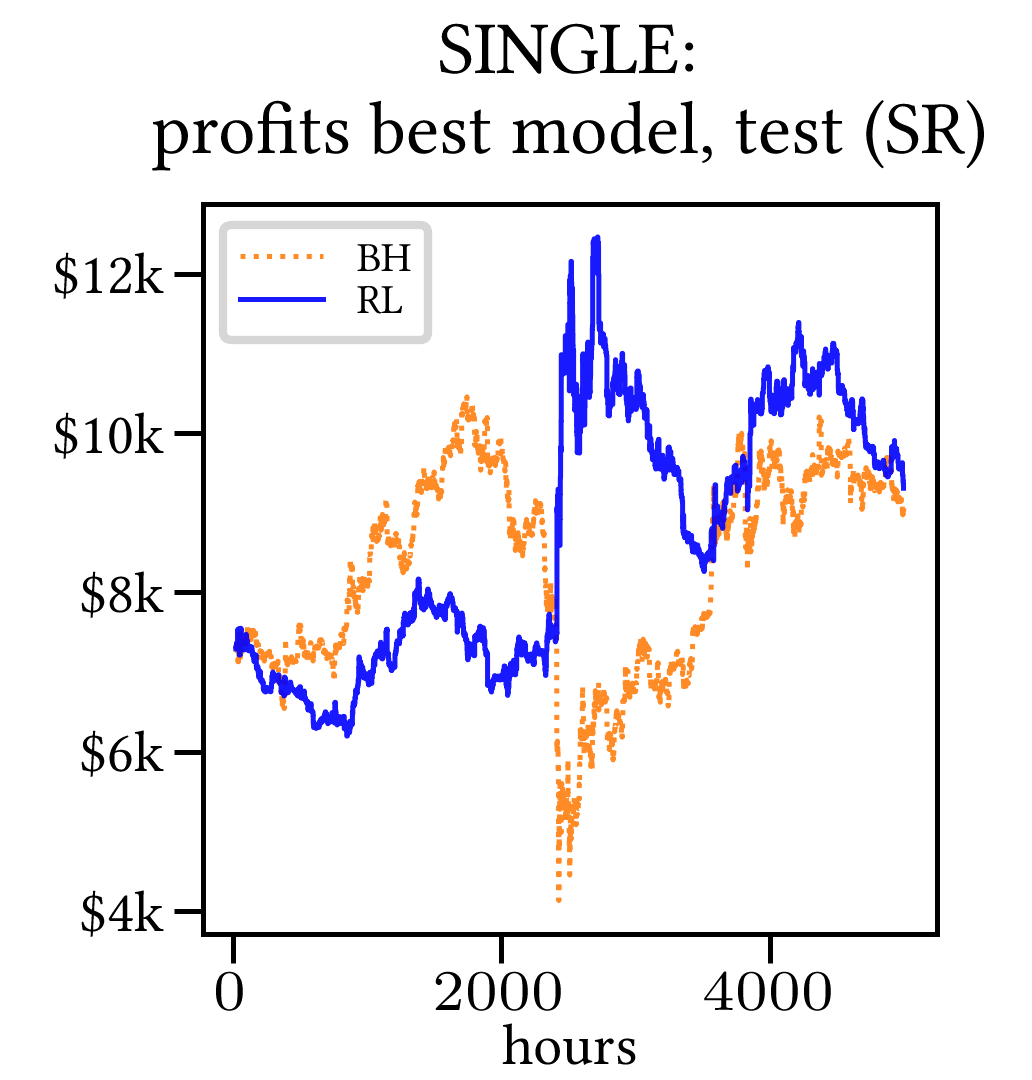}

\includegraphics[width=0.49\linewidth]{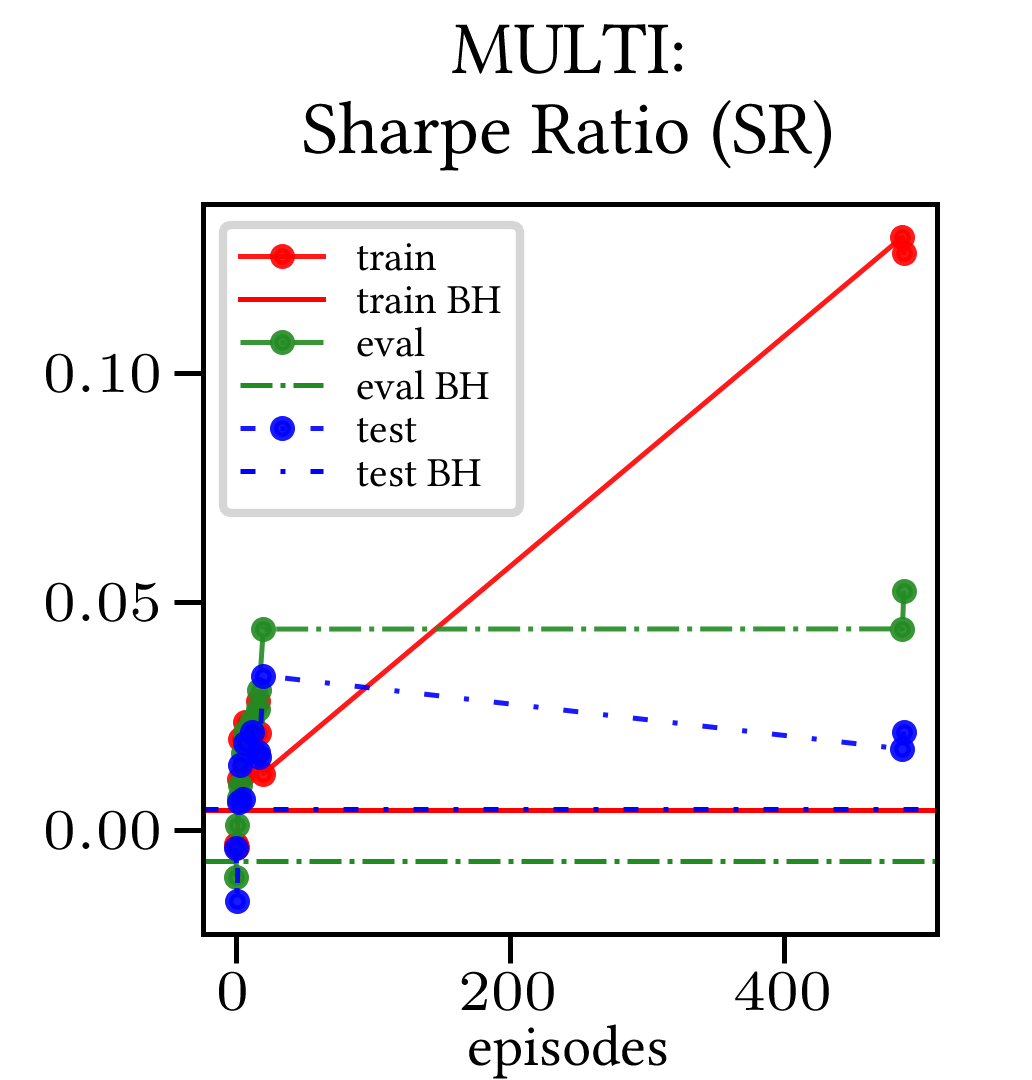}\includegraphics[width=0.49\linewidth]{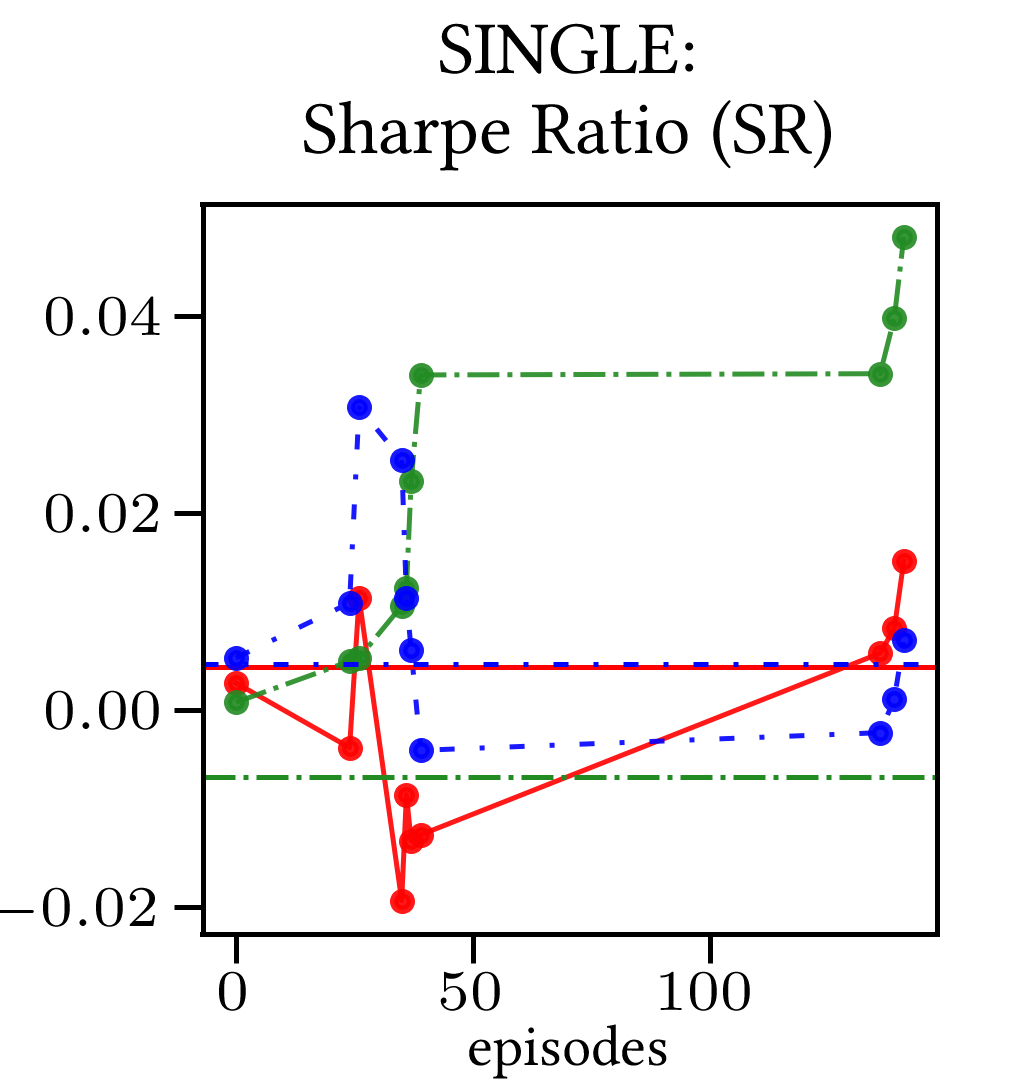}
  \caption{Top (\emph{Bottom}): Best model for profits (\emph{performance for training/evaluation/test set based on SR}), \btc}
  \Description{ADD DESCRIPTION}
  \label{Predictability1}
\end{figure}

\begin{figure}[h]
\includegraphics[width=0.49\linewidth]{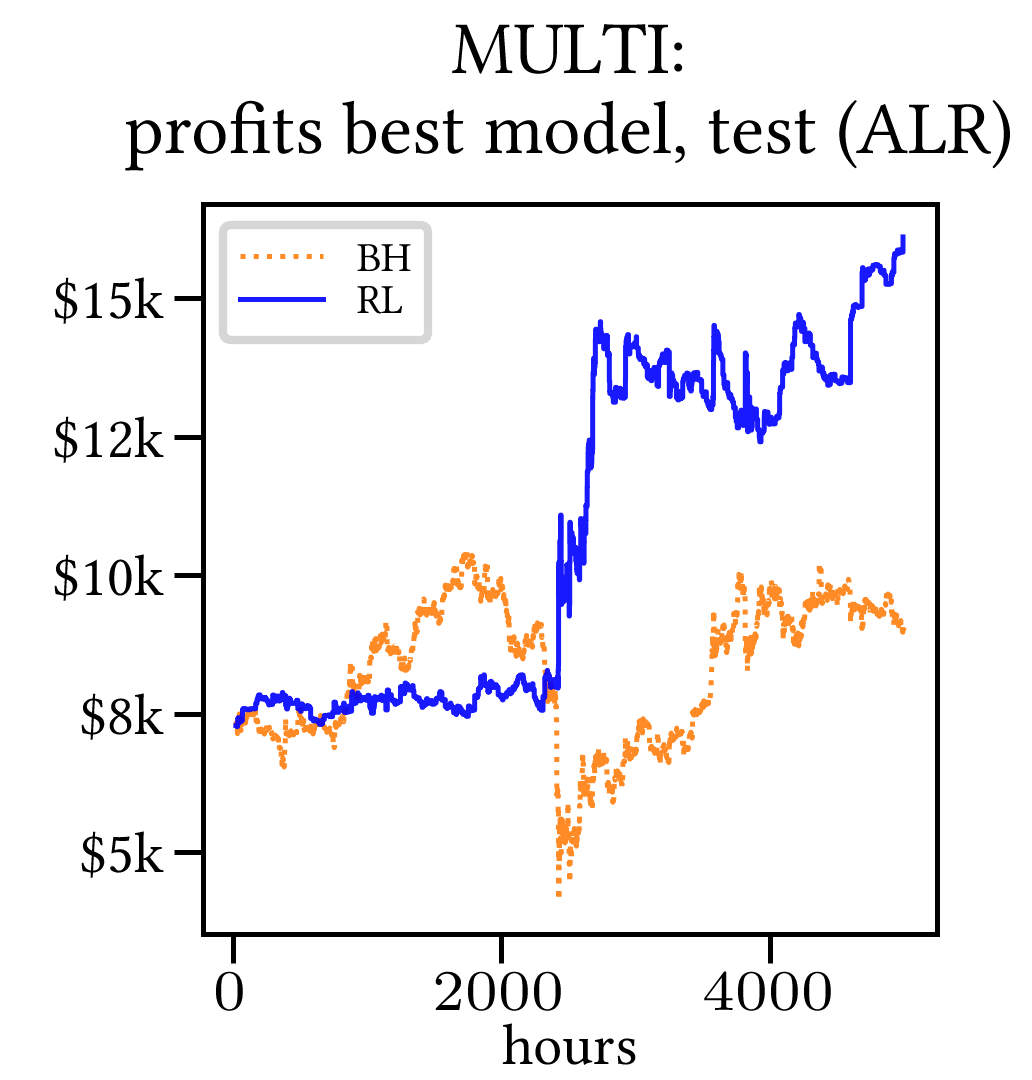}\includegraphics[width=0.49\linewidth]{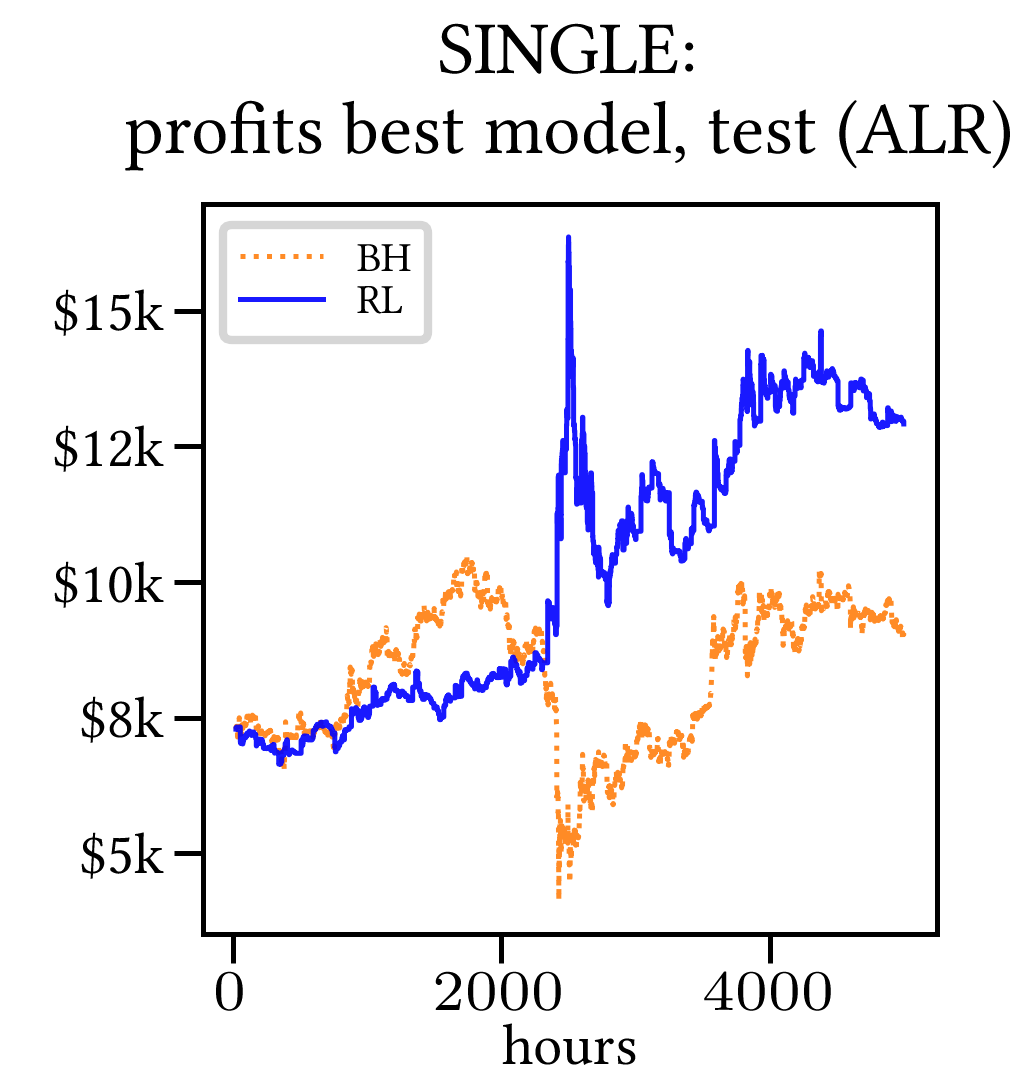}

\includegraphics[width=0.49\linewidth]{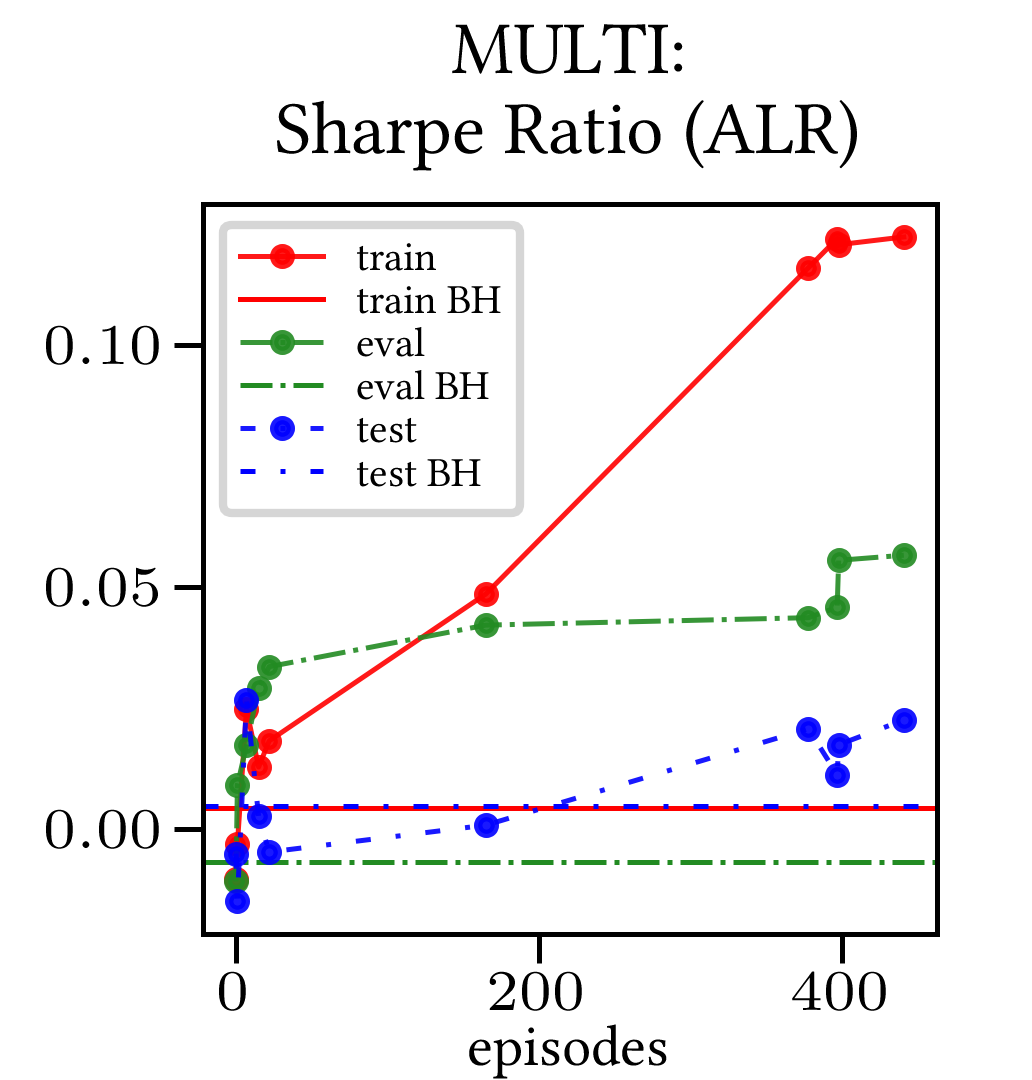}\includegraphics[width=0.49\linewidth]{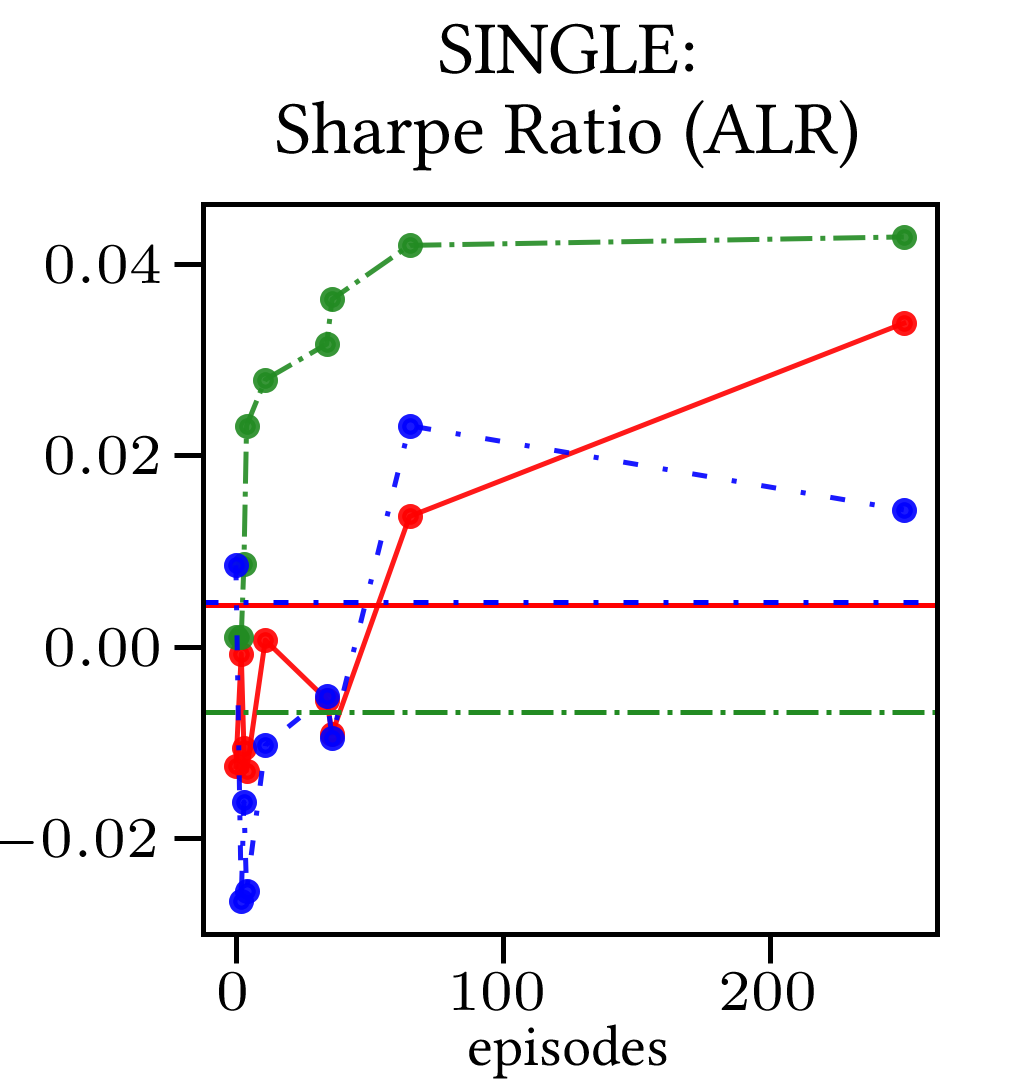}
   \caption{Top (\emph{Bottom}): Best model for profits (\emph{performance for training/evaluation/test set based on ALR}), \btc}
     \Description{ADD DESCRIPTION}
  \label{Predictability2}
\end{figure}

\begin{figure}[h]
\includegraphics[width=0.49\linewidth]{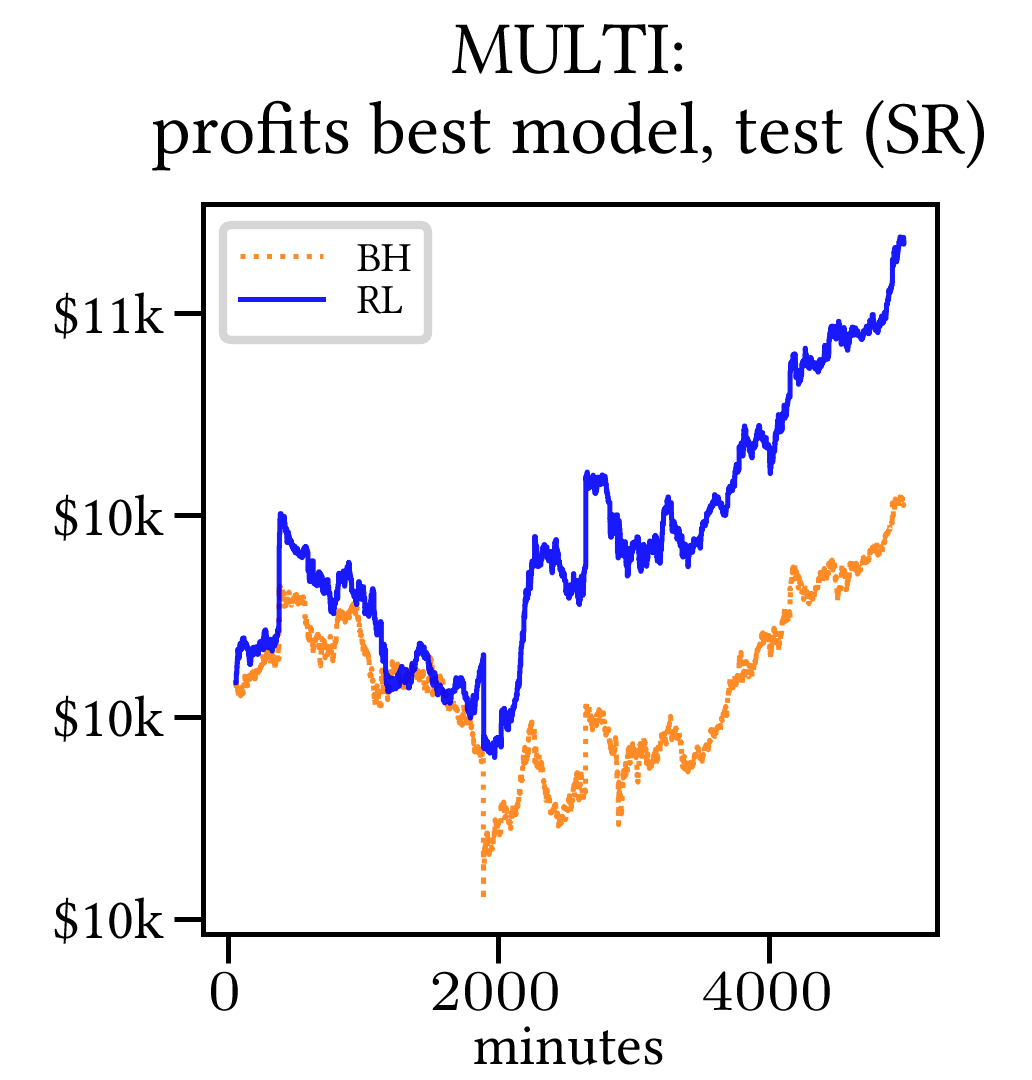}\includegraphics[width=0.49\linewidth]{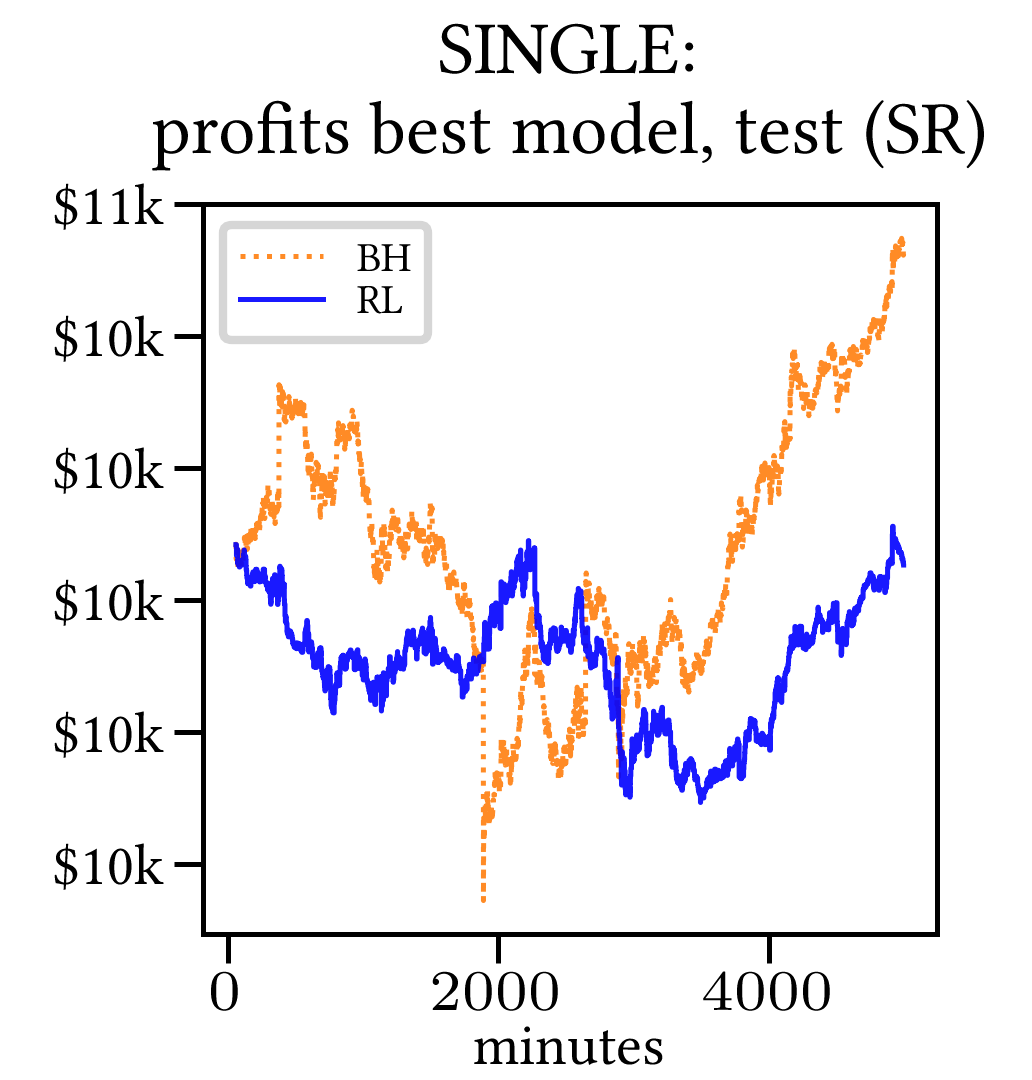}

\includegraphics[width=0.49\linewidth]{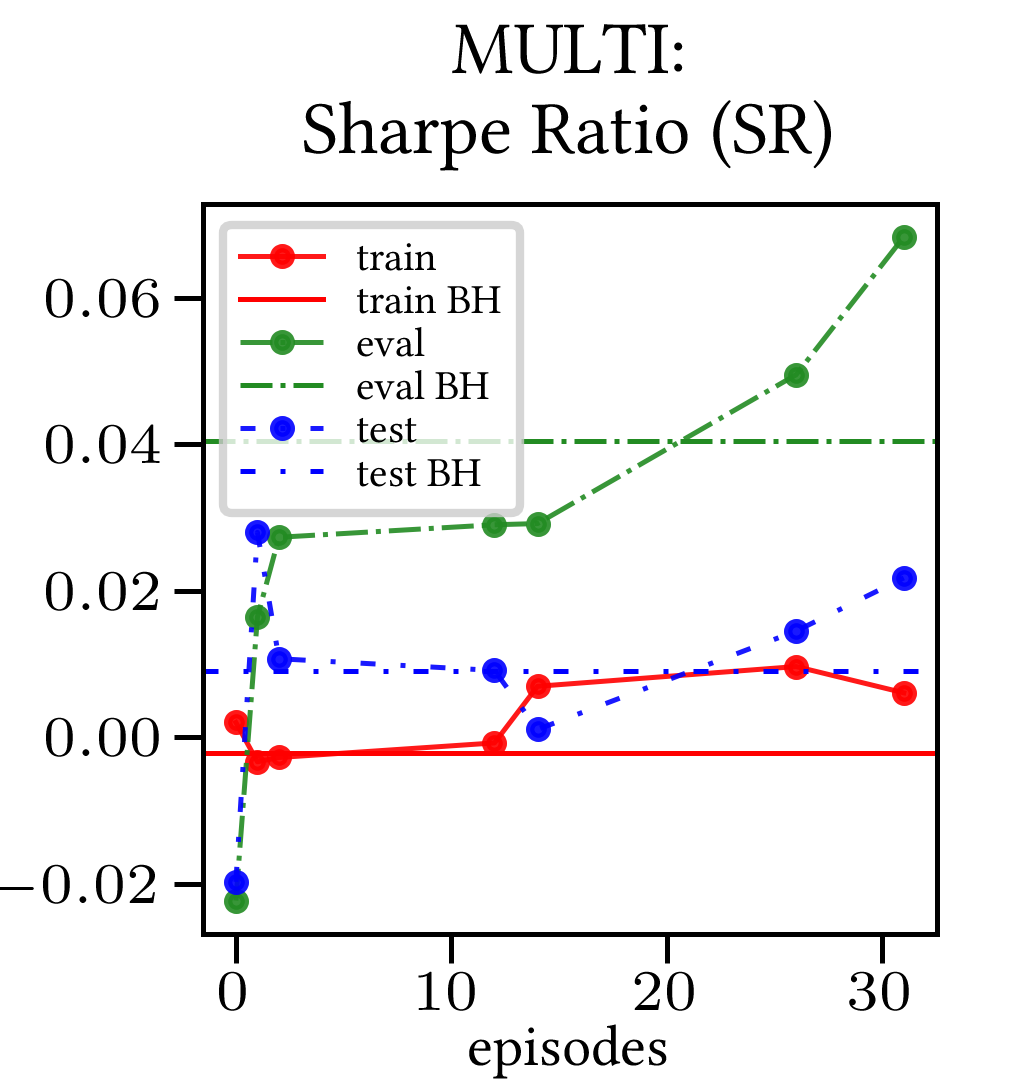}\includegraphics[width=0.49\linewidth]{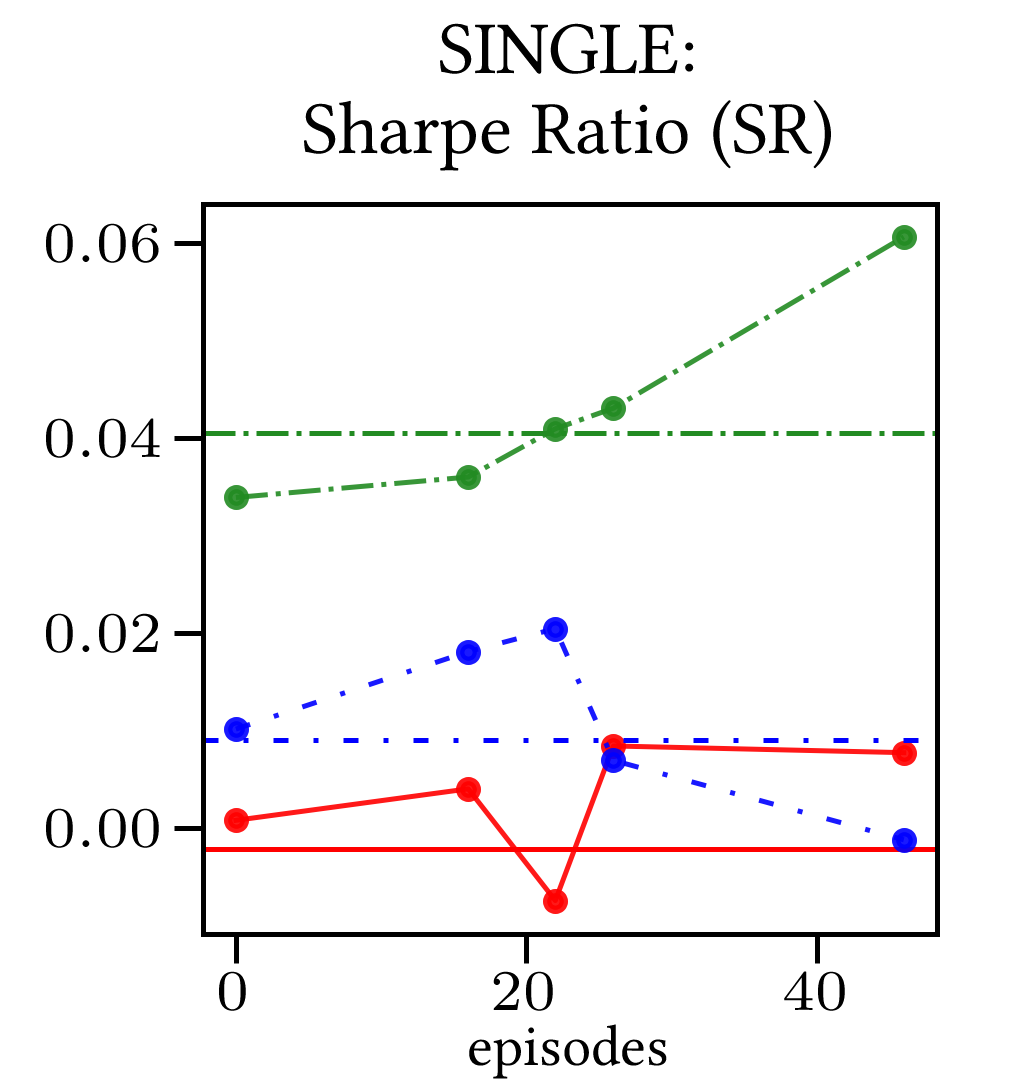}
   \caption{Top (\emph{Bottom}): Best model for profits (\emph{performance for training/evaluation/test set based on SR}), \nifty}
     \Description{ADD DESCRIPTION}
  \label{Predictability3}
\end{figure}

A commonly used model selection strategy is to pick the best performing model on
the evaluation set. In Figures \ref{Predictability1}, \ref{Predictability2}, \ref{Predictability3}, the
performance of such best performing model is shown (for training, evaluation, test)
as it progresses through the episodes.
We observe that the performance on the evaluation set (in terms of the Sharpe Ratio)
is consistently good. In particular, the performance of the Multi-Reward model
is at least as good as that of the Single-Reward model, while also being much
more stable\footnote{Further work will be needed in the future to account for
the exact impact of intrinsic RL noise.}. Furthermore, the profits on the test
sets are higher and more consistent in the Multi-Reward case (especially in the
case of \nifty{}, see Figure \ref{Predictability3}).
The results in Figures \ref{Predictability1}, \ref{Predictability2}, \ref{Predictability3} also show
that the performance on the test set is loosely correlated with the one on the
evaluation set. This is likely due to the noisiness of the learning process, and
a neat difference between evaluation and test set. In any case, the performance
on the evaluation is a reliable indication for the improved stability, as it
depends on the stability of the learning process.

 \begin{figure}[h]
   \centering
\includegraphics[]{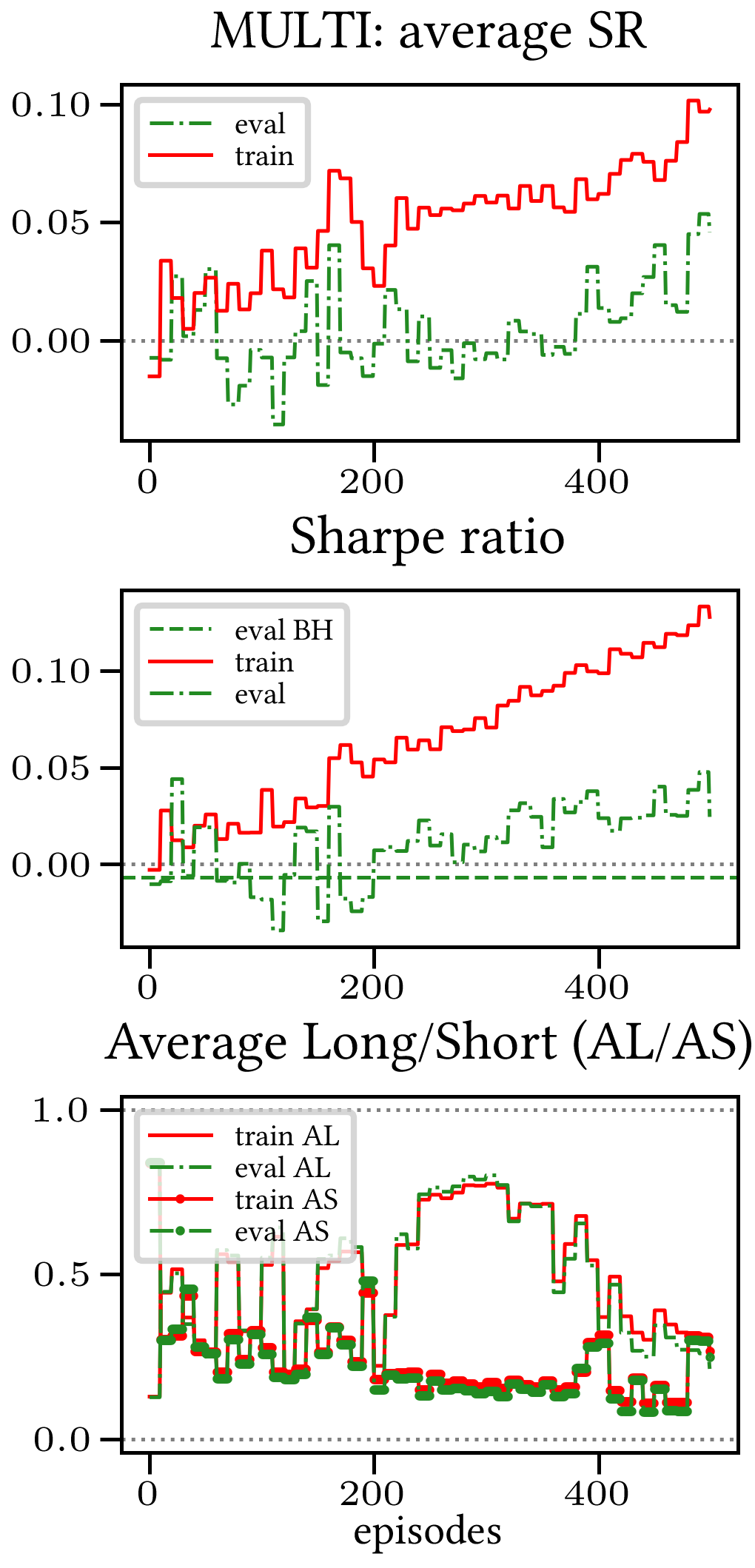}\includegraphics[]{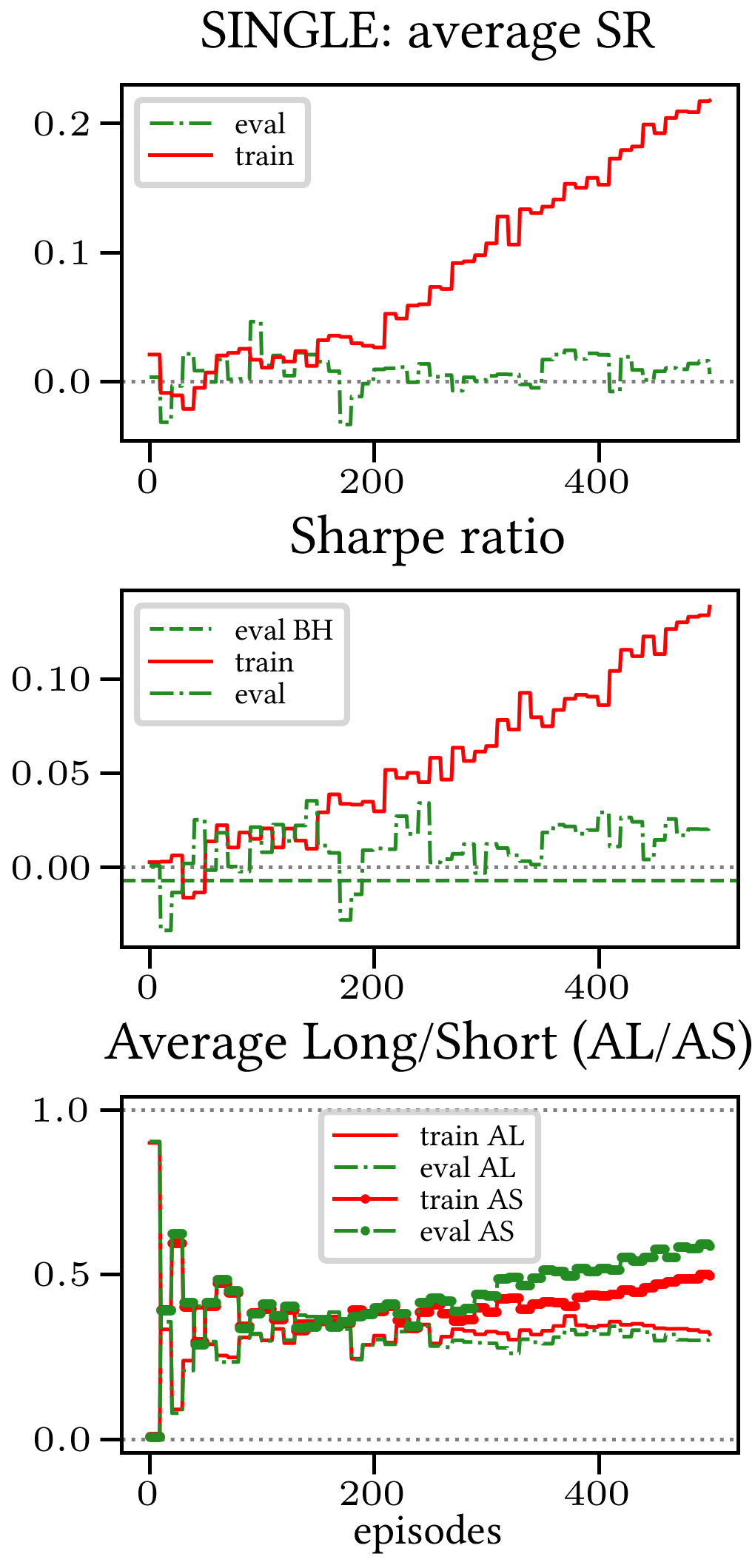}
  \caption{Reward Performances, SR/\LSPcompact/\btctwenty, regularized agent's network}
  \Description{pred1}
  \label{pred1}
\end{figure}

\begin{figure}[h]
   \centering
\includegraphics[]{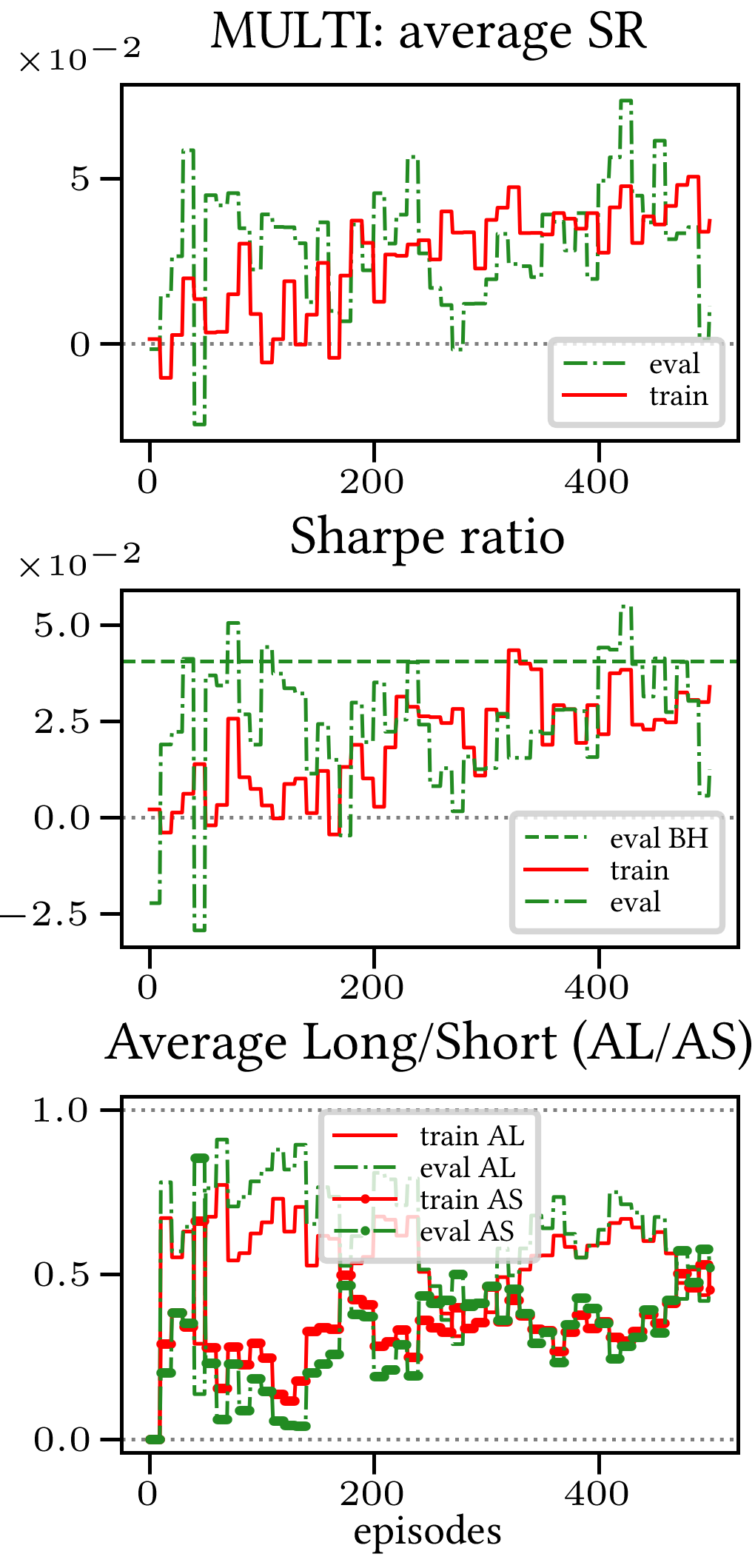}\includegraphics[]{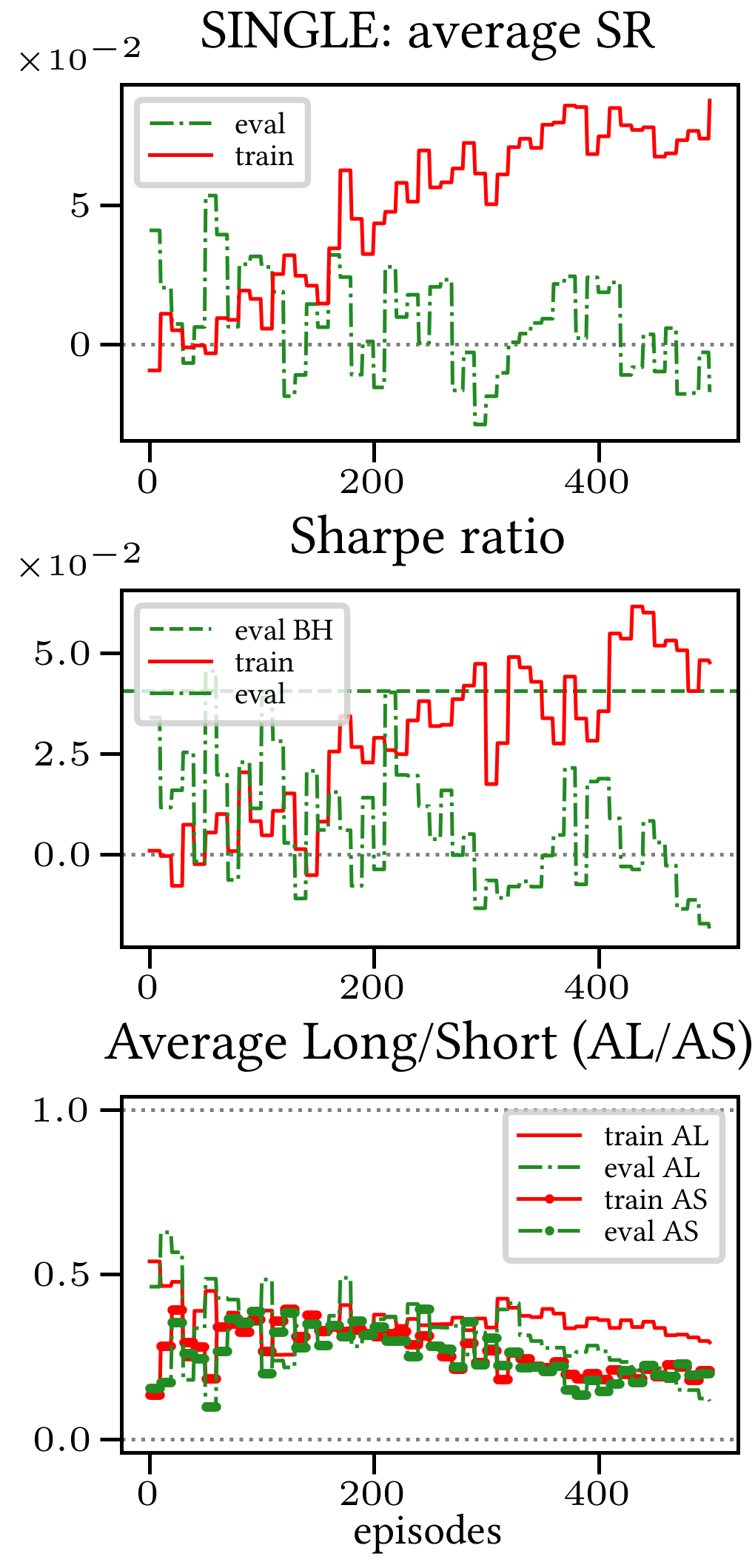}
  \caption{Reward Performances, ALR/\LSPcompact/\niftytwenty, regularized agent's network}
  \Description{pred2}
  \label{pred2}
\end{figure}

\section{Conclusions}\label{Conclusions}

Firstly, we have validated the generalization properties of a Multi-Reward, Reinforcement Learning method with Hindsight Experience Replay (in the declination given by \cite{friedman2018generalizing}) by running experiments on several important, single-asset datasets (\AAPL, \SPY, \ETH, \XRP, \btc, \nifty).

Secondly, even though a full statistical analysis would require further work, we have provided a number of consistent statistical indicators confirming the improved stability of our Multi-Reward method over its Single-Reward counterpart: distribution of performance over independent runs, 
convergence of prediction indicators,
and profits for best performing models.

Thirdly, we verified that the Multi-Reward has a clear edge over its Single-Reward counterparts in the case of sparse, heavily position-dependent reward mechanisms.

Finally, we have partially validated the generalization property regarding the discount factor (as suggested by \cite{friedman2018generalizing}), even though more work is required to consolidate the claim.

\subsection{Limits of the RL setting}\label{Limits}

The obtained results are, in nearly all occasions, subject to noise: more specifically, relevant performance indicators occasionally oscillate around -- rather than approach -- their limiting value (with respect to the number of epochs). 
This behavior, which is visible in the cumulative reward plots (Figures \ref{POWCRewardsLSP}--\ref{POWCRewardsLSPNifty}--\ref{rgammaRewardsLP-SR}--\ref{rgammaRewardsLSP-SR}--\ref{pred1}--\ref{pred2}), is consistent with what can be expected from a critic-only Reinforcement Learning approach. 
Additionally, despite the (not at all short) length of training, evaluation, and test set, the data is obviously correlated in time, thus the effectiveness of the method is reduced.
The impact of the noise is mitigated by running several experiments (Figures \ref{Fig_multi}, \ref{single_various_SR}, \ref{walk_for_LR}, \ref{single_various_POWC}, \ref{distributions}).

\section{Future Outlook}\label{Outlook}

While we have highlighted some of the benefits that a Multi-Reward approach has over a Single-Reward approach for predictive properties of a critic-only RL paradigm for single-asset financial data, many questions remain partially answered, or even wide open.

Firstly, while the Multi-Reward approach can stabilize and improve results of some rewards using the other ones, it is not clear exactly how the rewards are influenced by each other (normalization approaches different from \eqref{RewardRescaling} could be investigated).
Secondly, it would be interesting to conduct a more thorough research into different lengths and non-uniform sampling mechanisms for the experience replay (see \cite{fedus2020revisiting}).
Thirdly, a more thorough analysis on the use of a random discount factor should be conducted.
Fourthly, one might perform a sensitivity analysis on more hyperparameters.
Fifthly, a more in depth analysis of the prediction power to provide statistical evidence is still needed.
Finally, it would be interesting to further address the noisy convergence of the performance metrics (see Section \ref{Limits}).

\appendix
\section{RL algorithms}\label{AppendixRL_Algos}

The original Multi-Objective critic-only RL algorithm of Fontaine and Friedman \cite{friedman2018generalizing} in summarized in Algorithm \ref{alg:HERRL}, while our modified method (inclusive of, among others, random access point and discount factor generalization, see Subsection \ref{AbstractSetup}) is reported in Algorithm \ref{alg:HERRL-DFG} (only the changes against Algorithm \ref{alg:HERRL} are highlighted).

\begin{algorithm}[tb]
\caption{\alg}
\label{alg:HERRL}
\textbf{Input}: {\tt MultiReward} $\in\{\mbox{True, False}\}$\\
\textbf{Parameters}: ${\tt tol}\in (0,1)$, ${\tt batchsize},k,M \in \mathbb{N}$, $S \subset \{1,\dots,M\}$\\
\textbf{Output}: Trained Multi-Reward agent.\\
\begin{algorithmic}[1] 
\resetline{1}
\STATE Take one-hot encoding vector $\f{w}$.
\STATE Initialize network $Q\colon (\f{s},\f{w}) \mapsto [Q_1(\f{s},\f{w}),\dots,Q_P(\f{s},\f{w})]$ mapping state variables $\f{s}$ and weight vector $\f{w}$ to expected discounted return of every action $a\in\{1,\dots,P\}$. \label{state-actionNN}
\FOR{$i = 1,\dots, M$ }
\STATE Reset training environment.
\WHILE{Episode $i$ {\bfseries not} finished}
\IF {{\tt MultiReward} {\bfseries is} True}
\STATE Sample random reward weights $\f{w}$. \label{line-random-weight}
\ENDIF
\IF {Unif$(0,1) < tol$} 
\STATE Choose random action $a$
\ELSE
\STATE Pick $a \leftarrow \arg\max_{a}{Q(\f{s},\f{w})}$. \label{max-action}
\ENDIF
\STATE Conduct one step (get to new state $\f{s}_{new}$)  \label{line-step}
\STATE Get associated reward $\f{w}\cdot \f{r}(\f{s},a)$ \label{line-reward}
\STATE Append single experience $(\f{s},\f{w},\f{w}\cdot \f{r}(\f{s},a),a,\f{s}_{new})$ to experience replay $\mathcal{R}$ \label{line-append}
\STATE Set $\f{s}\leftarrow\f{s}_{new}$\\
\IF {{\tt MultiReward} {\bfseries is} True}
\STATE Add other $k$ experiences to Replay by re-running lines \ref{line-random-weight}-\ref{line-step}-\ref{line-reward}-\ref{line-append} $k$ times. \label{add-experiences}
\ENDIF \\
\IF {$i\in S$}
\STATE Randomly sample {\tt batchsize} units from $\mathcal{R}$. \label{sampleareplay}
\STATE Update $Q$-values associated with each sampled unit using Bellman equation \eqref{BellmanUpdate}.
\STATE Fit model on estimated $Q$-values \label{train}
\ENDIF 
\ENDWHILE
\ENDFOR
\end{algorithmic}
\end{algorithm}

\begin{algorithm}[tb]
\caption{\alg with discount factor generalization and random access point \emph{(Changes to lines of Algorithm \ref{alg:HERRL}})}
\label{alg:HERRL-DFG}
\textbf{Input}: {\tt Anchoring for cross-validation} $\in\{\mbox{True; False}\}$\\
\begin{algorithmic}[1] 
\resetline{2}
\STATE Initialize network $Q\colon (\f{s},\f{w},\gamma) \mapsto [Q_1(\f{s},\f{w},\gamma),\dots,Q_P(\f{s},\f{w},\gamma)]$ mapping state variables $\f{s}$, weight vector $\f{w}$, and discount factor $\gamma$ to expected discounted return of every action $a\in\{1,\dots,P\}$. \\
\begin{center}$\vdots$\end{center}
\resetline{4}
\IF {not {\tt Anchoring for cross-validation}}
\STATE Randomly select subset of training set, and reset associated environment.
\ENDIF
\begin{center}$\vdots$\end{center}
\resetline{7}
\STATE Sample random reward weights $\f{w}$ and discount factor $\gamma$. \\
\begin{center}$\vdots$\end{center}
\resetline{12}
\STATE Pick $a \leftarrow \arg\max_{a}{Q(\f{s},\f{w},\gamma)}$. \\
\begin{center}$\vdots$\end{center}
\resetline{16}
\STATE Append single experience $(\f{s},\gamma,\f{w},\f{w}\cdot \f{r}(\f{s},a),a,\f{s}_{new})$ to experience replay $\mathcal{R}$ \\
\begin{center}$\vdots$\end{center}
\resetline{22} 
\STATE Randomly sample {\tt batchsize} units from $\mathcal{R}$, and normalize sampled $Q$-values according to Subsection \ref{Normalization}.\label{sampleandnormalize}
\end{algorithmic}
\end{algorithm}

\bibliographystyle{ACM-Reference-Format}
\bibliography{KDD_bibliography}


\begin{thebibliography}{36}


\ifx \showCODEN    \undefined \def \showCODEN     #1{\unskip}     \fi
\ifx \showDOI      \undefined \def \showDOI       #1{#1}\fi
\ifx \showISBNx    \undefined \def \showISBNx     #1{\unskip}     \fi
\ifx \showISBNxiii \undefined \def \showISBNxiii  #1{\unskip}     \fi
\ifx \showISSN     \undefined \def \showISSN      #1{\unskip}     \fi
\ifx \showLCCN     \undefined \def \showLCCN      #1{\unskip}     \fi
\ifx \shownote     \undefined \def \shownote      #1{#1}          \fi
\ifx \showarticletitle \undefined \def \showarticletitle #1{#1}   \fi
\ifx \showURL      \undefined \def \showURL       {\relax}        \fi
\providecommand\bibfield[2]{#2}
\providecommand\bibinfo[2]{#2}
\providecommand\natexlab[1]{#1}
\providecommand\showeprint[2][]{arXiv:#2}

\bibitem[\protect\citeauthoryear{Andrychowicz, Wolski, Ray, Schneider, Fong,
  Welinder, McGrew, Tobin, Pieter~Abbeel, and Zaremba}{Andrychowicz
  et~al\mbox{.}}{2017}]%
        {andrychowicz2017hindsight}
\bibfield{author}{\bibinfo{person}{Marcin Andrychowicz}, \bibinfo{person}{Filip
  Wolski}, \bibinfo{person}{Alex Ray}, \bibinfo{person}{Jonas Schneider},
  \bibinfo{person}{Rachel Fong}, \bibinfo{person}{Peter Welinder},
  \bibinfo{person}{Bob McGrew}, \bibinfo{person}{Josh Tobin},
  \bibinfo{person}{OpenAI Pieter~Abbeel}, {and} \bibinfo{person}{Wojciech
  Zaremba}.} \bibinfo{year}{2017}\natexlab{}.
\newblock \showarticletitle{Hindsight experience replay}.
\newblock \bibinfo{journal}{\emph{Advances in neural information processing
  systems}}  \bibinfo{volume}{30} (\bibinfo{year}{2017}).
\newblock


\bibitem[\protect\citeauthoryear{Barrett and Narayanan}{Barrett and
  Narayanan}{2008}]%
        {barrett2008learning}
\bibfield{author}{\bibinfo{person}{Leon Barrett} {and} \bibinfo{person}{Srini
  Narayanan}.} \bibinfo{year}{2008}\natexlab{}.
\newblock \showarticletitle{Learning all optimal policies with multiple
  criteria}. In \bibinfo{booktitle}{\emph{Proceedings of the 25th international
  conference on Machine learning}}. \bibinfo{pages}{41--47}.
\newblock


\bibitem[\protect\citeauthoryear{Bekiros}{Bekiros}{2010}]%
        {bekiros2010heterogeneous}
\bibfield{author}{\bibinfo{person}{Stelios~D Bekiros}.}
  \bibinfo{year}{2010}\natexlab{}.
\newblock \showarticletitle{Heterogeneous trading strategies with adaptive
  fuzzy actor--critic reinforcement learning: A behavioral approach}.
\newblock \bibinfo{journal}{\emph{Journal of Economic Dynamics and Control}}
  \bibinfo{volume}{34}, \bibinfo{number}{6} (\bibinfo{year}{2010}),
  \bibinfo{pages}{1153--1170}.
\newblock


\bibitem[\protect\citeauthoryear{Bisht and Kumar}{Bisht and Kumar}{2020}]%
        {bisht2020deep}
\bibfield{author}{\bibinfo{person}{Kiran Bisht} {and} \bibinfo{person}{Arun
  Kumar}.} \bibinfo{year}{2020}\natexlab{}.
\newblock \showarticletitle{Deep Reinforcement Learning based Multi-Objective
  Systems for Financial Trading}. In \bibinfo{booktitle}{\emph{2020 5th IEEE
  International Conference on Recent Advances and Innovations in Engineering
  (ICRAIE)}}. IEEE, \bibinfo{pages}{1--6}.
\newblock


\bibitem[\protect\citeauthoryear{Chan and Shelton}{Chan and Shelton}{2001}]%
        {RePEc:sce:scecf1:146}
\bibfield{author}{\bibinfo{person}{Nicholas~T. Chan} {and}
  \bibinfo{person}{Christian Shelton}.} \bibinfo{year}{2001}\natexlab{}.
\newblock \bibinfo{booktitle}{\emph{An Adaptive Electronic Market-Maker}}.
\newblock \bibinfo{type}{Computing in Economics and Finance 2001} 146.
  \bibinfo{institution}{Society for Computational Economics}.
\newblock
\urldef\tempurl%
\url{https://EconPapers.repec.org/RePEc:sce:scecf1:146}
\showURL{%
\tempurl}


\bibitem[\protect\citeauthoryear{Corazza and Bertoluzzo}{Corazza and
  Bertoluzzo}{2014}]%
        {RePEc:ven:wpaper:2014:15}
\bibfield{author}{\bibinfo{person}{Marco Corazza} {and}
  \bibinfo{person}{Francesco Bertoluzzo}.} \bibinfo{year}{2014}\natexlab{}.
\newblock \bibinfo{booktitle}{\emph{Q-Learning-based financial trading systems
  with applications}}.
\newblock \bibinfo{type}{Working Papers} 2014:15.
  \bibinfo{institution}{Department of Economics, University of Venice "Ca'
  Foscari"}.
\newblock
\urldef\tempurl%
\url{https://EconPapers.repec.org/RePEc:ven:wpaper:2014:15}
\showURL{%
\tempurl}


\bibitem[\protect\citeauthoryear{Dempster and Leemans}{Dempster and
  Leemans}{2006}]%
        {dempster2006automated}
\bibfield{author}{\bibinfo{person}{Michael~AH Dempster} {and}
  \bibinfo{person}{Vasco Leemans}.} \bibinfo{year}{2006}\natexlab{}.
\newblock \showarticletitle{An automated FX trading system using adaptive
  reinforcement learning}.
\newblock \bibinfo{journal}{\emph{Expert Systems with Applications}}
  \bibinfo{volume}{30}, \bibinfo{number}{3} (\bibinfo{year}{2006}),
  \bibinfo{pages}{543--552}.
\newblock


\bibitem[\protect\citeauthoryear{Dempster, Payne, Romahi, and
  Thompson}{Dempster et~al\mbox{.}}{2001}]%
        {dempster2001computational}
\bibfield{author}{\bibinfo{person}{Michael~AH Dempster}, \bibinfo{person}{Tom~W
  Payne}, \bibinfo{person}{Yazann Romahi}, {and} \bibinfo{person}{Giles~WP
  Thompson}.} \bibinfo{year}{2001}\natexlab{}.
\newblock \showarticletitle{Computational learning techniques for intraday FX
  trading using popular technical indicators}.
\newblock \bibinfo{journal}{\emph{IEEE Transactions on neural networks}}
  \bibinfo{volume}{12}, \bibinfo{number}{4} (\bibinfo{year}{2001}),
  \bibinfo{pages}{744--754}.
\newblock


\bibitem[\protect\citeauthoryear{Deng, Bao, Kong, Ren, and Dai}{Deng
  et~al\mbox{.}}{2016}]%
        {deng2016deep}
\bibfield{author}{\bibinfo{person}{Yue Deng}, \bibinfo{person}{Feng Bao},
  \bibinfo{person}{Youyong Kong}, \bibinfo{person}{Zhiquan Ren}, {and}
  \bibinfo{person}{Qionghai Dai}.} \bibinfo{year}{2016}\natexlab{}.
\newblock \showarticletitle{Deep direct reinforcement learning for financial
  signal representation and trading}.
\newblock \bibinfo{journal}{\emph{IEEE transactions on neural networks and
  learning systems}} \bibinfo{volume}{28}, \bibinfo{number}{3}
  (\bibinfo{year}{2016}), \bibinfo{pages}{653--664}.
\newblock


\bibitem[\protect\citeauthoryear{Eilers, Dunis, von Mettenheim, and
  Breitner}{Eilers et~al\mbox{.}}{2014}]%
        {eilers2014intelligent}
\bibfield{author}{\bibinfo{person}{Dennis Eilers}, \bibinfo{person}{Christian~L
  Dunis}, \bibinfo{person}{Hans-J{\"o}rg von Mettenheim}, {and}
  \bibinfo{person}{Michael~H Breitner}.} \bibinfo{year}{2014}\natexlab{}.
\newblock \showarticletitle{Intelligent trading of seasonal effects: A decision
  support algorithm based on reinforcement learning}.
\newblock \bibinfo{journal}{\emph{Decision support systems}}
  \bibinfo{volume}{64} (\bibinfo{year}{2014}), \bibinfo{pages}{100--108}.
\newblock


\bibitem[\protect\citeauthoryear{Fedus, Ramachandran, Agarwal, Bengio,
  Larochelle, Rowland, and Dabney}{Fedus et~al\mbox{.}}{2020}]%
        {fedus2020revisiting}
\bibfield{author}{\bibinfo{person}{William Fedus}, \bibinfo{person}{Prajit
  Ramachandran}, \bibinfo{person}{Rishabh Agarwal}, \bibinfo{person}{Yoshua
  Bengio}, \bibinfo{person}{Hugo Larochelle}, \bibinfo{person}{Mark Rowland},
  {and} \bibinfo{person}{Will Dabney}.} \bibinfo{year}{2020}\natexlab{}.
\newblock \showarticletitle{Revisiting fundamentals of experience replay}. In
  \bibinfo{booktitle}{\emph{International Conference on Machine Learning}}.
  PMLR, \bibinfo{pages}{3061--3071}.
\newblock


\bibitem[\protect\citeauthoryear{Fischer}{Fischer}{2018}]%
        {fischer2018reinforcement}
\bibfield{author}{\bibinfo{person}{Thomas~G Fischer}.}
  \bibinfo{year}{2018}\natexlab{}.
\newblock \bibinfo{booktitle}{\emph{Reinforcement learning in financial
  markets-a survey}}.
\newblock \bibinfo{type}{{T}echnical {R}eport}. \bibinfo{institution}{FAU
  Discussion Papers in Economics}.
\newblock


\bibitem[\protect\citeauthoryear{Friedman and Fontaine}{Friedman and
  Fontaine}{2018}]%
        {friedman2018generalizing}
\bibfield{author}{\bibinfo{person}{Eli Friedman} {and} \bibinfo{person}{Fred
  Fontaine}.} \bibinfo{year}{2018}\natexlab{}.
\newblock \showarticletitle{Generalizing across multi-objective reward
  functions in deep reinforcement learning}.
\newblock \bibinfo{journal}{\emph{arXiv preprint arXiv:1809.06364}}
  (\bibinfo{year}{2018}).
\newblock


\bibitem[\protect\citeauthoryear{Gold}{Gold}{2003}]%
        {gold2003fx}
\bibfield{author}{\bibinfo{person}{Carl Gold}.}
  \bibinfo{year}{2003}\natexlab{}.
\newblock \showarticletitle{FX trading via recurrent reinforcement learning}.
  In \bibinfo{booktitle}{\emph{2003 IEEE International Conference on
  Computational Intelligence for Financial Engineering, 2003. Proceedings.}}
  IEEE, \bibinfo{pages}{363--370}.
\newblock


\bibitem[\protect\citeauthoryear{Gu, Mabu, Yang, Li, and Hirasawa}{Gu
  et~al\mbox{.}}{2011}]%
        {gu2011trading}
\bibfield{author}{\bibinfo{person}{Yunqing Gu}, \bibinfo{person}{Shingo Mabu},
  \bibinfo{person}{Yang Yang}, \bibinfo{person}{Jianhua Li}, {and}
  \bibinfo{person}{Kotaro Hirasawa}.} \bibinfo{year}{2011}\natexlab{}.
\newblock \showarticletitle{Trading rules on stock markets using Genetic
  Network Programming-Sarsa learning with plural subroutines}. In
  \bibinfo{booktitle}{\emph{SICE Annual Conference 2011}}. IEEE,
  \bibinfo{pages}{143--148}.
\newblock


\bibitem[\protect\citeauthoryear{Jangmin, Lee, Lee, and Zhang}{Jangmin
  et~al\mbox{.}}{2006}]%
        {jangmin2006adaptive}
\bibfield{author}{\bibinfo{person}{O Jangmin}, \bibinfo{person}{Jongwoo Lee},
  \bibinfo{person}{Jae~Won Lee}, {and} \bibinfo{person}{Byoung-Tak Zhang}.}
  \bibinfo{year}{2006}\natexlab{}.
\newblock \showarticletitle{Adaptive stock trading with dynamic asset
  allocation using reinforcement learning}.
\newblock \bibinfo{journal}{\emph{Information Sciences}} \bibinfo{volume}{176},
  \bibinfo{number}{15} (\bibinfo{year}{2006}), \bibinfo{pages}{2121--2147}.
\newblock


\bibitem[\protect\citeauthoryear{Jiang, Xu, and Liang}{Jiang
  et~al\mbox{.}}{2017}]%
        {jiang2017deep}
\bibfield{author}{\bibinfo{person}{Zhengyao Jiang}, \bibinfo{person}{Dixing
  Xu}, {and} \bibinfo{person}{Jinjun Liang}.} \bibinfo{year}{2017}\natexlab{}.
\newblock \showarticletitle{A deep reinforcement learning framework for the
  financial portfolio management problem}.
\newblock \bibinfo{journal}{\emph{arXiv preprint arXiv:1706.10059}}
  (\bibinfo{year}{2017}).
\newblock


\bibitem[\protect\citeauthoryear{Jin and El-Saawy}{Jin and El-Saawy}{2016}]%
        {jin2016portfolio}
\bibfield{author}{\bibinfo{person}{Olivier Jin} {and} \bibinfo{person}{Hamza
  El-Saawy}.} \bibinfo{year}{2016}\natexlab{}.
\newblock \showarticletitle{Portfolio management using reinforcement learning}.
\newblock \bibinfo{journal}{\emph{Stanford University}} (\bibinfo{year}{2016}).
\newblock


\bibitem[\protect\citeauthoryear{Kaur}{Kaur}{2017}]%
        {kaur2017algorithmic}
\bibfield{author}{\bibinfo{person}{S Kaur}.} \bibinfo{year}{2017}\natexlab{}.
\newblock \bibinfo{booktitle}{\emph{Algorithmic trading using reinforcement
  learning augmented with hidden Markov model}}.
\newblock \bibinfo{type}{{T}echnical {R}eport}. \bibinfo{institution}{Working
  paper, Stanford University}.
\newblock


\bibitem[\protect\citeauthoryear{Kober, Bagnell, and Peters}{Kober
  et~al\mbox{.}}{2013}]%
        {kober2013reinforcement}
\bibfield{author}{\bibinfo{person}{Jens Kober}, \bibinfo{person}{J~Andrew
  Bagnell}, {and} \bibinfo{person}{Jan Peters}.}
  \bibinfo{year}{2013}\natexlab{}.
\newblock \showarticletitle{Reinforcement learning in robotics: A survey}.
\newblock \bibinfo{journal}{\emph{The International Journal of Robotics
  Research}} \bibinfo{volume}{32}, \bibinfo{number}{11} (\bibinfo{year}{2013}),
  \bibinfo{pages}{1238--1274}.
\newblock


\bibitem[\protect\citeauthoryear{Lee and Jangmin}{Lee and Jangmin}{2002}]%
        {lee2002multi}
\bibfield{author}{\bibinfo{person}{Jae~Won Lee} {and} \bibinfo{person}{O
  Jangmin}.} \bibinfo{year}{2002}\natexlab{}.
\newblock \showarticletitle{A multi-agent Q-learning framework for optimizing
  stock trading systems}. In \bibinfo{booktitle}{\emph{International Conference
  on Database and Expert Systems Applications}}. Springer,
  \bibinfo{pages}{153--162}.
\newblock


\bibitem[\protect\citeauthoryear{Li, Dagli, and Enke}{Li et~al\mbox{.}}{2007}]%
        {li2007short}
\bibfield{author}{\bibinfo{person}{Hailin Li}, \bibinfo{person}{Cihan~H Dagli},
  {and} \bibinfo{person}{David Enke}.} \bibinfo{year}{2007}\natexlab{}.
\newblock \showarticletitle{Short-term stock market timing prediction under
  reinforcement learning schemes}. In \bibinfo{booktitle}{\emph{2007 IEEE
  International Symposium on Approximate Dynamic Programming and Reinforcement
  Learning}}. IEEE, \bibinfo{pages}{233--240}.
\newblock


\bibitem[\protect\citeauthoryear{Mao, Alizadeh, Menache, and Kandula}{Mao
  et~al\mbox{.}}{2016}]%
        {mao2016resource}
\bibfield{author}{\bibinfo{person}{Hongzi Mao}, \bibinfo{person}{Mohammad
  Alizadeh}, \bibinfo{person}{Ishai Menache}, {and} \bibinfo{person}{Srikanth
  Kandula}.} \bibinfo{year}{2016}\natexlab{}.
\newblock \showarticletitle{Resource management with deep reinforcement
  learning}. In \bibinfo{booktitle}{\emph{Proceedings of the 15th ACM workshop
  on hot topics in networks}}. \bibinfo{pages}{50--56}.
\newblock


\bibitem[\protect\citeauthoryear{Mnih, Kavukcuoglu, Silver, Graves, Antonoglou,
  Wierstra, and Riedmiller}{Mnih et~al\mbox{.}}{2013}]%
        {mnih2013playing}
\bibfield{author}{\bibinfo{person}{Volodymyr Mnih}, \bibinfo{person}{Koray
  Kavukcuoglu}, \bibinfo{person}{David Silver}, \bibinfo{person}{Alex Graves},
  \bibinfo{person}{Ioannis Antonoglou}, \bibinfo{person}{Daan Wierstra}, {and}
  \bibinfo{person}{Martin Riedmiller}.} \bibinfo{year}{2013}\natexlab{}.
\newblock \showarticletitle{Playing atari with deep reinforcement learning}.
\newblock \bibinfo{journal}{\emph{arXiv preprint arXiv:1312.5602}}
  (\bibinfo{year}{2013}).
\newblock


\bibitem[\protect\citeauthoryear{Moody, Wu, Liao, and Saffell}{Moody
  et~al\mbox{.}}{1998}]%
        {moody1998performance}
\bibfield{author}{\bibinfo{person}{John Moody}, \bibinfo{person}{Lizhong Wu},
  \bibinfo{person}{Yuansong Liao}, {and} \bibinfo{person}{Matthew Saffell}.}
  \bibinfo{year}{1998}\natexlab{}.
\newblock \showarticletitle{Performance functions and reinforcement learning
  for trading systems and portfolios}.
\newblock \bibinfo{journal}{\emph{Journal of Forecasting}}
  \bibinfo{volume}{17}, \bibinfo{number}{5-6} (\bibinfo{year}{1998}),
  \bibinfo{pages}{441--470}.
\newblock


\bibitem[\protect\citeauthoryear{Natarajan and Tadepalli}{Natarajan and
  Tadepalli}{2005}]%
        {natarajan2005dynamic}
\bibfield{author}{\bibinfo{person}{Sriraam Natarajan} {and}
  \bibinfo{person}{Prasad Tadepalli}.} \bibinfo{year}{2005}\natexlab{}.
\newblock \showarticletitle{Dynamic preferences in multi-criteria reinforcement
  learning}. In \bibinfo{booktitle}{\emph{Proceedings of the 22nd international
  conference on Machine learning}}. \bibinfo{pages}{601--608}.
\newblock


\bibitem[\protect\citeauthoryear{Neuneier}{Neuneier}{1995}]%
        {neuneier1995optimal}
\bibfield{author}{\bibinfo{person}{Ralph Neuneier}.}
  \bibinfo{year}{1995}\natexlab{}.
\newblock \showarticletitle{Optimal asset allocation using adaptive dynamic
  programming}.
\newblock \bibinfo{journal}{\emph{Advances in Neural Information Processing
  Systems}}  \bibinfo{volume}{8} (\bibinfo{year}{1995}).
\newblock


\bibitem[\protect\citeauthoryear{Nevmyvaka, Feng, and Kearns}{Nevmyvaka
  et~al\mbox{.}}{2006}]%
        {nevmyvaka2006reinforcement}
\bibfield{author}{\bibinfo{person}{Yuriy Nevmyvaka}, \bibinfo{person}{Yi Feng},
  {and} \bibinfo{person}{Michael Kearns}.} \bibinfo{year}{2006}\natexlab{}.
\newblock \showarticletitle{Reinforcement learning for optimized trade
  execution}. In \bibinfo{booktitle}{\emph{Proceedings of the 23rd
  international conference on Machine learning}}. \bibinfo{pages}{673--680}.
\newblock


\bibitem[\protect\citeauthoryear{Roijers, Vamplew, Whiteson, and
  Dazeley}{Roijers et~al\mbox{.}}{2013}]%
        {roijers2013survey}
\bibfield{author}{\bibinfo{person}{Diederik~M Roijers}, \bibinfo{person}{Peter
  Vamplew}, \bibinfo{person}{Shimon Whiteson}, {and} \bibinfo{person}{Richard
  Dazeley}.} \bibinfo{year}{2013}\natexlab{}.
\newblock \showarticletitle{A survey of multi-objective sequential
  decision-making}.
\newblock \bibinfo{journal}{\emph{Journal of Artificial Intelligence Research}}
   \bibinfo{volume}{48} (\bibinfo{year}{2013}), \bibinfo{pages}{67--113}.
\newblock


\bibitem[\protect\citeauthoryear{Sherstov and Stone}{Sherstov and
  Stone}{2004}]%
        {sherstov2004three}
\bibfield{author}{\bibinfo{person}{Alexander~A Sherstov} {and}
  \bibinfo{person}{Peter Stone}.} \bibinfo{year}{2004}\natexlab{}.
\newblock \showarticletitle{Three automated stock-trading agents: A comparative
  study}. In \bibinfo{booktitle}{\emph{International Workshop on Agent-Mediated
  Electronic Commerce}}. Springer, \bibinfo{pages}{173--187}.
\newblock


\bibitem[\protect\citeauthoryear{Si, Li, Ding, and Rao}{Si
  et~al\mbox{.}}{2017}]%
        {si2017multi}
\bibfield{author}{\bibinfo{person}{Weiyu Si}, \bibinfo{person}{Jinke Li},
  \bibinfo{person}{Peng Ding}, {and} \bibinfo{person}{Ruonan Rao}.}
  \bibinfo{year}{2017}\natexlab{}.
\newblock \showarticletitle{A multi-objective deep reinforcement learning
  approach for stock index future’s intraday trading}. In
  \bibinfo{booktitle}{\emph{2017 10th International symposium on computational
  intelligence and design (ISCID)}}, Vol.~\bibinfo{volume}{2}. IEEE,
  \bibinfo{pages}{431--436}.
\newblock


\bibitem[\protect\citeauthoryear{Sutton and Barto}{Sutton and Barto}{2018}]%
        {sutton2018reinforcement}
\bibfield{author}{\bibinfo{person}{Richard~S Sutton} {and}
  \bibinfo{person}{Andrew~G Barto}.} \bibinfo{year}{2018}\natexlab{}.
\newblock \bibinfo{booktitle}{\emph{Reinforcement learning: An introduction}}.
\newblock \bibinfo{publisher}{MIT press}.
\newblock


\bibitem[\protect\citeauthoryear{Tan, Quek, and Cheng}{Tan
  et~al\mbox{.}}{2011}]%
        {tan2011stock}
\bibfield{author}{\bibinfo{person}{Zhiyong Tan}, \bibinfo{person}{Chai Quek},
  {and} \bibinfo{person}{Philip~YK Cheng}.} \bibinfo{year}{2011}\natexlab{}.
\newblock \showarticletitle{Stock trading with cycles: A financial application
  of ANFIS and reinforcement learning}.
\newblock \bibinfo{journal}{\emph{Expert Systems with Applications}}
  \bibinfo{volume}{38}, \bibinfo{number}{5} (\bibinfo{year}{2011}),
  \bibinfo{pages}{4741--4755}.
\newblock


\bibitem[\protect\citeauthoryear{Watts}{Watts}{2015}]%
        {watts2015hedging}
\bibfield{author}{\bibinfo{person}{Samuel Watts}.}
  \bibinfo{year}{2015}\natexlab{}.
\newblock \bibinfo{booktitle}{\emph{Hedging basis risk using reinforcement
  learning}}.
\newblock \bibinfo{type}{{T}echnical {R}eport}. \bibinfo{institution}{Technical
  report, Working Paper, University of Oxford}.
\newblock


\bibitem[\protect\citeauthoryear{Zheng, Zhang, Zheng, Xiang, Yuan, Xie, and
  Li}{Zheng et~al\mbox{.}}{2018}]%
        {zheng2018drn}
\bibfield{author}{\bibinfo{person}{Guanjie Zheng}, \bibinfo{person}{Fuzheng
  Zhang}, \bibinfo{person}{Zihan Zheng}, \bibinfo{person}{Yang Xiang},
  \bibinfo{person}{Nicholas~Jing Yuan}, \bibinfo{person}{Xing Xie}, {and}
  \bibinfo{person}{Zhenhui Li}.} \bibinfo{year}{2018}\natexlab{}.
\newblock \showarticletitle{DRN: A deep reinforcement learning framework for
  news recommendation}. In \bibinfo{booktitle}{\emph{Proceedings of the 2018
  World Wide Web Conference}}. \bibinfo{pages}{167--176}.
\newblock


\bibitem[\protect\citeauthoryear{Zitzler, Knowles, and Thiele}{Zitzler
  et~al\mbox{.}}{2008}]%
        {zitzler2008quality}
\bibfield{author}{\bibinfo{person}{Eckart Zitzler}, \bibinfo{person}{Joshua
  Knowles}, {and} \bibinfo{person}{Lothar Thiele}.}
  \bibinfo{year}{2008}\natexlab{}.
\newblock \showarticletitle{Quality assessment of pareto set approximations}.
\newblock \bibinfo{journal}{\emph{Multiobjective optimization}}
  (\bibinfo{year}{2008}), \bibinfo{pages}{373--404}.
\newblock


\end{thebibliography}



\begin{thebibliography}{49}


\ifx \showCODEN    \undefined \def \showCODEN     #1{\unskip}     \fi
\ifx \showDOI      \undefined \def \showDOI       #1{#1}\fi
\ifx \showISBNx    \undefined \def \showISBNx     #1{\unskip}     \fi
\ifx \showISBNxiii \undefined \def \showISBNxiii  #1{\unskip}     \fi
\ifx \showISSN     \undefined \def \showISSN      #1{\unskip}     \fi
\ifx \showLCCN     \undefined \def \showLCCN      #1{\unskip}     \fi
\ifx \shownote     \undefined \def \shownote      #1{#1}          \fi
\ifx \showarticletitle \undefined \def \showarticletitle #1{#1}   \fi
\ifx \showURL      \undefined \def \showURL       {\relax}        \fi
\providecommand\bibfield[2]{#2}
\providecommand\bibinfo[2]{#2}
\providecommand\natexlab[1]{#1}
\providecommand\showeprint[2][]{arXiv:#2}

\bibitem[\protect\citeauthoryear{Abels, Roijers, Lenaerts, Now{\'e}, and
  Steckelmacher}{Abels et~al\mbox{.}}{2019}]%
        {abels2019dynamic}
\bibfield{author}{\bibinfo{person}{Axel Abels}, \bibinfo{person}{Diederik
  Roijers}, \bibinfo{person}{Tom Lenaerts}, \bibinfo{person}{Ann Now{\'e}},
  {and} \bibinfo{person}{Denis Steckelmacher}.}
  \bibinfo{year}{2019}\natexlab{}.
\newblock \showarticletitle{Dynamic weights in multi-objective deep
  reinforcement learning}. In \bibinfo{booktitle}{\emph{International
  Conference on Machine Learning}}. PMLR, \bibinfo{pages}{11--20}.
\newblock


\bibitem[\protect\citeauthoryear{Andrychowicz, Wolski, Ray, Schneider, Fong,
  Welinder, McGrew, Tobin, Pieter~Abbeel, and Zaremba}{Andrychowicz
  et~al\mbox{.}}{2017}]%
        {andrychowicz2017hindsight}
\bibfield{author}{\bibinfo{person}{Marcin Andrychowicz}, \bibinfo{person}{Filip
  Wolski}, \bibinfo{person}{Alex Ray}, \bibinfo{person}{Jonas Schneider},
  \bibinfo{person}{Rachel Fong}, \bibinfo{person}{Peter Welinder},
  \bibinfo{person}{Bob McGrew}, \bibinfo{person}{Josh Tobin},
  \bibinfo{person}{OpenAI Pieter~Abbeel}, {and} \bibinfo{person}{Wojciech
  Zaremba}.} \bibinfo{year}{2017}\natexlab{}.
\newblock \showarticletitle{Hindsight experience replay}.
\newblock \bibinfo{journal}{\emph{Advances in neural information processing
  systems}}  \bibinfo{volume}{30} (\bibinfo{year}{2017}).
\newblock


\bibitem[\protect\citeauthoryear{Barrett and Narayanan}{Barrett and
  Narayanan}{2008}]%
        {barrett2008learning}
\bibfield{author}{\bibinfo{person}{Leon Barrett} {and} \bibinfo{person}{Srini
  Narayanan}.} \bibinfo{year}{2008}\natexlab{}.
\newblock \showarticletitle{Learning all optimal policies with multiple
  criteria}. In \bibinfo{booktitle}{\emph{Proceedings of the 25th international
  conference on Machine learning}}. \bibinfo{pages}{41--47}.
\newblock


\bibitem[\protect\citeauthoryear{Bekiros}{Bekiros}{2010}]%
        {bekiros2010heterogeneous}
\bibfield{author}{\bibinfo{person}{Stelios~D Bekiros}.}
  \bibinfo{year}{2010}\natexlab{}.
\newblock \showarticletitle{Heterogeneous trading strategies with adaptive
  fuzzy actor--critic reinforcement learning: A behavioral approach}.
\newblock \bibinfo{journal}{\emph{Journal of Economic Dynamics and Control}}
  \bibinfo{volume}{34}, \bibinfo{number}{6} (\bibinfo{year}{2010}),
  \bibinfo{pages}{1153--1170}.
\newblock


\bibitem[\protect\citeauthoryear{Bisht and Kumar}{Bisht and Kumar}{2020}]%
        {bisht2020deep}
\bibfield{author}{\bibinfo{person}{Kiran Bisht} {and} \bibinfo{person}{Arun
  Kumar}.} \bibinfo{year}{2020}\natexlab{}.
\newblock \showarticletitle{Deep Reinforcement Learning based Multi-Objective
  Systems for Financial Trading}. In \bibinfo{booktitle}{\emph{2020 5th IEEE
  International Conference on Recent Advances and Innovations in Engineering
  (ICRAIE)}}. IEEE, \bibinfo{pages}{1--6}.
\newblock


\bibitem[\protect\citeauthoryear{Castelletti, Pianosi, and
  Restelli}{Castelletti et~al\mbox{.}}{2012}]%
        {castelletti2012tree}
\bibfield{author}{\bibinfo{person}{Andrea Castelletti},
  \bibinfo{person}{Francesca Pianosi}, {and} \bibinfo{person}{Marcello
  Restelli}.} \bibinfo{year}{2012}\natexlab{}.
\newblock \showarticletitle{Tree-based fitted Q-iteration for multi-objective
  Markov decision problems}. In \bibinfo{booktitle}{\emph{The 2012
  international joint conference on neural networks (IJCNN)}}. IEEE,
  \bibinfo{pages}{1--8}.
\newblock


\bibitem[\protect\citeauthoryear{Chan and Shelton}{Chan and Shelton}{2001}]%
        {RePEc:sce:scecf1:146}
\bibfield{author}{\bibinfo{person}{Nicholas~T. Chan} {and}
  \bibinfo{person}{Christian Shelton}.} \bibinfo{year}{2001}\natexlab{}.
\newblock \bibinfo{booktitle}{\emph{An Adaptive Electronic Market-Maker}}.
\newblock \bibinfo{type}{Computing in Economics and Finance 2001} 146.
  \bibinfo{institution}{Society for Computational Economics}.
\newblock
\urldef\tempurl%
\url{https://EconPapers.repec.org/RePEc:sce:scecf1:146}
\showURL{%
\tempurl}


\bibitem[\protect\citeauthoryear{Corazza and Bertoluzzo}{Corazza and
  Bertoluzzo}{2014}]%
        {RePEc:ven:wpaper:2014:15}
\bibfield{author}{\bibinfo{person}{Marco Corazza} {and}
  \bibinfo{person}{Francesco Bertoluzzo}.} \bibinfo{year}{2014}\natexlab{}.
\newblock \bibinfo{booktitle}{\emph{Q-Learning-based financial trading systems
  with applications}}.
\newblock \bibinfo{type}{Working Papers} 2014:15.
  \bibinfo{institution}{Department of Economics, University of Venice "Ca'
  Foscari"}.
\newblock
\urldef\tempurl%
\url{https://EconPapers.repec.org/RePEc:ven:wpaper:2014:15}
\showURL{%
\tempurl}


\bibitem[\protect\citeauthoryear{Dempster and Leemans}{Dempster and
  Leemans}{2006}]%
        {dempster2006automated}
\bibfield{author}{\bibinfo{person}{Michael~AH Dempster} {and}
  \bibinfo{person}{Vasco Leemans}.} \bibinfo{year}{2006}\natexlab{}.
\newblock \showarticletitle{An automated FX trading system using adaptive
  reinforcement learning}.
\newblock \bibinfo{journal}{\emph{Expert Systems with Applications}}
  \bibinfo{volume}{30}, \bibinfo{number}{3} (\bibinfo{year}{2006}),
  \bibinfo{pages}{543--552}.
\newblock


\bibitem[\protect\citeauthoryear{Dempster, Payne, Romahi, and
  Thompson}{Dempster et~al\mbox{.}}{2001}]%
        {dempster2001computational}
\bibfield{author}{\bibinfo{person}{Michael~AH Dempster}, \bibinfo{person}{Tom~W
  Payne}, \bibinfo{person}{Yazann Romahi}, {and} \bibinfo{person}{Giles~WP
  Thompson}.} \bibinfo{year}{2001}\natexlab{}.
\newblock \showarticletitle{Computational learning techniques for intraday FX
  trading using popular technical indicators}.
\newblock \bibinfo{journal}{\emph{IEEE Transactions on neural networks}}
  \bibinfo{volume}{12}, \bibinfo{number}{4} (\bibinfo{year}{2001}),
  \bibinfo{pages}{744--754}.
\newblock


\bibitem[\protect\citeauthoryear{Deng, Bao, Kong, Ren, and Dai}{Deng
  et~al\mbox{.}}{2016}]%
        {deng2016deep}
\bibfield{author}{\bibinfo{person}{Yue Deng}, \bibinfo{person}{Feng Bao},
  \bibinfo{person}{Youyong Kong}, \bibinfo{person}{Zhiquan Ren}, {and}
  \bibinfo{person}{Qionghai Dai}.} \bibinfo{year}{2016}\natexlab{}.
\newblock \showarticletitle{Deep direct reinforcement learning for financial
  signal representation and trading}.
\newblock \bibinfo{journal}{\emph{IEEE transactions on neural networks and
  learning systems}} \bibinfo{volume}{28}, \bibinfo{number}{3}
  (\bibinfo{year}{2016}), \bibinfo{pages}{653--664}.
\newblock


\bibitem[\protect\citeauthoryear{Dusparic and Cahill}{Dusparic and
  Cahill}{2009}]%
        {dusparic2009distributed}
\bibfield{author}{\bibinfo{person}{Ivana Dusparic} {and} \bibinfo{person}{Vinny
  Cahill}.} \bibinfo{year}{2009}\natexlab{}.
\newblock \showarticletitle{Distributed w-learning: Multi-policy optimization
  in self-organizing systems}. In \bibinfo{booktitle}{\emph{2009 Third IEEE
  international conference on self-adaptive and self-organizing systems}}.
  IEEE, \bibinfo{pages}{20--29}.
\newblock


\bibitem[\protect\citeauthoryear{Eilers, Dunis, von Mettenheim, and
  Breitner}{Eilers et~al\mbox{.}}{2014}]%
        {eilers2014intelligent}
\bibfield{author}{\bibinfo{person}{Dennis Eilers}, \bibinfo{person}{Christian~L
  Dunis}, \bibinfo{person}{Hans-J{\"o}rg von Mettenheim}, {and}
  \bibinfo{person}{Michael~H Breitner}.} \bibinfo{year}{2014}\natexlab{}.
\newblock \showarticletitle{Intelligent trading of seasonal effects: A decision
  support algorithm based on reinforcement learning}.
\newblock \bibinfo{journal}{\emph{Decision support systems}}
  \bibinfo{volume}{64} (\bibinfo{year}{2014}), \bibinfo{pages}{100--108}.
\newblock


\bibitem[\protect\citeauthoryear{Ernst, Geurts, and Wehenkel}{Ernst
  et~al\mbox{.}}{2005}]%
        {ernst2005tree}
\bibfield{author}{\bibinfo{person}{Damien Ernst}, \bibinfo{person}{Pierre
  Geurts}, {and} \bibinfo{person}{Louis Wehenkel}.}
  \bibinfo{year}{2005}\natexlab{}.
\newblock \showarticletitle{Tree-based batch mode reinforcement learning}.
\newblock \bibinfo{journal}{\emph{Journal of Machine Learning Research}}
  \bibinfo{volume}{6} (\bibinfo{year}{2005}).
\newblock


\bibitem[\protect\citeauthoryear{Fedus, Ramachandran, Agarwal, Bengio,
  Larochelle, Rowland, and Dabney}{Fedus et~al\mbox{.}}{2020}]%
        {fedus2020revisiting}
\bibfield{author}{\bibinfo{person}{William Fedus}, \bibinfo{person}{Prajit
  Ramachandran}, \bibinfo{person}{Rishabh Agarwal}, \bibinfo{person}{Yoshua
  Bengio}, \bibinfo{person}{Hugo Larochelle}, \bibinfo{person}{Mark Rowland},
  {and} \bibinfo{person}{Will Dabney}.} \bibinfo{year}{2020}\natexlab{}.
\newblock \showarticletitle{Revisiting fundamentals of experience replay}. In
  \bibinfo{booktitle}{\emph{International Conference on Machine Learning}}.
  PMLR, \bibinfo{pages}{3061--3071}.
\newblock


\bibitem[\protect\citeauthoryear{Fischer}{Fischer}{2018}]%
        {fischer2018reinforcement}
\bibfield{author}{\bibinfo{person}{Thomas~G Fischer}.}
  \bibinfo{year}{2018}\natexlab{}.
\newblock \bibinfo{booktitle}{\emph{Reinforcement learning in financial
  markets-a survey}}.
\newblock \bibinfo{type}{{T}echnical {R}eport}. \bibinfo{institution}{FAU
  Discussion Papers in Economics}.
\newblock


\bibitem[\protect\citeauthoryear{Friedman and Fontaine}{Friedman and
  Fontaine}{2018}]%
        {friedman2018generalizing}
\bibfield{author}{\bibinfo{person}{Eli Friedman} {and} \bibinfo{person}{Fred
  Fontaine}.} \bibinfo{year}{2018}\natexlab{}.
\newblock \showarticletitle{Generalizing across multi-objective reward
  functions in deep reinforcement learning}.
\newblock \bibinfo{journal}{\emph{arXiv preprint arXiv:1809.06364}}
  (\bibinfo{year}{2018}).
\newblock


\bibitem[\protect\citeauthoryear{Gold}{Gold}{2003}]%
        {gold2003fx}
\bibfield{author}{\bibinfo{person}{Carl Gold}.}
  \bibinfo{year}{2003}\natexlab{}.
\newblock \showarticletitle{FX trading via recurrent reinforcement learning}.
  In \bibinfo{booktitle}{\emph{2003 IEEE International Conference on
  Computational Intelligence for Financial Engineering, 2003. Proceedings.}}
  IEEE, \bibinfo{pages}{363--370}.
\newblock


\bibitem[\protect\citeauthoryear{Gu, Mabu, Yang, Li, and Hirasawa}{Gu
  et~al\mbox{.}}{2011}]%
        {gu2011trading}
\bibfield{author}{\bibinfo{person}{Yunqing Gu}, \bibinfo{person}{Shingo Mabu},
  \bibinfo{person}{Yang Yang}, \bibinfo{person}{Jianhua Li}, {and}
  \bibinfo{person}{Kotaro Hirasawa}.} \bibinfo{year}{2011}\natexlab{}.
\newblock \showarticletitle{Trading rules on stock markets using Genetic
  Network Programming-Sarsa learning with plural subroutines}. In
  \bibinfo{booktitle}{\emph{SICE Annual Conference 2011}}. IEEE,
  \bibinfo{pages}{143--148}.
\newblock


\bibitem[\protect\citeauthoryear{Jangmin, Lee, Lee, and Zhang}{Jangmin
  et~al\mbox{.}}{2006}]%
        {jangmin2006adaptive}
\bibfield{author}{\bibinfo{person}{O Jangmin}, \bibinfo{person}{Jongwoo Lee},
  \bibinfo{person}{Jae~Won Lee}, {and} \bibinfo{person}{Byoung-Tak Zhang}.}
  \bibinfo{year}{2006}\natexlab{}.
\newblock \showarticletitle{Adaptive stock trading with dynamic asset
  allocation using reinforcement learning}.
\newblock \bibinfo{journal}{\emph{Information Sciences}} \bibinfo{volume}{176},
  \bibinfo{number}{15} (\bibinfo{year}{2006}), \bibinfo{pages}{2121--2147}.
\newblock


\bibitem[\protect\citeauthoryear{Jiang, Xu, and Liang}{Jiang
  et~al\mbox{.}}{2017}]%
        {jiang2017deep}
\bibfield{author}{\bibinfo{person}{Zhengyao Jiang}, \bibinfo{person}{Dixing
  Xu}, {and} \bibinfo{person}{Jinjun Liang}.} \bibinfo{year}{2017}\natexlab{}.
\newblock \showarticletitle{A deep reinforcement learning framework for the
  financial portfolio management problem}.
\newblock \bibinfo{journal}{\emph{arXiv preprint arXiv:1706.10059}}
  (\bibinfo{year}{2017}).
\newblock


\bibitem[\protect\citeauthoryear{Jin and El-Saawy}{Jin and El-Saawy}{2016}]%
        {jin2016portfolio}
\bibfield{author}{\bibinfo{person}{Olivier Jin} {and} \bibinfo{person}{Hamza
  El-Saawy}.} \bibinfo{year}{2016}\natexlab{}.
\newblock \showarticletitle{Portfolio management using reinforcement learning}.
\newblock \bibinfo{journal}{\emph{Stanford University}} (\bibinfo{year}{2016}).
\newblock


\bibitem[\protect\citeauthoryear{K{\"a}llstr{\"o}m and
  Heintz}{K{\"a}llstr{\"o}m and Heintz}{2019}]%
        {kallstrom2019tunable}
\bibfield{author}{\bibinfo{person}{Johan K{\"a}llstr{\"o}m} {and}
  \bibinfo{person}{Fredrik Heintz}.} \bibinfo{year}{2019}\natexlab{}.
\newblock \showarticletitle{Tunable dynamics in agent-based simulation using
  multi-objective reinforcement learning}. In
  \bibinfo{booktitle}{\emph{Adaptive and Learning Agents Workshop (ALA-19) at
  AAMAS, Montreal, Canada, May 13-14, 2019}}. \bibinfo{pages}{1--7}.
\newblock


\bibitem[\protect\citeauthoryear{Kaur}{Kaur}{2017}]%
        {kaur2017algorithmic}
\bibfield{author}{\bibinfo{person}{S Kaur}.} \bibinfo{year}{2017}\natexlab{}.
\newblock \bibinfo{booktitle}{\emph{Algorithmic trading using reinforcement
  learning augmented with hidden Markov model}}.
\newblock \bibinfo{type}{{T}echnical {R}eport}. \bibinfo{institution}{Working
  paper, Stanford University}.
\newblock


\bibitem[\protect\citeauthoryear{Kober, Bagnell, and Peters}{Kober
  et~al\mbox{.}}{2013}]%
        {kober2013reinforcement}
\bibfield{author}{\bibinfo{person}{Jens Kober}, \bibinfo{person}{J~Andrew
  Bagnell}, {and} \bibinfo{person}{Jan Peters}.}
  \bibinfo{year}{2013}\natexlab{}.
\newblock \showarticletitle{Reinforcement learning in robotics: A survey}.
\newblock \bibinfo{journal}{\emph{The International Journal of Robotics
  Research}} \bibinfo{volume}{32}, \bibinfo{number}{11} (\bibinfo{year}{2013}),
  \bibinfo{pages}{1238--1274}.
\newblock


\bibitem[\protect\citeauthoryear{Lee and Jangmin}{Lee and Jangmin}{2002}]%
        {lee2002multi}
\bibfield{author}{\bibinfo{person}{Jae~Won Lee} {and} \bibinfo{person}{O
  Jangmin}.} \bibinfo{year}{2002}\natexlab{}.
\newblock \showarticletitle{A multi-agent Q-learning framework for optimizing
  stock trading systems}. In \bibinfo{booktitle}{\emph{International Conference
  on Database and Expert Systems Applications}}. Springer,
  \bibinfo{pages}{153--162}.
\newblock


\bibitem[\protect\citeauthoryear{Lee, Park, Jangmin, Lee, and Hong}{Lee
  et~al\mbox{.}}{2007}]%
        {lee2007multiagent}
\bibfield{author}{\bibinfo{person}{Jae~Won Lee}, \bibinfo{person}{Jonghun
  Park}, \bibinfo{person}{O Jangmin}, \bibinfo{person}{Jongwoo Lee}, {and}
  \bibinfo{person}{Euyseok Hong}.} \bibinfo{year}{2007}\natexlab{}.
\newblock \showarticletitle{A multiagent approach to $ q $-learning for daily
  stock trading}.
\newblock \bibinfo{journal}{\emph{IEEE Transactions on Systems, Man, and
  Cybernetics-Part A: Systems and Humans}} \bibinfo{volume}{37},
  \bibinfo{number}{6} (\bibinfo{year}{2007}), \bibinfo{pages}{864--877}.
\newblock


\bibitem[\protect\citeauthoryear{Lee, Zhang, et~al\mbox{.}}{Lee
  et~al\mbox{.}}{2002}]%
        {lee2002stock}
\bibfield{author}{\bibinfo{person}{Jae~Won Lee}, \bibinfo{person}{Byoung-Tak
  Zhang}, {et~al\mbox{.}}} \bibinfo{year}{2002}\natexlab{}.
\newblock \showarticletitle{Stock trading system using reinforcement learning
  with cooperative agents}. In \bibinfo{booktitle}{\emph{Proceedings of the
  Nineteenth International Conference on Machine Learning}}.
  \bibinfo{pages}{451--458}.
\newblock


\bibitem[\protect\citeauthoryear{Li, Dagli, and Enke}{Li et~al\mbox{.}}{2007}]%
        {li2007short}
\bibfield{author}{\bibinfo{person}{Hailin Li}, \bibinfo{person}{Cihan~H Dagli},
  {and} \bibinfo{person}{David Enke}.} \bibinfo{year}{2007}\natexlab{}.
\newblock \showarticletitle{Short-term stock market timing prediction under
  reinforcement learning schemes}. In \bibinfo{booktitle}{\emph{2007 IEEE
  International Symposium on Approximate Dynamic Programming and Reinforcement
  Learning}}. IEEE, \bibinfo{pages}{233--240}.
\newblock


\bibitem[\protect\citeauthoryear{Mao, Alizadeh, Menache, and Kandula}{Mao
  et~al\mbox{.}}{2016}]%
        {mao2016resource}
\bibfield{author}{\bibinfo{person}{Hongzi Mao}, \bibinfo{person}{Mohammad
  Alizadeh}, \bibinfo{person}{Ishai Menache}, {and} \bibinfo{person}{Srikanth
  Kandula}.} \bibinfo{year}{2016}\natexlab{}.
\newblock \showarticletitle{Resource management with deep reinforcement
  learning}. In \bibinfo{booktitle}{\emph{Proceedings of the 15th ACM workshop
  on hot topics in networks}}. \bibinfo{pages}{50--56}.
\newblock


\bibitem[\protect\citeauthoryear{Mnih, Kavukcuoglu, Silver, Graves, Antonoglou,
  Wierstra, and Riedmiller}{Mnih et~al\mbox{.}}{2013}]%
        {mnih2013playing}
\bibfield{author}{\bibinfo{person}{Volodymyr Mnih}, \bibinfo{person}{Koray
  Kavukcuoglu}, \bibinfo{person}{David Silver}, \bibinfo{person}{Alex Graves},
  \bibinfo{person}{Ioannis Antonoglou}, \bibinfo{person}{Daan Wierstra}, {and}
  \bibinfo{person}{Martin Riedmiller}.} \bibinfo{year}{2013}\natexlab{}.
\newblock \showarticletitle{Playing atari with deep reinforcement learning}.
\newblock \bibinfo{journal}{\emph{arXiv preprint arXiv:1312.5602}}
  (\bibinfo{year}{2013}).
\newblock


\bibitem[\protect\citeauthoryear{Moody, Wu, Liao, and Saffell}{Moody
  et~al\mbox{.}}{1998}]%
        {moody1998performance}
\bibfield{author}{\bibinfo{person}{John Moody}, \bibinfo{person}{Lizhong Wu},
  \bibinfo{person}{Yuansong Liao}, {and} \bibinfo{person}{Matthew Saffell}.}
  \bibinfo{year}{1998}\natexlab{}.
\newblock \showarticletitle{Performance functions and reinforcement learning
  for trading systems and portfolios}.
\newblock \bibinfo{journal}{\emph{Journal of Forecasting}}
  \bibinfo{volume}{17}, \bibinfo{number}{5-6} (\bibinfo{year}{1998}),
  \bibinfo{pages}{441--470}.
\newblock


\bibitem[\protect\citeauthoryear{Mossalam, Assael, Roijers, and
  Whiteson}{Mossalam et~al\mbox{.}}{2016}]%
        {mossalam2016multi}
\bibfield{author}{\bibinfo{person}{Hossam Mossalam}, \bibinfo{person}{Yannis~M
  Assael}, \bibinfo{person}{Diederik~M Roijers}, {and} \bibinfo{person}{Shimon
  Whiteson}.} \bibinfo{year}{2016}\natexlab{}.
\newblock \showarticletitle{Multi-objective deep reinforcement learning}.
\newblock \bibinfo{journal}{\emph{arXiv preprint arXiv:1610.02707}}
  (\bibinfo{year}{2016}).
\newblock


\bibitem[\protect\citeauthoryear{Natarajan and Tadepalli}{Natarajan and
  Tadepalli}{2005}]%
        {natarajan2005dynamic}
\bibfield{author}{\bibinfo{person}{Sriraam Natarajan} {and}
  \bibinfo{person}{Prasad Tadepalli}.} \bibinfo{year}{2005}\natexlab{}.
\newblock \showarticletitle{Dynamic preferences in multi-criteria reinforcement
  learning}. In \bibinfo{booktitle}{\emph{Proceedings of the 22nd international
  conference on Machine learning}}. \bibinfo{pages}{601--608}.
\newblock


\bibitem[\protect\citeauthoryear{Neuneier}{Neuneier}{1995}]%
        {neuneier1995optimal}
\bibfield{author}{\bibinfo{person}{Ralph Neuneier}.}
  \bibinfo{year}{1995}\natexlab{}.
\newblock \showarticletitle{Optimal asset allocation using adaptive dynamic
  programming}.
\newblock \bibinfo{journal}{\emph{Advances in Neural Information Processing
  Systems}}  \bibinfo{volume}{8} (\bibinfo{year}{1995}).
\newblock


\bibitem[\protect\citeauthoryear{Nevmyvaka, Feng, and Kearns}{Nevmyvaka
  et~al\mbox{.}}{2006}]%
        {nevmyvaka2006reinforcement}
\bibfield{author}{\bibinfo{person}{Yuriy Nevmyvaka}, \bibinfo{person}{Yi Feng},
  {and} \bibinfo{person}{Michael Kearns}.} \bibinfo{year}{2006}\natexlab{}.
\newblock \showarticletitle{Reinforcement learning for optimized trade
  execution}. In \bibinfo{booktitle}{\emph{Proceedings of the 23rd
  international conference on Machine learning}}. \bibinfo{pages}{673--680}.
\newblock


\bibitem[\protect\citeauthoryear{Nguyen, Nguyen, Vamplew, Nahavandi, Dazeley,
  and Lim}{Nguyen et~al\mbox{.}}{2020}]%
        {nguyen2020multi}
\bibfield{author}{\bibinfo{person}{Thanh~Thi Nguyen}, \bibinfo{person}{Ngoc~Duy
  Nguyen}, \bibinfo{person}{Peter Vamplew}, \bibinfo{person}{Saeid Nahavandi},
  \bibinfo{person}{Richard Dazeley}, {and} \bibinfo{person}{Chee~Peng Lim}.}
  \bibinfo{year}{2020}\natexlab{}.
\newblock \showarticletitle{A multi-objective deep reinforcement learning
  framework}.
\newblock \bibinfo{journal}{\emph{Engineering Applications of Artificial
  Intelligence}}  \bibinfo{volume}{96} (\bibinfo{year}{2020}),
  \bibinfo{pages}{103915}.
\newblock


\bibitem[\protect\citeauthoryear{R{\u{a}}dulescu, Mannion, Roijers, and
  Now{\'e}}{R{\u{a}}dulescu et~al\mbox{.}}{2020}]%
        {ruadulescu2020multi}
\bibfield{author}{\bibinfo{person}{Roxana R{\u{a}}dulescu},
  \bibinfo{person}{Patrick Mannion}, \bibinfo{person}{Diederik~M Roijers},
  {and} \bibinfo{person}{Ann Now{\'e}}.} \bibinfo{year}{2020}\natexlab{}.
\newblock \showarticletitle{Multi-objective multi-agent decision making: a
  utility-based analysis and survey}.
\newblock \bibinfo{journal}{\emph{Autonomous Agents and Multi-Agent Systems}}
  \bibinfo{volume}{34}, \bibinfo{number}{1} (\bibinfo{year}{2020}),
  \bibinfo{pages}{1--52}.
\newblock


\bibitem[\protect\citeauthoryear{Reymond and Now{\'e}}{Reymond and
  Now{\'e}}{2019}]%
        {reymond2019pareto}
\bibfield{author}{\bibinfo{person}{Mathieu Reymond} {and} \bibinfo{person}{Ann
  Now{\'e}}.} \bibinfo{year}{2019}\natexlab{}.
\newblock \showarticletitle{Pareto-DQN: Approximating the Pareto front in
  complex multi-objective decision problems}. In
  \bibinfo{booktitle}{\emph{Proceedings of the adaptive and learning agents
  workshop (ALA-19) at AAMAS}}.
\newblock


\bibitem[\protect\citeauthoryear{Roijers, Vamplew, Whiteson, and
  Dazeley}{Roijers et~al\mbox{.}}{2013}]%
        {roijers2013survey}
\bibfield{author}{\bibinfo{person}{Diederik~M Roijers}, \bibinfo{person}{Peter
  Vamplew}, \bibinfo{person}{Shimon Whiteson}, {and} \bibinfo{person}{Richard
  Dazeley}.} \bibinfo{year}{2013}\natexlab{}.
\newblock \showarticletitle{A survey of multi-objective sequential
  decision-making}.
\newblock \bibinfo{journal}{\emph{Journal of Artificial Intelligence Research}}
   \bibinfo{volume}{48} (\bibinfo{year}{2013}), \bibinfo{pages}{67--113}.
\newblock


\bibitem[\protect\citeauthoryear{Sherstov and Stone}{Sherstov and
  Stone}{2004}]%
        {sherstov2004three}
\bibfield{author}{\bibinfo{person}{Alexander~A Sherstov} {and}
  \bibinfo{person}{Peter Stone}.} \bibinfo{year}{2004}\natexlab{}.
\newblock \showarticletitle{Three automated stock-trading agents: A comparative
  study}. In \bibinfo{booktitle}{\emph{International Workshop on Agent-Mediated
  Electronic Commerce}}. Springer, \bibinfo{pages}{173--187}.
\newblock


\bibitem[\protect\citeauthoryear{Si, Li, Ding, and Rao}{Si
  et~al\mbox{.}}{2017}]%
        {si2017multi}
\bibfield{author}{\bibinfo{person}{Weiyu Si}, \bibinfo{person}{Jinke Li},
  \bibinfo{person}{Peng Ding}, {and} \bibinfo{person}{Ruonan Rao}.}
  \bibinfo{year}{2017}\natexlab{}.
\newblock \showarticletitle{A multi-objective deep reinforcement learning
  approach for stock index future’s intraday trading}. In
  \bibinfo{booktitle}{\emph{2017 10th International symposium on computational
  intelligence and design (ISCID)}}, Vol.~\bibinfo{volume}{2}. IEEE,
  \bibinfo{pages}{431--436}.
\newblock


\bibitem[\protect\citeauthoryear{Sutton and Barto}{Sutton and Barto}{2018}]%
        {sutton2018reinforcement}
\bibfield{author}{\bibinfo{person}{Richard~S Sutton} {and}
  \bibinfo{person}{Andrew~G Barto}.} \bibinfo{year}{2018}\natexlab{}.
\newblock \bibinfo{booktitle}{\emph{Reinforcement learning: An introduction}}.
\newblock \bibinfo{publisher}{MIT press}.
\newblock


\bibitem[\protect\citeauthoryear{Tajmajer}{Tajmajer}{2017}]%
        {tajmajer2017multi}
\bibfield{author}{\bibinfo{person}{Tomasz Tajmajer}.}
  \bibinfo{year}{2017}\natexlab{}.
\newblock \showarticletitle{Multi-objective deep Q-learning with subsumption
  architecture}.
\newblock \bibinfo{journal}{\emph{arXiv preprint arXiv:1704.06676}}
  (\bibinfo{year}{2017}).
\newblock


\bibitem[\protect\citeauthoryear{Tajmajer}{Tajmajer}{2018}]%
        {tajmajer2018modular}
\bibfield{author}{\bibinfo{person}{Tomasz Tajmajer}.}
  \bibinfo{year}{2018}\natexlab{}.
\newblock \showarticletitle{Modular multi-objective deep reinforcement learning
  with decision values}. In \bibinfo{booktitle}{\emph{2018 Federated conference
  on computer science and information systems (FedCSIS)}}. IEEE,
  \bibinfo{pages}{85--93}.
\newblock


\bibitem[\protect\citeauthoryear{Tan, Quek, and Cheng}{Tan
  et~al\mbox{.}}{2011}]%
        {tan2011stock}
\bibfield{author}{\bibinfo{person}{Zhiyong Tan}, \bibinfo{person}{Chai Quek},
  {and} \bibinfo{person}{Philip~YK Cheng}.} \bibinfo{year}{2011}\natexlab{}.
\newblock \showarticletitle{Stock trading with cycles: A financial application
  of ANFIS and reinforcement learning}.
\newblock \bibinfo{journal}{\emph{Expert Systems with Applications}}
  \bibinfo{volume}{38}, \bibinfo{number}{5} (\bibinfo{year}{2011}),
  \bibinfo{pages}{4741--4755}.
\newblock


\bibitem[\protect\citeauthoryear{Watts}{Watts}{2015}]%
        {watts2015hedging}
\bibfield{author}{\bibinfo{person}{Samuel Watts}.}
  \bibinfo{year}{2015}\natexlab{}.
\newblock \bibinfo{booktitle}{\emph{Hedging basis risk using reinforcement
  learning}}.
\newblock \bibinfo{type}{{T}echnical {R}eport}. \bibinfo{institution}{Technical
  report, Working Paper, University of Oxford}.
\newblock


\bibitem[\protect\citeauthoryear{Zheng, Zhang, Zheng, Xiang, Yuan, Xie, and
  Li}{Zheng et~al\mbox{.}}{2018}]%
        {zheng2018drn}
\bibfield{author}{\bibinfo{person}{Guanjie Zheng}, \bibinfo{person}{Fuzheng
  Zhang}, \bibinfo{person}{Zihan Zheng}, \bibinfo{person}{Yang Xiang},
  \bibinfo{person}{Nicholas~Jing Yuan}, \bibinfo{person}{Xing Xie}, {and}
  \bibinfo{person}{Zhenhui Li}.} \bibinfo{year}{2018}\natexlab{}.
\newblock \showarticletitle{DRN: A deep reinforcement learning framework for
  news recommendation}. In \bibinfo{booktitle}{\emph{Proceedings of the 2018
  World Wide Web Conference}}. \bibinfo{pages}{167--176}.
\newblock


\bibitem[\protect\citeauthoryear{Zitzler, Knowles, and Thiele}{Zitzler
  et~al\mbox{.}}{2008}]%
        {zitzler2008quality}
\bibfield{author}{\bibinfo{person}{Eckart Zitzler}, \bibinfo{person}{Joshua
  Knowles}, {and} \bibinfo{person}{Lothar Thiele}.}
  \bibinfo{year}{2008}\natexlab{}.
\newblock \showarticletitle{Quality assessment of pareto set approximations}.
\newblock \bibinfo{journal}{\emph{Multiobjective optimization}}
  (\bibinfo{year}{2008}), \bibinfo{pages}{373--404}.
\newblock


\end{thebibliography}

\end{document}